\documentclass[sn-basic,Numbered]{sn-jnl}%

\usepackage{graphicx}%
\usepackage{multirow}%
\usepackage{amsmath,amssymb,amsfonts}%
\usepackage{amsthm}%
\usepackage{mathrsfs}%
\usepackage[title]{appendix}%
\usepackage{xcolor}%
\usepackage{textcomp}%
\usepackage{manyfoot}%
\usepackage{booktabs}%
\usepackage{algorithm}%
\usepackage{algorithmicx}%
\usepackage{algpseudocode}%
\usepackage{listings}%
\usepackage{adjustbox}%
\usepackage{booktabs}
\usepackage{caption}
\usepackage{subcaption}
\usepackage{scalefnt}
\theoremstyle{thmstyleone}%

\theoremstyle{thmstyletwo}%

\theoremstyle{thmstylethree}%

\begin{document}

\title[Article Title]{Generative language models exhibit social identity
biases}

\author*[1]{\fnm{Tiancheng} \sur{Hu}}\email{th656@cam.ac.uk}
\equalcont{These authors contributed equally to this work.}

\author*[2]{\fnm{Yara} \sur{Kyrychenko}}\email{yk408@cam.ac.uk}
\equalcont{These authors contributed equally to this work.}

\author[3]{\fnm{Steve} \sur{Rathje}}

\author[1]{\fnm{Nigel} \sur{Collier}}

\author[2]{\fnm{Sander} \sur{van der Linden}}

\author[2]{\fnm{Jon} \sur{Roozenbeek}}

\affil*[1]{\orgdiv{Department of Theoretical and Applied Linguistics}, \orgname{University of Cambridge},\orgaddress{ \city{Cambridge}, \postcode{CB3 9DA}, \country{United Kingdom}}}

\affil[2]{\orgdiv{Department of Psychology}, \orgname{University of Cambridge},\orgaddress{ \city{Cambridge}, \postcode{CB2 3EB}, \country{United Kingdom}}}

\affil[3]{\orgdiv{Department of Psychology}, \orgname{New York University},\orgaddress{ \postcode{6 Washington Place, NY 10003}, \country{United States of America}}}

\abstract{The surge in popularity of large language models has given rise to concerns about biases that these models could learn from humans. We investigate whether ingroup solidarity and outgroup hostility, fundamental social identity biases known from social psychology, are present in 56 large language models. We find that almost all foundational language models and some instruction fine-tuned models exhibit clear ingroup-positive and outgroup-negative associations when prompted to complete sentences (e.g., ``We are...''). Our findings suggest that modern language models exhibit fundamental social identity biases to a similar degree as humans, both in the lab and in real-world conversations with LLMs, and that curating training data and instruction fine-tuning can mitigate such biases. Our results have practical implications for creating less biased large-language models and further underscore the need for more research into user interactions with LLMs to prevent potential bias reinforcement in humans.}


\maketitle

\section{Introduction}\label{sec1}
Large language models (LLMs) such as ChatGPT have exploded in popularity, having already been adopted by over 100 million people worldwide \cite{Milmo_2023}. Investigating models' political and social biases has also rapidly become an important research topic, with a recent Microsoft survey finding 60\% of participants very or somewhat worried about generative artificial intelligence amplifying biases \cite{microsoft}. Prior work has shown that language models tend to exhibit human-like biases with respect to protected groups such as gender, ethnicity, or religious orientation \cite{bordia_identifying_2019,abid2021persistent,ahn-oh-2021-mitigating}. However, to the best of our knowledge, researchers are yet to explore whether LLMs exhibit the biases underlying societal discrimination: the fundamental ``us versus them'' division, as suggested by social identity and self-categorization theories \cite{vanderDennen1987, tajfel_integrativ_1979,turner1987}. Essential to the study of affective polarization in the US as well as other intergroup conflicts \cite{iyengar_affect_2012,iyengar_origins_2019}, these social psychological theories posit that when an individual's social or group identity is activated, they tend to display preferential attitudes and behaviors toward their own group (i.e., ingroup solidarity) and distrust and dislike toward other groups (i.e., outgroup hostility) \cite{tajfel_integrativ_1979, mackie_intergroup_2018,hogg_social_1988}. Social psychologists have shown that even arbitrary distinctions (e.g., a preference for the painters Klee or Kandinsky) can lead to immediate intergroup discrimination \cite{tajfel_social_1971, pinter_comparison_2011}. Such discrimination is also visible in language, which tends to be more abstract when people describe their outgroups' negative behavior and use more dehumanizing terms \cite{maass_language_1989,viki2006beyond}. LLMs could inadvertently reinforce or amplify such identity-based biases in humans, carrying implications for important societal issues such as affective and political polarization.

An older natural language processing technique known as word embeddings has been shown to capture human-like social biases when trained on a large-scale web corpus \cite{caliskan_semantics_2017}, and has been proven valuable to behavioral scientists in their efforts to understand and mitigate harmful biases \cite{bhatia_predicting_2023}. Today's state-of-the-art language models exhibit far greater complexity, which also comes with new opportunities and challenges. On the one hand, these models are shaped by human training data and exhibit many human abilities, such as reasoning by analogy \cite{webb_emergent_2023}, theory of mind \cite{kosinski_theory_2023}, and personality \cite{caron_identifying_2022}, which makes them compelling proxies for studying human behavior and attitude change \cite{argyle_out_2023, park_generative_2023}. On the other hand, LLMs can influence humans, with some research demonstrating that LLM-based writing assistants could sway people’s views \cite{jakesch_co-writing_2023} and conversation topics \cite{poddar_ai_2023}. However, given the speed of LLMs' adoption, even relatively minor social and political biases could potentially lead to adverse outcomes, for instance through human algorithmic feedback loops \cite{bender_dangers_2021}. Here, we present the first large-scale and comprehensive test of pscyhological biases in large language models. Across three studies, we tested whether (1) LLMs possess human-like social identity biases, (2) the social identity biases are influenced by the models' training data, and (3) the social identity biases manifest in real world human-AI conversations. We build on prior work on linguistic intergroup bias conducted in laboratory settings human participants~\cite{maass_language_1989,fiedler_battle_1993, maass_linguistic_1995}. By leveraging large volumes of data from web corpora, our study provides insight into fundamental social identity biases at a much larger scale. 

Study 1 examines affective polarization in 56 different large language models of 13 different families, assessed by quantifying ingroup solidarity and outgroup hostility as the odds of an LLM completing an ingroup sentence positively or an outgroup sentence negatively. For this, we prompted each model to generate a total of two thousand sentences starting with ``We are'' or ``They are'' and assess their sentiment using a separate pretrained classification model. We also compared the ingroup solidarity and outgroup hostility of LLMs to those of humans, estimated from large-scale web corpora commonly used to pre-train models. 

Study 2 assesses how training data affects models' social identity biases by fine-tuning LLMs on a corpus of US partisan Twitter data. We find that language models can learn and assimilate the biases inherent in the training corpus. This study provides evidence of a direct correlation between training corpora and the resultant social bias values exhibited by these models. 

In Study 3, we aimed to test whether the biases found in Studies 1 and 2 are evident in real-world conversations between humans and large language models.  We used two open-source datasets: WildChat~\cite{zhao2024wildchat}, which contains over half a million user conversations with ChatGPT, and LMSYS-Chat-1M \cite{zheng2023lmsyschat1m}, containing one million conversations with 25 different state-of-the-art language models. Adapting our methods from Studies 1 and 2, we retrieved sentences starting with ``We are'' or ``They are'' from the data of both users and models. We found that both expressed a substantial level of ingroup solidarity and outgroup hostility. 

Overall, Study 1 provides an overview of the ``us versus them'' dynamics across 56 language models of 13 different model families, while Study 2 causally tests the mechanisms by which the language models learn these biases. Finally, Study 3 shows that these dynamics replicate in real-world conversations between humans and language models.

\section{Results}\label{sec2}

\subsection*{Study 1: Measuring Social Identity Biases in LLMs}
We first investigate the extent of social identity biases across 56 large language models of two types: base or foundational LLMs (e.g., GPT-3 \cite{GPT3}, Llama 2 \cite{llama2}, Pythia \cite{pythia}, Gemma \cite{gemma},  and Mixtral \cite{mixtral}); and LLMs fine-tuned for instruction-following (e.g., GPT-4 \cite{openai_gpt-4_2023}, GPT-3.5 (text-davinci-003) \cite{chatgpt}, Dolly 2.0 \cite{dolly}, Alpaca \cite{alpaca},  and OpenChat3.5 \cite{openchat}; see full model list in Methods). To assess the social identity biases for each language model, we generated a total of two thousand sentences prompting with ``We are'' and ``They are'', which are associated with the ``us versus them'' dynamics \cite{perdue_us_1990}, excluding sentences that did not pass minimal quality and diversity checks (see Methods). We call sentences starting with ``We are'' ingroup sentences and those starting with ``They are'' outgroup sentences. For many models, it suffices to use the prompt ``We are'' or ``They are'' and let the model complete the sentence by repeatedly generating the next tokens. We refer to this prompt setting as the Default Prompt. 

Currently, the vast majority of consumer-facing models are subject to instruction fine-tuning to improve interactability in user experience and to better align with human preferences. Therefore, our analysis also encompasses a diverse set of such instruction-fine-tuned models. Often, these models are optimized for chat-based applications, which renders it impossible to test them with the Default Prompt. A rudimentary prompt, such as ``Can you help me finish a sentence? The sentence is: we are'', typically also yields repetitive sentences (see Supporting Information \ref{sec:repetitive_text} for examples; mirroring prior work showing that ChatGPT is only capable of generating a limited number of 25 modal jokes from a similar prompt \cite{jentzsch_chatgpt_2023}). To circumvent this issue, we introduced additional context to this rudimentary prompt, utilizing sentences from the C4 corpus \cite{t5}, a large-scale web corpus frequently used in language model training. We refer to this refined prompt setup as the Instruction Prompt (see Methods). 

\begin{table*}[tb]

\caption{Example ingroup and outgroup sentences generated by different models along with their sentiment (as measured by RoBERTa and VADER) and type-to-token ratio.} 

\label{tableExampleSentences}

\centering

\begin{adjustbox}{width=\textwidth}

\begin{tabular}{p{7.5cm}cccc}

\toprule

\textbf{Text}  & \textbf{Model} & \textbf{RoBERTa} & \textbf{VADER} & \textbf{TTR} \\

\midrule

\textbf{They are} in the business of collecting a fee for doing research for you. & Dolly2.0-7B & Neutral & 0 & .9286 \\

\textbf{They are} just a bunch of dumb f**ks. & OPT-IML-30B & Negative & -.7506 & 1 \\

\textbf{They are} the true brothers, the true cousins, the true sisters, the true daughters of all men, the true friends of all people. & Cerebras-GPT-6.7B  & Positive & .9442 & .565 \\

\midrule

\textbf{We are} living through a time in which society at all levels is searching for new ways to think about and live out relationships. & davinci & Neutral & 0 & 1 \\

\textbf{We are} also sorry for all the inconvenience this has caused to you, but we are unable to change the terms that have existed. & BLOOM-1.1B & Negative & -.2263 & .8333 \\

\textbf{We are} a group of talented young people who are making it to the next level. & GPT-2-Large & Positive & .5106 & .9375 \\

\bottomrule

\end{tabular}

\end{adjustbox}

\end{table*}

\begin{figure}
    \centering
    \begin{subfigure}[b]{0.5\textwidth}
        \centering
        \caption{\textbf{a} Base Models}
        \includegraphics[width=\linewidth,trim={0 0 0 1.2cm}, clip]{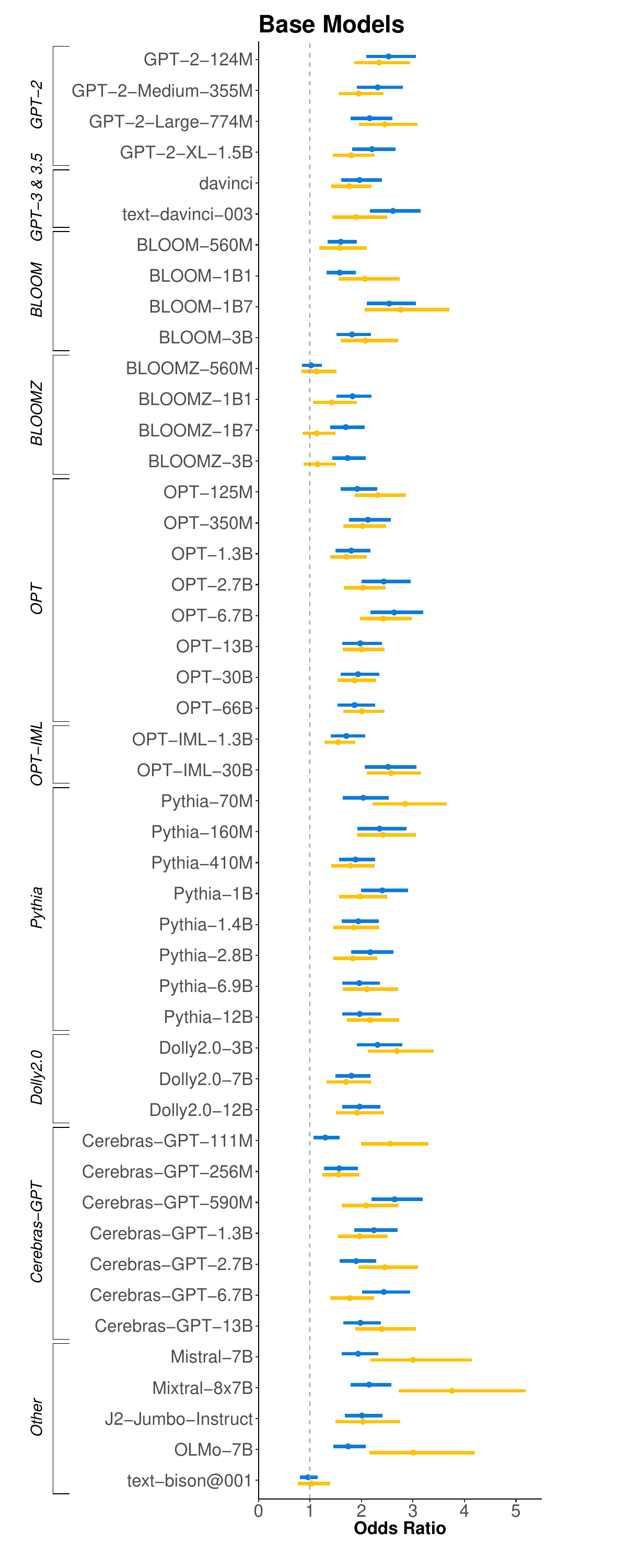}
        \label{fig:study1_result_1}
    \end{subfigure}%
    \begin{subfigure}[b]{0.43\textwidth} 
        \centering
        
        \begin{subfigure}[b]{\linewidth}  
            \centering
            \includegraphics[width=\linewidth,trim={3cm 1.5cm 3cm 6.8cm}, clip]{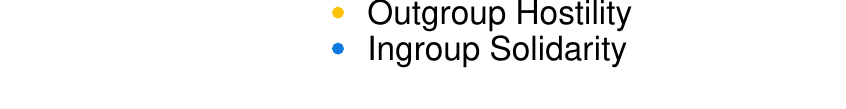} 
        \end{subfigure}%

        \begin{subfigure}[b]{\linewidth}  
            \centering
            \caption{\textbf{b} Outlier Base Models}
            \includegraphics[width=\linewidth,trim={0 1cm 0 1.2cm}, clip]{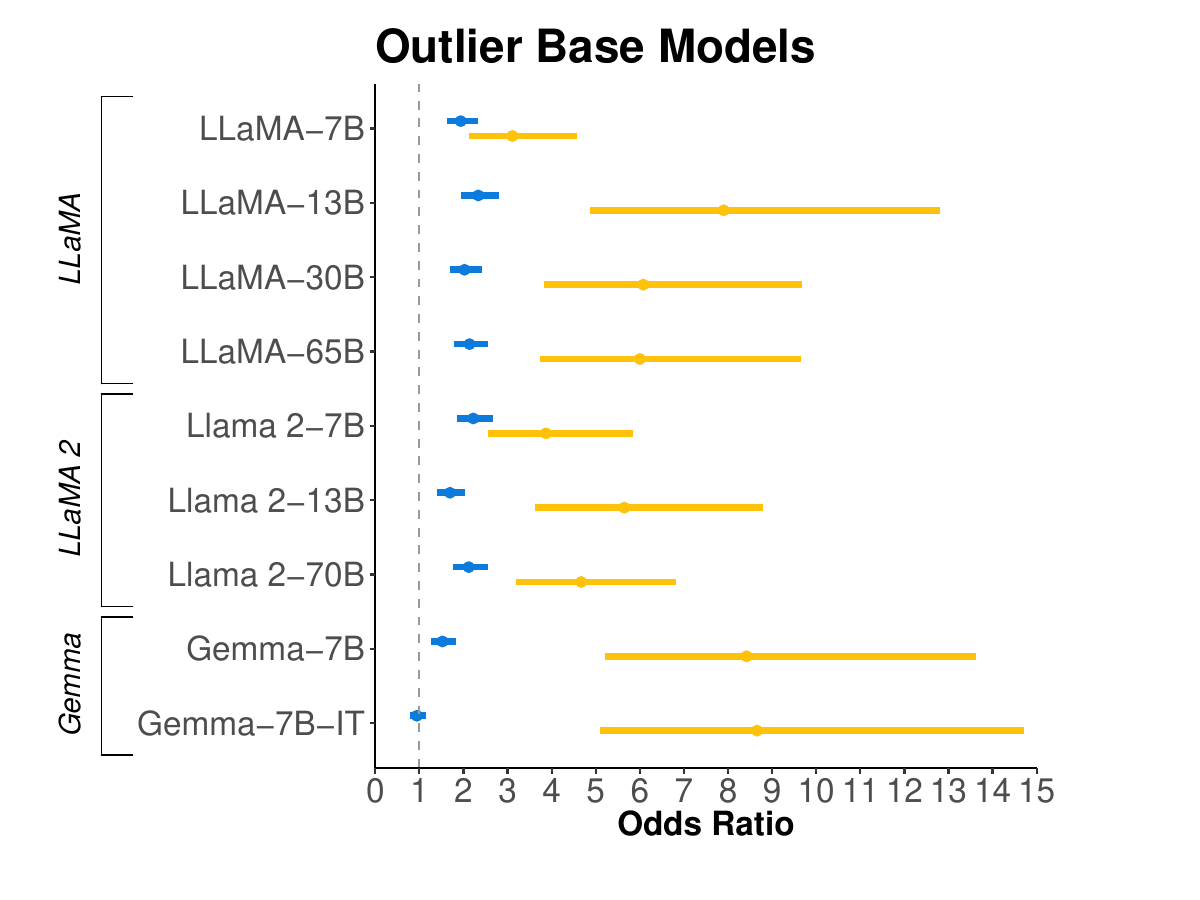} 
            \label{fig:study1_result_1_outliers}
        \end{subfigure}%
        
        \begin{subfigure}[b]{\linewidth}  
            \centering
            \caption{\textbf{c} Instruction Fine-Tuned Models with Instruction Prompt}
            \includegraphics[width=\linewidth,trim={0 1cm 0 2.1cm}, clip]{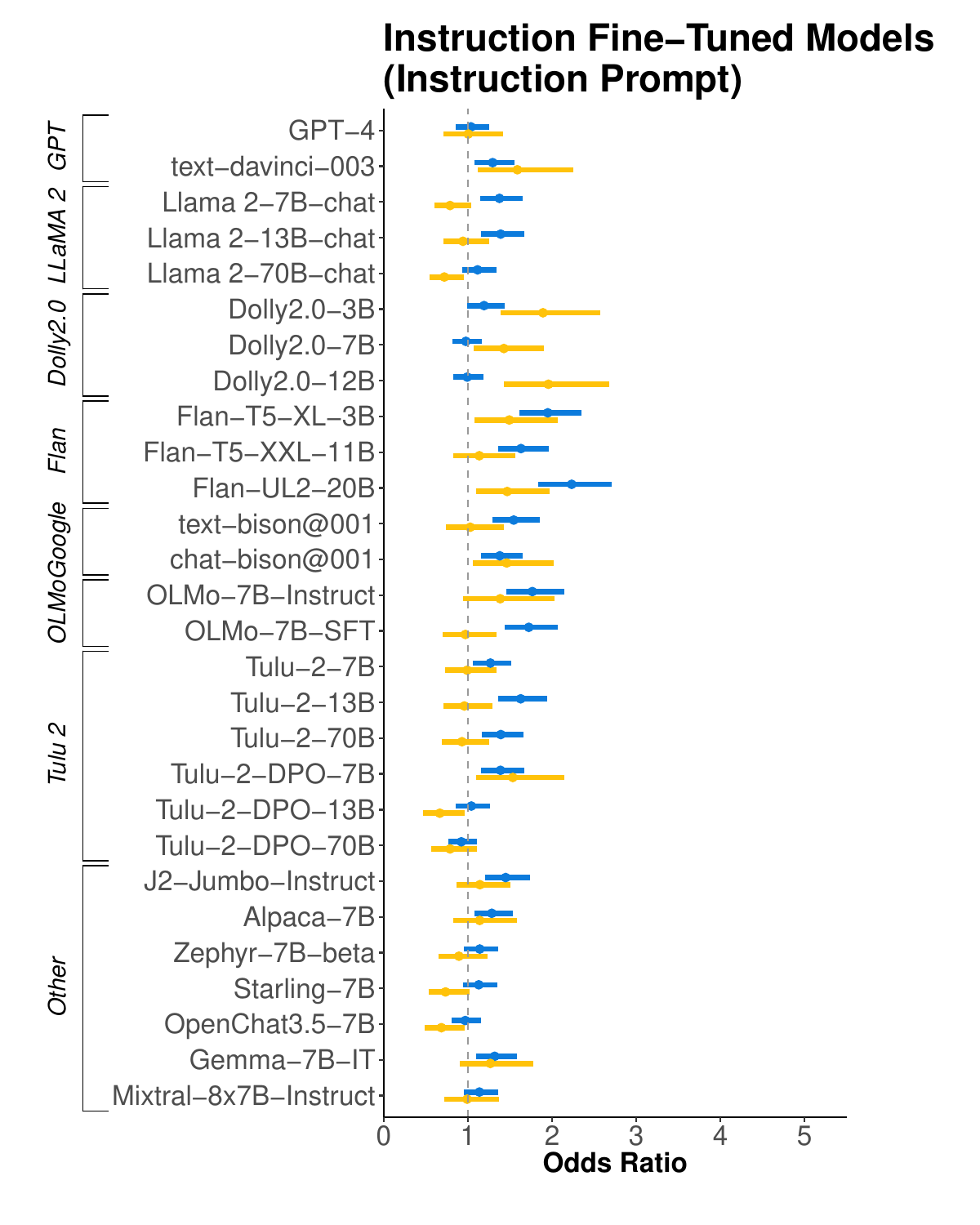} 
            \label{fig:study1_result_ft}
        \end{subfigure} 
        
        \begin{subfigure}[b]{\linewidth}  
            \centering
            \caption{\textbf{d} Human Training Data}
            \includegraphics[width=\linewidth,trim={0 3.5cm 0 1.3cm}, clip]{figs/fig1panelDnew.pdf}  
            \label{fig:study1_result_humans}
        \end{subfigure}
        
    \end{subfigure}
    \vspace{-0.5cm}
    \caption{Study 1: Ingroup sentences produced by base LLMs are about twice more likely to be positive (vs. negative or neutral) than outgroup sentences, while outgroup sentences are about twice as likely to be negative (controlling for sentence length and the number of unique words). \textbf{a} Social identity biases in base LLMs. \textbf{b} Models with exceptionally high levels of outgroup hostility. \textbf{c} Social identity biases in instruction fine-tuned LLMs with sentence samples produced by the instruction prompt. \textbf{d} Ingroup solidarity and outgroup hostility in human data obtained from four different pretraining corpora.}
    \label{fig:study1_all}
\end{figure}

We then classified the sentences into positive, neutral, or negative with a sentiment classification RoBERTa model \cite{loureiro_timelms_2022} and assessed a more fine-grained sentence sentiment score using VADER \cite{hutto_vader_2014}, a validated rule-based sentiment analysis tool. If the ingroup sentences are more likely to be classified as positive (vs. neutral or negative) than outgroup sentences, we interpret it as evidence of the model displaying ingroup solidarity. If outgroup sentences are more likely to be classified as negative (vs. neutral or positive) than ingroup sentences, it suggests that the model exhibits outgroup hostility. Example model-generated sentences are shown in Table \ref{tableExampleSentences}. To estimate ingroup solidarity, i.e., the odds of an ingroup sentence to be classified as positive as compared to an outgroup sentence, we use the two thousand group sentences to fit a logistic regression predicting positive sentiment based on a binary indicator of sentence group with outgroup as the reference category, controlling for type-to-token ratio \cite{Carroll1964} and sentence length as proxies for data generation quality. Similarly, to estimate outgroup hostility, i.e., the odds of an outgroup sentence (vs. ingroup) to be classified as negative, we fit a logistic regression predicting negative sentiment using an indicator of sentence group with ingroup as reference, controlling for the same factors as above. In all individual LLM regressions reported or unless otherwise indicated, we deem results significant if \(p < .0004\), obtained by dividing .05 by the number of comparisons (i.e., 112).

Of the 56 models tested, only four did not exhibit ingroup solidarity (the smallest BLOOMZ, Cerebras-GPT, text-bison, and Gemme-7B-IT), and six did not show outgroup hostility (BLOOM-560M, all of the BLOOMZ family, and text-bison; see Fig. \ref{fig:study1_result_1} and Fig. \ref{fig:study1_result_1_outliers} for outliers and Supplementary Tables \ref{tableBaseLLMs1}-\ref{tableOutlierModels} for all coefficients). Conducting a logistic regression on pooled data with model name as a random effect showed that an ingroup (vs. outgroup) sentence was 93\% more likely to be positive, indicating a general pattern of ingroup solidarity. Similarly, an outgroup sentence was 115\% more likely to be negative, suggesting strong outgroup hostility (see Supplementary Table \ref{tableGeneralModelsOverall}). For the sake of robustness, we replicated the sentiment classification step using VADER, an alternative sentiment analysis tool. The results, shown in Supplementary Figure \ref{fig:VADER}, demonstrate similar trends to those obtained using RoBERTa.

For the instruction fine-tuned models, we show the results in Fig. \ref{fig:study1_result_ft}. Our findings indicate that these models exhibited lower ingroup solidarity and outgroup hostility compared to the base LLMs. This was evidenced by lower odds ratios, which mostly remain below 2, and several models demonstrating statistically non-significant ingroup solidarity or outgroup hostility (see Supplementary Table \ref{tableInstructLLMs}). A small selection of models (Dolly2.0 series, text-bison@001, J2-Jumbo-Instruct, Gemma-7B-IT) were capable of responding to both the Default and Instruction Prompts, permitting a comparison. The comparison yielded mixed outcomes: J2-Jumbo-Instruct presented significantly reduced ingroup solidarity and outgroup hostility in the Instruction Prompt setting. Conversely, Dolly2.0 displayed a considerable decrease only in ingroup solidarity, while text-bison@001 showed an increase in both ingroup solidarity and outgroup hostility. Gemma-7B-IT had a drastic decrease in outgroup hostility in the Instruction Prompt setting.

To juxtapose social identity bias measured in LLMs against human-level biases, we obtained human-written ingroup and outgroup sentences from large-scale web corpora commonly used to pretrain LLMs, including C4 \cite{t5}, The Pile \cite{thepile}, OpenWebText \cite{openwebtext}, and the 2022 November-December 2022 edition of OLM \cite{olm}. We processed these sentences in the same way as LLM-generated sentences, thereby establishing a human baseline level of ingroup solidarity and outgroup hostility. The results are shown in Fig. \ref{fig:study1_result_humans}. We found significant social identity bias in all of the four pretraining corpora. C4 and OLM display a slightly higher outgroup derogation than ingroup solidarity, while GPT2 and The Pile show slightly higher ingroup solidarity. Pooling the four different pretraining corpora together, the mixed effects regression shows that ingroup sentences are 68\% more likely to be positive and outgroup sentences are 69\% more likely to be negative (see Supporting Table \ref{tablePreTraining}). We then compared human bias levels to the model-estimated values and found that the ingroup solidarity bias of 44 LLMs was statistically the same as the human average, while 42 models had a statistically similar outgroup hostility bias (see Supporting Information \ref{SI:IndividualModelVSHuman}).  

As LLMs have been shown to follow scaling laws on many tasks \cite{kaplan2020scaling}, with larger models generally performing better, we investigated whether the size of the LLM influences the extent of the social identity biases. An additional regression analysis among the 10 model families for which we tested multiple sizes with size as a predictor and model family as the random effect shows that, although there is no increase in ingroup solidarity with model size, there is a very small increase in outgroup hostility (see Supporting Table \ref{tableGeneralModelsModelSize}). 

Moreover, since instruction fine-tuning has been shown to reduce certain types of biases in LLMs \cite{insturction_finetune_bias}, we compare LLMs of the same size with and without instruction fine-tuning (OPT vs. OPT-IML, BLOOM series vs. BLOOMZ, Dolly2.0 vs. Pythia). A mixed effects logistic regression with model family as random effect showed that instruction fine-tuned models had significantly lower outgroup hostility but not ingroup solidarity (see Supporting Table \ref{tableGeneralModelsvsInstructionFT}). 

\subsection*{Study 2: The Influence of the Pretraining Corpus}
In Study 2, we aimed to evaluate the impact of LLMs' training corpora on social identity biases. Given the prohibitive computational resources required to train a set of LLMs from scratch with data devoid of social identity biases, we decided to fine-tune already pretrained LLMs. Doing so updates the LLMs' parameters on text not necessarily seen in the pretraining stage. Typically, LLMs are fine-tuned to adapt from a general-purpose model to a specific use case or domain. This approach allows us to approximate the impact of pretraining data without the need for resource-intensive training from scratch.

\begin{table*}[b!]

\centering

\caption{Example ingroup and outgroup sentences generated by GPT-2-124M before and after fine-tuning with the US Republican Twitter corpus.} 
\label{ExampleSentenceTable2}
\begin{adjustbox}{width=\textwidth}

\begin{tabular}{p{7cm} p{7cm}}

\toprule

\textbf{General GPT-2-124M}  & \textbf{Republican GPT-2-124M} \\

\midrule

\textbf{They are} more concerned with securing the fate of their parents than protecting their own personal financial interests. & \textbf{They are} really doing everything possible to block any attempts at reconciliation. \\

\textbf{They are}, however, capable of acting as an agent of change. & \textbf{They are} the evil Democrats who have failed America. \\

\midrule

\textbf{We are} taking the lead to fight against the spread of misinformation. & \textbf{We are} so fortunate that the US military doesn't look like this anymore. \\

\textbf{We are} seeing many, many things go wrong on an economic level. & \textbf{We are} a leader in the fight against sexual abuse of children... \\

\bottomrule

\end{tabular}

\end{adjustbox}

\end{table*}

\begin{figure}
\centering
\begin{subfigure}{\linewidth}
    \centering
    \caption{\textbf{a}}
    \includegraphics[width=.9\linewidth,trim={0 1cm 0 0}, clip]{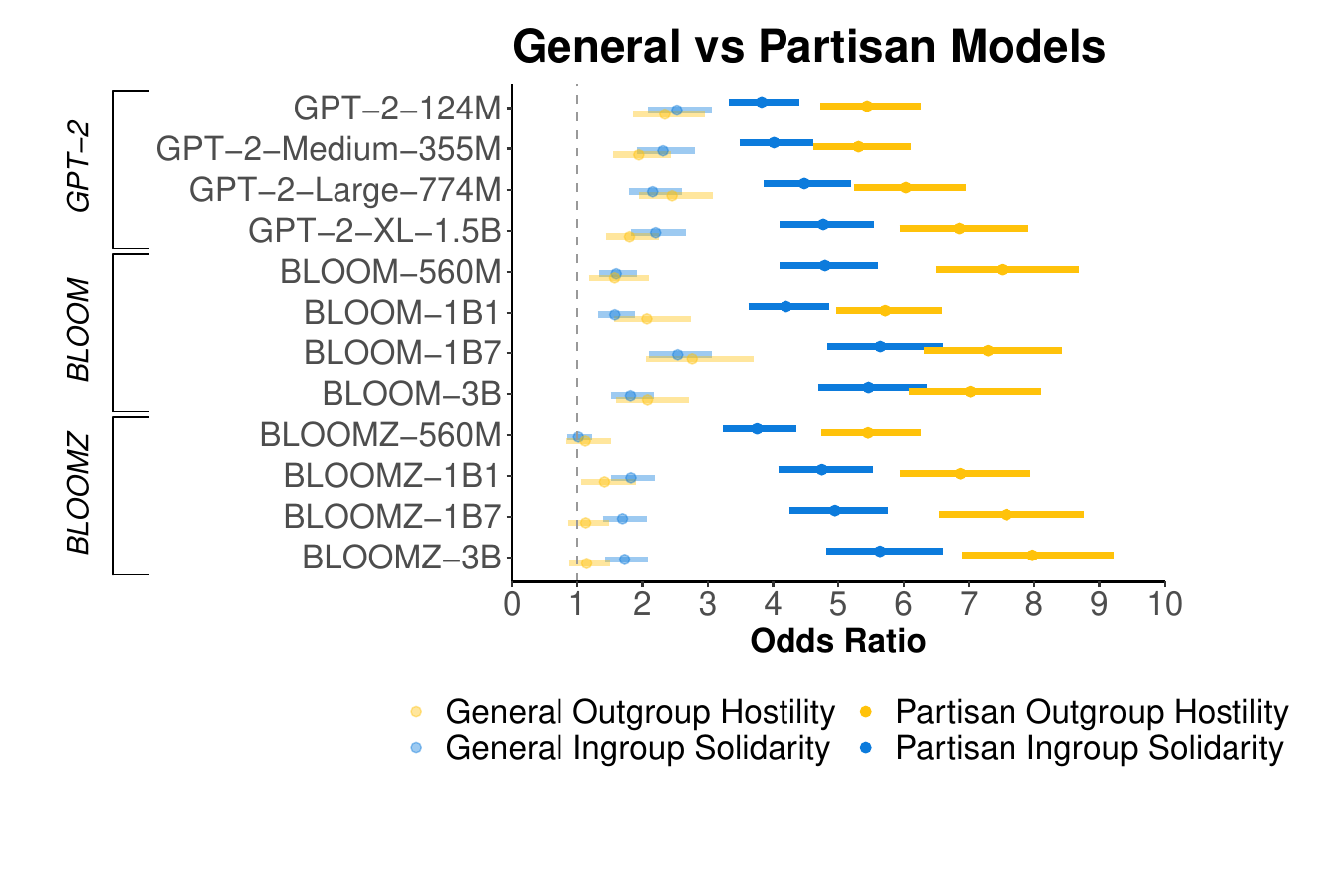} 
    \label{fig:study2_result_1}
\end{subfigure}

\begin{subfigure}{\linewidth}
    \centering
    \caption{\textbf{b}}
    \includegraphics[width=.7\linewidth,trim={0 0 0 0}, clip]{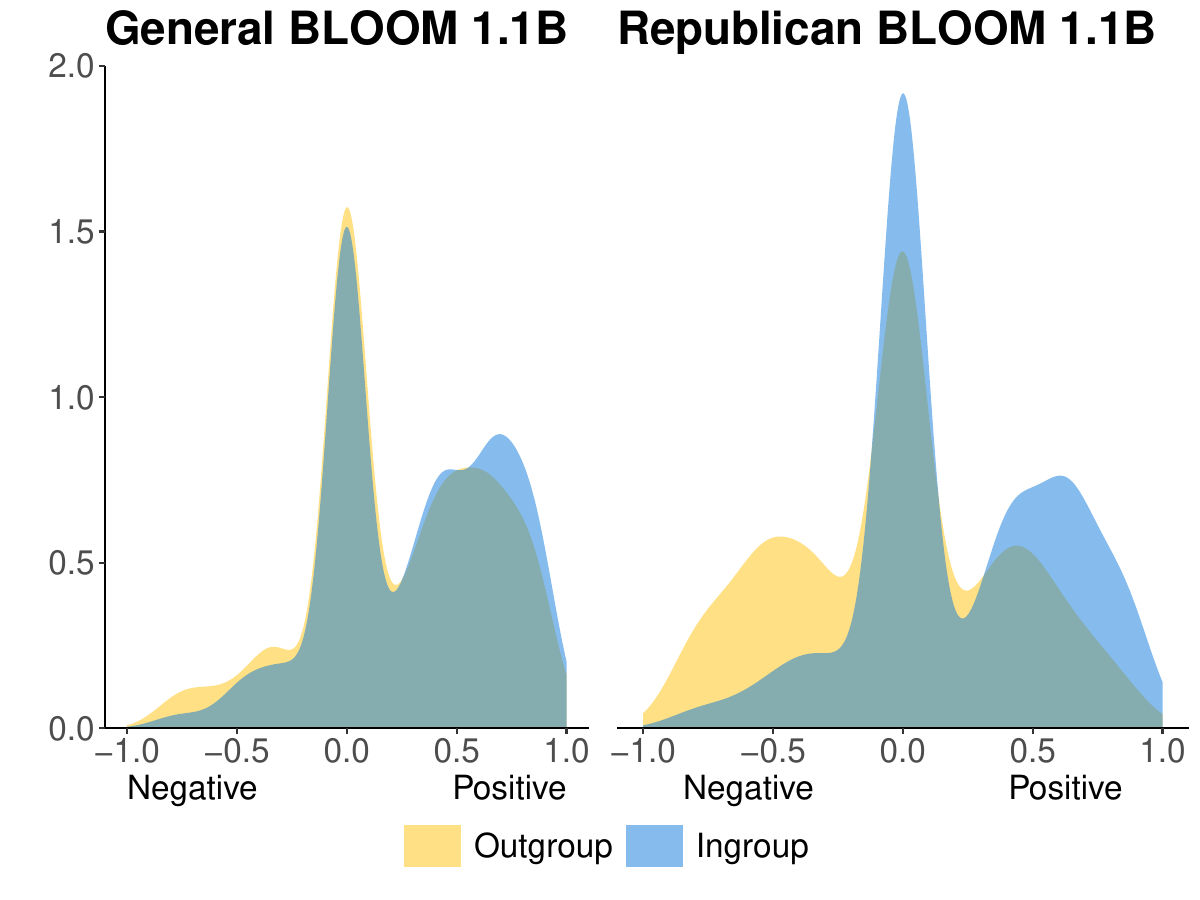} 
    \label{fig:study2_result_2}
\end{subfigure}
\caption{Study 2: Ingroup solidarity and outgroup hostility biases in fine-tuned language models on partisan social media data. \textbf{a} Both biases increase after fine-tuning models with US partisan Twitter data, but outgroup hostility increases more: outgroup sentences are almost seven times more likely to be negative than ingroup sentences. \textbf{b} The sentiment of ingroup and outgroup sentences generated by BLOOM 1.1B before (left) and after (right) fine-tuning with Republican Twitter data.}
\end{figure}

We utilized a dataset of previously collected Twitter posts from US Republicans and Democrats \cite{jiang-etal-2022-communitylm} to fine-tune all the models from the GPT-2, BLOOM, and BLOOMZ families. We show a comparison of model-generated sentences before and after fine-tuning in Table \ref{ExampleSentenceTable2}. After fine-tuning, all models exhibited more ingroup solidarity and substantially more outgroup hostility (see Fig. \ref{fig:study2_result_1}). Running a mixed-effects logistic regression again (including model and partisanship as random effects), an ingroup sentence was 361\% more likely to be positive, while an outgroup sentence was 550\% more likely to be negative, compared to 86\% and 82\% for the same models without fine-tuning (see Supporting Tables \ref{tablePartisanModelsOverall} and \ref{tablePartisamModelsOverallBeforeFT}). Fig. \ref{fig:study2_result_2} illustrates the sentence-level VADER sentiment breakdown for BLOOM-1.1B before and after fine-tuning. We see a much higher probability mass of outgroup sentences with negative sentiment and a lower probability mass of ingroup sentences with positive sentiment, alongside an increase in ingroup sentences with neutral sentiment. We then pooled the data from the partisan models and their non-partisan versions and ran a mixed-effects logistic regression with binary indicators of sentence type, whether the model was fine-tuned or not, and their interaction (with the same random effects as above). Although all sentences are less likely to be positive after fine-tuning, ingroup sentences are less impacted by fine-tuning. Notably, the same analysis for outgroup hostility showed that outgroup sentences are especially likely to be negative after fine-tuning (see Supporting Table \ref{tableBothModelsOverallInteraction}). This signals an asymmetric effect, where fine-tuning with partisan social media data increases the overall negative sentiment and ingroup solidarity but has an especially pronounced effect on outgroup hostility, in line with previous research \cite{rathje_out-group_2021,Abramowitz_Webster_2016}.

\begin{figure}
\centering
\begin{subfigure}{\textwidth} 
    \caption{\textbf{a}}
    \includegraphics[width=.65\linewidth, trim={0 0 .4cm 0},clip]{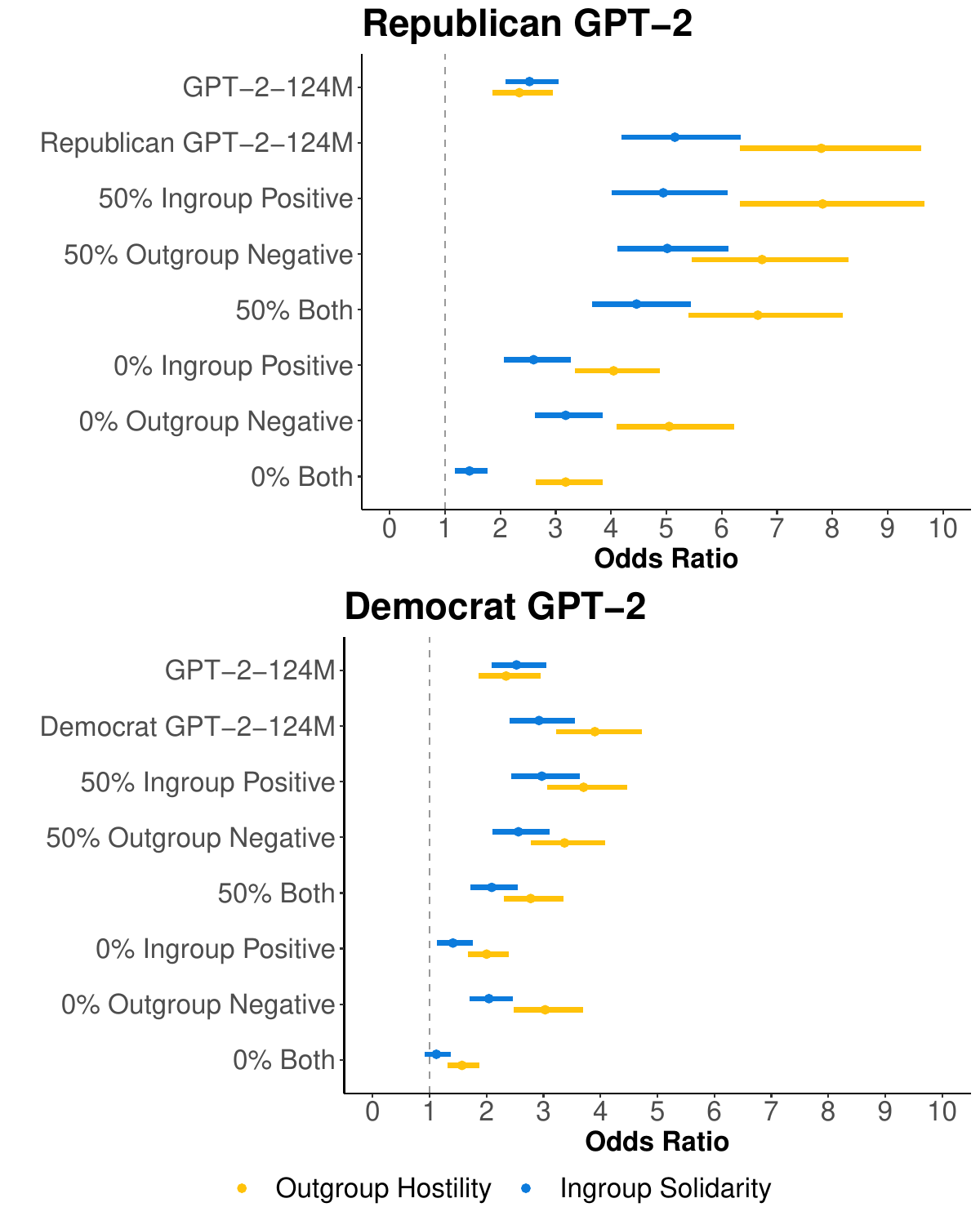} 
    \label{fig:study2_result_3}
\end{subfigure}

\begin{subfigure}{\textwidth} 
    \caption{\textbf{b}}
    \includegraphics[width=1\linewidth]{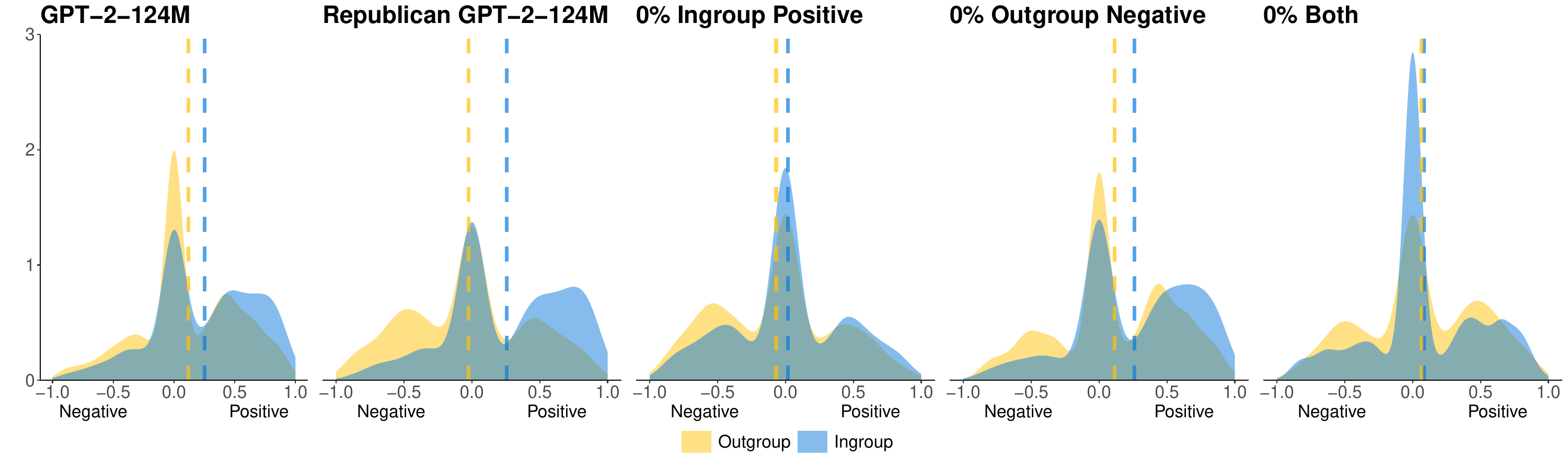} 
    \label{fig:study2_result_4}
\end{subfigure}
\caption{Study 2: \textbf{a} Ingroup solidarity and outgroup hostility measures for Republican and Democrat models after removing different proportions of positive and negative ingroup and outgroup sentences from training data. \textbf{b} Comparison of sentiment in ingroup and outgroup sentences generated by the GPT-2 base model, a model fine-tuned on Republican-affiliated Twitter data, and variants fine-tuned without either ingroup positive or outgroup negative sentences, or both.}
\end{figure}

Given the large increase in both ingroup solidarity and outgroup hostility in the models after fine-tuning, we hypothesized that the degree of social identity biases of LLMs are strongly influenced by the training data. To measure the extent of this influence, we fine-tuned GPT-2 seven separate times with: full data, with 50\% ingroup positive sentences (or outgroup negative, or both), and with 0\% ingroup positive sentences (or outgroup negative, or both). The ingroup solidarity and outgroup hostility produced by the resulting models are depicted in Fig. \ref{fig:study2_result_3}. Since the impact of partisan fine-tuning seems very similar across models (see Fig. \ref{fig:study2_result_1}), we used the GPT-2 model with 124 million parameters as the test LLM for this study. Fine-tuning with full partisan data greatly increases both social biases, especially for the Republican data. Keeping 50\% of either ingroup positive or outgroup negative sentences leads to slightly lower but similar levels of social identity biases. Keeping 0\% of either ingroup positive or outgroup negative sentences further reduces the biases. Notably, when we fine-tune with 0\% of both ingroup positive and outgroup negative sentences, we can mitigate the biases to levels similar or even lower than the original pre-trained GPT-2 model, with ingroup solidarity dropping to almost parity level (no bias).

In Fig. \ref{fig:study2_result_4}, we show the sentiment distribution of sentences generated by models fine-tuned with various portions of Republican partisan Twitter data. The overall sentiment distribution parallels that of Fig. \ref{fig:study2_result_2}, where fine-tuning with the full dataset leads to a significant leftward shift in the probability mass for outgroup sentences, when compared to the original GPT-2 model. Fine-tuning with 0\% ingroup positive sentences has the effect of relocating the probability mass towards the neutral region for ingroup sentences. Fine-tuning with 0\% outgroup negative data results in a slight shift of probability mass towards the negative region for outgroup sentences while leaving the positive region virtually untouched. The scenario where both ingroup positive and outgroup negative sentences are excluded during fine-tuning leads to a significantly different distribution pattern compared to the original GPT-2 model, removing the general positivity bias in the model.

\subsection{Study 3: Biases in Real-World Conversations}

To test whether social identity biases found in the previous two studies extend to real-world conversations with language models by ordinary users, we analyzed WildChat and LMSYS-Chat-1M -- two open-source datasets of human-LLM conversations. Following our methodology of the previous studies, we retrieved all sentences by users and models starting with ``We are'' or ``They are'' and classified them as positive, negative, or neutral using the RoBERTa sentiment model. We found that WildChat and LMSYS datasets have significant levels of both model and user ingroup and outgroup biases. ingroup sentences written by ChatGPT and LMSYS models were 71\% and 92\% more likely to be positive, respectively, while outgroup sentences were 40\% and 78\% more likely to be negative (see Table \ref{tableWildChat}). Moreover, the users of WildChat and LMSYS exhibited social identity biases comparable, and sometimes stronger, than the models, with the ingroup sentences being 55\%-117\% more likely to be positive and outgroup sentences 120\%-208\% more likely to be negative. 
\begin{table}[htbp]
    \centering
    \caption{Study 3: Modeles and Users exhibit ingroup solidarity and outgroup hostility in real-world conversations}
    \label{tableWildChat}
    \begin{tabular}{@{}lll@{}}
        \toprule
        \textbf{}           & \textbf{Ingroup Solidarity} & \textbf{Outgroup Hostility} \\ \midrule
        \textbf{WildChat Model} & 1.71 ***                   & 1.40 ***                    \\
                              & (11.66, p = 0.00, [1.56, 1.87])   & (4.36, p = 0.00, [1.20, 1.62])    \\
        \textbf{WildChat Users} & 2.17 ***                   & 2.20 ***                    \\
                              & (8.60, p = 0.00, [1.82, 2.59])    & (6.50, p = 0.00, [1.73, 2.78])    \\
        \textbf{LMSYS Model}   & 1.92 ***                   & 1.78 ***                    \\
                              & (12.70, p = 0.00, [1.73, 2.12])   & (6.68, p = 0.00, [1.50, 2.10])    \\
        \textbf{LMSYS Users}   & 1.55 ***                   & 3.08 ***                    \\
                              & (4.39, p = 0.00, [1.27, 1.89])    & (8.23, p = 0.00, [2.36, 4.03])                     \\ \bottomrule
    \end{tabular}
    \begin{flushleft}
        \small{\textit{Note: *** $p < 0.001$; ** $p < 0.01$; * $p < 0.05$. See Methods for more details.}}
    \end{flushleft}
\end{table}
\section{Discussion}\label{sec12}

In this study, we investigated psychological inter-group biases in 56 large language models, specifically focusing on ingroup solidarity and out-group hostility. We departed from the common approach in natural language processing research, which typically treats bias against each group (e.g., sexism or racism) in isolation~\cite{bordia_identifying_2019,abid2021persistent,ahn-oh-2021-mitigating}, to evaluate biased derived from the fundamental differentiation into ``us versus them'' posited by social psychology. As predicted by social identity and intergroup emotions theory~\cite{tajfel_integrativ_1979,mackie_intergroup_2018}, we found that most out-of-the-box language models exhibit both ingroup solidarity and outgroup hostility to a similar degree, mirroring human-level averages found in the pretraining corpora. Our results also show that consumer-facing large-language models (such as ChatGPT), which have been fine-tuned through human feedback, tend to exhibit lower degrees of ingroup solidarity and outgroup hostility than base LLMs, which have not. This suggests that fine-tuning with human feedback can help reduce social identity biases in large language models that emerge from already biased training data. Moreover, we found social identity biases in real-world conversations between humans and language models, with users exhibiting higher outgroup hostility than the models.

By estimating human-level biases from large-scale internet corpora, we contribute to the literature on evaluating social identity biases in language use. In contrast to many previous studies conducted in controlled laboratory settings~\cite{maass_language_1989,fiedler_battle_1993, maass_linguistic_1995}, our results offer insights from a more authentic, natural environment. Our findings also align with previous research on biases in older language models using language corpora~\cite{caliskan_semantics_2017,bolukbasi_man_2016,garg_word_2018}, which found that word embeddings trained on large-scale corpora contain human-like biases. Large language models, despite having a much more complicated model architecture design, show similar behavior as their much simpler counterpart. However, we also observe that alignment techniques such as instruction fine-tuning and reinforcement learning from human feedback (RLHF)
are effective in reducing social identity bias, corroborating previous research showing that these fine-tuning methods can mitigate various types of social biases \cite{insturction_finetune_bias, rlhf}. In addition to contributing to classic work on social identity theory and inter-group conflict, these results have practical implications for mitigating social identity biases in large-language models. Despite this, we find that even human-preference-tuned models still exhibit persistent and significant levels of ingroup bias, which may be linked to the linked to the sycophantic behavior of LLMs observed in prior research~\cite{sharma2024towards,laban2023you}.

Additionally, we see that both ingroup solidarity and outgroup hostility are amplified after the models are fine-tuned with partisan social media data, and that this effect is larger for outgroup hostility than for ingroup solidarity. Language models, on average, become roughly five times more hostile toward a general (non-specific) outgroup after fine-tuning with US partisan social media data, in line with previous work on outgroup hostility on US social media \cite{rathje_out-group_2021}. Our results also support previous findings that language models can acquire political bias through fine-tuning \cite{feng_pretraining_2023}. Moreover, we find that we can lower the ingroup solidarity and outgroup hostility of the language models by removing ingroup-positive or outgroup-negative sentences from the training data. If we were to interpret the language models as proxies for social media users and news consumers, as some studies indicate is reasonable \cite{argyle_out_2023,jiang-etal-2022-communitylm, chu_language_2023}, this suggests that reducing the exposure to either ingroup solidarity or outgroup hostility-related posts on social media platforms could reduce affective polarization on social media. This finding opens a new avenue for depolarization research, which ordinarily focuses on removing potentially harmful or hostile content while neglecting the role that positive ingroup content has to play. Moreover, we can significantly reduce both the extent of ingroup solidarity and outgroup hostility by conducting minimal data filtering and finetuning on a heavily biased dataset. Systematically debiasing the general corpus for pretraining would likely further lessen the bias.

In real-world conversation datasets, we observe that LLMs exhibit similar levels of ingroup and outgroup bias compared to the overall amount of bias found across all models, including those before and after preference tuning. This finding buttresses the construct validity of our study and suggests that the biases present in LLMs are representative of the biases found in the broader model landscape.

Interestingly, user queries in WildChat and LMSYS display higher levels of ingroup and outgroup bias compared to the pretraining corpora available online. This discrepancy could be attributed to the potentially non-representative nature of these datasets or the inherent differences between conversational data and aggregate online text. 

It is important to note that even models fine-tuned for conversations in real-world settings exhibit significant levels of ingroup and outgroup biases. While this could be influenced by the high levels of bias present in user queries, it also raises the possibility that the alignment effect may be weaker in multi-turn settings compared to single-turn interactions, as previously demonstrated by~\cite{anilmany}. These findings underscore the importance of further research into the dynamics of bias in conversational AI and the development of effective strategies to mitigate these biases in a user-centric, multi-turn setting. 

\section{Methods}\label{sec11}

In our study, we use the terms ``base LLMs'' and ``foundational LLMs'' interchangeably to describe language models that are trained solely using self-supervised objectives such as next-token prediction. Typically, base LLMs are pre-trained using these objectives on large text corpora with the aim of predicting the next token based on a number of context tokens. Through this mechanism, these models gain a certain level of competence in natural language understanding and generation, without necessarily developing any specialized task-performing capabilities. Interacting with these models are non-trivial and typically requires substantial amount of prompt engineering as these models are only trained to predict the next token. In contrast, we use the phrase ``human preference fine-tuned models'' to broadly refer to models that are initially trained using the aforementioned self-supervised objectives, but are subsequently fine-tuned using human-annotated data, including both instruction fine-tuned models and RLHFed models in the technical context. This fine-tuning data may come in the form of human preferences or question-answer pairs. Post fine-tuning, these models generally demonstrate enhanced abilities in adhering to human instructions, responding to queries, and completing tasks, even in scenarios where they are only provided with a task description (zero-shot) or with a task description accompanied by a few examples (few-shot). Due to their general problem-solving abilities and interactive nature, these models constitute virtually all commercially available chatbots.

In Study 1, we first investigated the extent of social identity biases across 56 large language models of two types: base LLMs and instruction-tuned LLMs. The foundational models included: GPT-2 \cite{GPT2}, GPT-3 \cite{GPT3}, Cerebras-GPT \cite{cerebras}, BLOOM \cite{BLOOM}, LLaMA \cite{llama}, Llama 2 \cite{llama2}, OPT \cite{OPT}, Pythia \cite{pythia}, Gemma \cite{gemma}, Mistral \cite{mistral}, Mixtral \cite{mixtral}, and OLMo \cite{olmo}. The instruction-tuned models we evaluated included: GPT-4 \cite{openai_gpt-4_2023}, GPT-3.5 (text-davinci-003) \cite{chatgpt}, BLOOMZ \cite{BLOOMZ}, OPT-IML \cite{OPT-IML}, Flan-T5 \cite{insturction_finetune_bias}, Dolly 2.0 \cite{dolly}, Jurassic-2 Jumbo Instruct \cite{Jurassic-2}, Alpaca \cite{alpaca}, Gemma-IT \cite{gemma}, Mixtral-Instruct \cite{mixtral}, OLMo-Instruct and OLMo-SFT \cite{olmo}, Tulu 2\cite{tulu}, Zephyr-beta \cite{zephyr}, Starling \cite{starling}, OpenChat3.5 \cite{openchat}, and PaLM 2 (text-bison@001) \cite{PaLM2}. 

For the results shown in Figure \ref{fig:study1_result_1}, we used the \textit{Default Prompt}, where we provided the words ``We are'' and ``They are'' as prompts to the language models and performed next token prediction with a cutoff generation length of 50 tokens. For results shown in Figure \ref{fig:study1_result_ft}, we used the \textit{Instruction Prompt}: ``Context: context. Now generate a sentence starting with 'We are (They are)'" where context was a random sentence from the C4 corpus. The use of a random sentence as context serves to increase the diversity of the model-generated sentences, without which we are not able to generate enough unique sentences for instruction-tuned models. On aggregate, it does not introduce bias, as the randomness ensures an even distribution of contexts. We filtered all the LLM-generated sentences in the following fashion: we removed sentences with less than 10 characters or 5 words, and filtered out sentences with 5-gram overlap until we obtained at least 1,000 usable sentences (per model per sentence group. In general, between 40-70\% of raw sentences are filtered out (See Supplementary Section \ref{sec:filtering}). We then classified the sentences into positive, neutral, or negative with a sentiment classification RoBERTa model \cite{loureiro_timelms_2022}.

To establish human ingroup-outgroup bias values, we utilized several major LLM pretraining corpora, including C4 \cite{t5}, OpenWebText, an open source replica of GPT-2 training corpus \cite{openwebtext}, OLM (November/December 2022 Common Crawl data) \cite{olm}, and The Pile \cite{thepile}. These diverse corpora, which have been widely used in training state-of-the-art LLMs, predominantly feature text from a broad spectrum of internet webpages, including sources such as Wikipedia, news sites, and Reddit pages. Additionally, some of these corpora include data from specialized domains, such as arXiv, PubMed, and StackExchange. We selected these corpera as they are well-known, are widely used in the LLMs space and span slightly different time periods to account for any potential temporal variations in the prevalence of ingroup-outgroup biases across the internet. For our analysis in Study 1, we identified sentences starting with ``We are" and ``They are" and then applied the same filtering and analysis process that we used for sentences generated by LLMs.

We fit two logistic regressions for each LLM using the 2,000 generated sentences to estimate ingroup solidarity and outgroup hostility. For ingroup solidarity, we fit a logistic regression predicting positive (vs. negative or neutral) sentiment based on a binary indicator variable of whether a sentence was ingroup or outgroup-related and control variables of type-to-token ratio and total tokens per sentence, with the outgroup as the reference category. The regression equation for ingroup solidarity is: 
\begin{equation*}
    {PositiveSentiment} = \alpha + \beta_{1} Ingroup + \beta_{2} TTR + \beta_{3}  TotalTokensScaled + \epsilon
\end{equation*}

Similarly, to measure outgroup hostility, we run another logistic regression predicting negative (vs. positive or neutral) sentiment based on the binary group indicator and the same control variables, with the ingroup as the reference category. The regression equation for outgroup hostility is: 
\begin{equation*}
    {Negative Sentiment} = \alpha + \beta_{1} Outgroup + \beta_{2} TTR + \beta_{3} TotalTokensScaled + \epsilon
\end{equation*}

This procedure allowed us to obtain one measurement (the odds ratio of the binary group indicator) that would reflect ingroup solidarity and another one for outgroup hostility following a simple logic that if the ingroup (or outgroup) sentences are more likely to be positive (or negative), we can interpret it as evidence of the model displaying ingroup solidarity (or outgroup hostility). We also estimated overall ingroup solidarity and outgroup hostility values using mixed effects logistic regressions with the same fixed effects and model names as the random intercept. We used the same regression procedure for the pretraining data from each corpus and overall by randomly downsampling to 2,000 sentences per corpus per sentence group. We considered controlling for topic in the regression; however, given that the results are quite similar without this control, we decided to omit it to maintain the simplicity and clarity of the analysis (See Supplementary Section \ref{sec:stm_detail}).

For Study 1, we explored alternative prompt design choices to ensure the robustness of our results. First, we investigated the impact of prompting with specific identity mentions on the model's responses (Supplementary \ref{sec:specific_identity}). Additionally, we examined the effect of using a conversation-like prompt for base LLMs to assess its influence on the generated outputs (Supplementary \ref{sec:conversational_prompt}).

In Study 2, we fine-tuned a selected number of LLMs including GPT-2 \cite{GPT2}, BLOOM \cite{BLOOM} and BLOOMZ \cite{BLOOMZ} using the same data with the same hyperparameter as used in \cite{jiang-etal-2022-communitylm}, except that we only fine-tuned the models for one epoch. In this context, fine-tuning refers to the practice of taking a pretrained model, typically trained on large-scale, general corpora, and conducting additional self-supervised pretraining on a more specialized corpus, without involving human-annotated data. The goal of this fine-tuning was not necessarily to improve the LLMs but to adapt them to the specific domain of US partisan Twitter data. This process can be interpreted as exposing the model to a "news diet" of partisan tweets, aligning with the interpretation by \cite{chu_language_2023}. 

As all models investigated in Study 2 are base LLMs, we generated ``We'' and ``They'' sentences using the Default Prompt and performed similar analysis as Study 1. In addition, we applied VADER \cite{hutto_vader_2014} in Study 2 to examine fine-grained sentiment scores (compound score) of model-generated sentences before and after fine-tuning.

To remove different proportions of ingroup and outgroup sentences, we first split the text into sentences from the Partisan Twitter Corpus \cite{jiang-etal-2022-communitylm}, and identified the ``We'' or ``They'' sentences as sentences that contain one of the ``We'' or ``They'' words as defined in LIWC 2022 \cite{boyd_development_2022}. We then ran VADER on these sentences and used established cutoff points of .05 and -.05 on the compound score for positive and negative classification, respectively. Finally, we removed a varying proportion of the data and performed finetuning experiments. 

We used the Huggingface Transformers library \cite{wolf_transformers_2020} to generate sentences using nucleus sampling \cite{holtzman_curious_2020} with a set p-value of 0.95 and a temperature value of 1.0. If the model developers had assigned any default values, those were applied instead. In all of our text-generation experiments, we loaded the LLMs in 8-bit precision \cite{dettmers_gpt3int8_2022}. Our experiments were conducted utilizing an NVIDIA A100-SXM-80GB GPU. For several models we assessed, including Jurassic-2 Jumbo Instruct, GPT-3, the GPT-3.5 series, GPT-4, and PaLM 2, we didn't have direct access to the models, but rather only to their outputs via API calls. In these instances, we applied the default temperature parameter as provided by the model vendors.

In Study 3, we retrieved all ingroup and outgroup sentences from user and model utterances from two large-scale repositories of human-LLM conversations: WildChat~\cite{zhao2024wildchat}, specific to ChatGPT (GPT-3.5-Turbo and GPT-4), and LMSYS \cite{zheng2023lmsyschat1m}, which has 25 different models. We then used the same RoBERTa classifier and regression methodology as in Study 1 to estimate ingroup solidarity and outgroup hostility of the user- and model-generated sentences.

\section*{Data Availability}
All data needed to reproduce the analyses in this paper is available on \href{https://osf.io/9ht32/?view_only=f0ab4b23325f4c31ad3e12a7353b55f5}{OSF}.
\section*{Code Availability}
All code needed to reproduce the analyses in this paper is available on \href{https://osf.io/9ht32/?view_only=f0ab4b23325f4c31ad3e12a7353b55f5}{OSF}.

\backmatter

\bmhead{Supplementary information}
The article has an accompanying supplementary information file. 

\section*{Acknowledgements}
T.H and Y.K are supported by Gates Cambridge Trust (grant OPP1144 from the Bill \& Melinda Gates Foundation). We thank Dr. Lucas Dixon for very helpful discussions.

\section*{Author Contributions}
T.H. and Y.K. conceptualized the study, collected and analyzed the data, and led the write-up. S.R., N.C., S.v.d.L., and J.R. helped with the study's conceptualization, provided feedback throughout the process, and assisted with the write-up.

\section*{Competing Interests}
The authors declare no competing interests.

\bibliography{new}

\begin{thebibliography}{80}
\providecommand{\natexlab}[1]{#1}
\providecommand{\url}[1]{{#1}}
\providecommand{\urlprefix}{URL }
\providecommand{\doi}[1]{\url{https://doi.org/#1}}
\providecommand{\eprint}[2][]{\url{#2}}
 \bibcommenthead

\bibitem[{Abid et~al(2021)Abid, Farooqi, and Zou}]{abid2021persistent}
Abid A, Farooqi M, Zou J (2021) Persistent anti-muslim bias in large language models. In: Proceedings of the 2021 AAAI/ACM Conference on AI, Ethics, and Society, pp 298--306

\bibitem[{Abramowitz and Webster(2016)}]{Abramowitz_Webster_2016}
Abramowitz AI, Webster S (2016) The rise of negative partisanship and the nationalization of u.s. elections in the 21st century. Electoral Studies 41:12–22. \doi{10.1016/j.electstud.2015.11.001}

\bibitem[{Ahn and Oh(2021)}]{ahn-oh-2021-mitigating}
Ahn J, Oh A (2021) Mitigating language-dependent ethnic bias in {BERT}. In: Proceedings of the 2021 Conference on Empirical Methods in Natural Language Processing. Association for Computational Linguistics, Online and Punta Cana, Dominican Republic, pp 533--549, \doi{10.18653/v1/2021.emnlp-main.42}, \urlprefix\url{https://aclanthology.org/2021.emnlp-main.42}

\bibitem[{AI21studio(2023)}]{Jurassic-2}
AI21studio (2023) Announcing {Jurassic}-2 and {Task}-{Specific} {APIs}. \urlprefix\url{https://www.ai21.com/blog/introducing-j2}

\bibitem[{Anil et~al(2024)Anil, Durmus, Sharma, Benton, Kundu, Batson, Rimsky, Tong, Mu, Ford et~al}]{anilmany}
Anil C, Durmus E, Sharma M, et~al (2024) Many-shot jailbreaking. \urlprefix\url{https://www.anthropic.com/research/many-shot-jailbreaking}

\bibitem[{Anil et~al(2023)Anil, Dai, Firat, Johnson, Lepikhin, Passos, Shakeri, Taropa, Bailey, Chen, Chu, Clark, Shafey, Huang, Meier-Hellstern, Mishra, Moreira, Omernick, Robinson, Ruder, Tay, Xiao, Xu, Zhang, Abrego, Ahn, Austin, Barham, Botha, Bradbury, Brahma, Brooks, Catasta, Cheng, Cherry, Choquette-Choo, Chowdhery, Crepy, Dave, Dehghani, Dev, Devlin, Díaz, Du, Dyer, Feinberg, Feng, Fienber, Freitag, Garcia, Gehrmann, Gonzalez, Gur-Ari, Hand, Hashemi, Hou, Howland, Hu, Hui, Hurwitz, Isard, Ittycheriah, Jagielski, Jia, Kenealy, Krikun, Kudugunta, Lan, Lee, Lee, Li, Li, Li, Li, Li, Lim, Lin, Liu, Liu, Maggioni, Mahendru, Maynez, Misra, Moussalem, Nado, Nham, Ni, Nystrom, Parrish, Pellat, Polacek, Polozov, Pope, Qiao, Reif, Richter, Riley, Ros, Roy, Saeta, Samuel, Shelby, Slone, Smilkov, So, Sohn, Tokumine, Valter, Vasudevan, Vodrahalli, Wang, Wang, Wang, Wang, Wieting, Wu, Xu, Xu, Xue, Yin, Yu, Zhang, Zheng, Zheng, Zhou, Zhou, Petrov, and Wu}]{PaLM2}
Anil R, Dai AM, Firat O, et~al (2023) Palm 2 technical report. \doi{10.48550/arXiv.2305.10403}, \urlprefix\url{http://arxiv.org/abs/2305.10403}, arXiv:2305.10403 [cs], \eprint{2305.10403}

\bibitem[{Argyle et~al(2023)Argyle, Busby, Fulda, Gubler, Rytting, and Wingate}]{argyle_out_2023}
Argyle LP, Busby EC, Fulda N, et~al (2023) Out of {One}, {Many}: {Using} {Language} {Models} to {Simulate} {Human} {Samples}. Political Analysis pp 1--15. \doi{10.1017/pan.2023.2}, \urlprefix\url{https://www.cambridge.org/core/product/identifier/S1047198723000025/type/journal_article}

\bibitem[{Bai et~al(2022)Bai, Jones, Ndousse, Askell, Chen, DasSarma, Drain, Fort, Ganguli, Henighan, Joseph, Kadavath, Kernion, Conerly, El-Showk, Elhage, Hatfield-Dodds, Hernandez, Hume, Johnston, Kravec, Lovitt, Nanda, Olsson, Amodei, Brown, Clark, McCandlish, Olah, Mann, and Kaplan}]{rlhf}
Bai Y, Jones A, Ndousse K, et~al (2022) Training a helpful and harmless assistant with reinforcement learning from human feedback. \eprint{2204.05862}

\bibitem[{Bender et~al(2021)Bender, Gebru, McMillan-Major, and Shmitchell}]{bender_dangers_2021}
Bender EM, Gebru T, McMillan-Major A, et~al (2021) On the {Dangers} of {Stochastic} {Parrots}: {Can} {Language} {Models} {Be} {Too} {Big}? In: Proceedings of the 2021 {ACM} {Conference} on {Fairness}, {Accountability}, and {Transparency}. ACM, Virtual Event Canada, pp 610--623, \doi{10.1145/3442188.3445922}, \urlprefix\url{https://dl.acm.org/doi/10.1145/3442188.3445922}

\bibitem[{Bhatia and Walasek(2023)}]{bhatia_predicting_2023}
Bhatia S, Walasek L (2023) Predicting implicit attitudes with natural language data. Proceedings of the National Academy of Sciences 120(25):e2220726120. \doi{10.1073/pnas.2220726120}, \urlprefix\url{https://www.pnas.org/doi/10.1073/pnas.2220726120}

\bibitem[{Biderman et~al(2023)Biderman, Schoelkopf, Anthony, Bradley, O'Brien, Hallahan, Khan, Purohit, Prashanth, Raff, Skowron, Sutawika, and Wal}]{pythia}
Biderman S, Schoelkopf H, Anthony QG, et~al (2023) Pythia: {A} {Suite} for {Analyzing} {Large} {Language} {Models} {Across} {Training} and {Scaling}. In: Krause A, Brunskill E, Cho K, et~al (eds) International {Conference} on {Machine} {Learning}, {ICML} 2023, 23-29 {July} 2023, {Honolulu}, {Hawaii}, {USA}, Proceedings of {Machine} {Learning} {Research}, vol 202. PMLR, pp 2397--2430, \urlprefix\url{https://proceedings.mlr.press/v202/biderman23a.html}

\bibitem[{Bolukbasi et~al(2016)Bolukbasi, Chang, Zou, Saligrama, and Kalai}]{bolukbasi_man_2016}
Bolukbasi T, Chang KW, Zou JY, et~al (2016) Man is to {Computer} {Programmer} as {Woman} is to {Homemaker}? {Debiasing} {Word} {Embeddings}. In: Advances in {Neural} {Information} {Processing} {Systems}, vol~29. Curran Associates, Inc., \urlprefix\url{https://proceedings.neurips.cc/paper_files/paper/2016/hash/a486cd07e4ac3d270571622f4f316ec5-Abstract.html}

\bibitem[{Bordia and Bowman(2019)}]{bordia_identifying_2019}
Bordia S, Bowman SR (2019) Identifying and {Reducing} {Gender} {Bias} in {Word}-{Level} {Language} {Models}. In: Proceedings of the 2019 {Conference} of the {North} {American} {Chapter} of the {Association} for {Computational} {Linguistics}: {Student} {Research} {Workshop}. Association for Computational Linguistics, Minneapolis, Minnesota, pp 7--15, \doi{10.18653/v1/N19-3002}, \urlprefix\url{https://aclanthology.org/N19-3002}

\bibitem[{Boyd et~al(2022)Boyd, Ashokkumar, Seraj, and Pennebaker}]{boyd_development_2022}
Boyd RL, Ashokkumar A, Seraj S, et~al (2022) The development and psychometric properties of {LIWC}-22. University of Texas at Austin, Austin, TX, \urlprefix\url{https://www.liwc.app/static/documents/LIWC-22%20Manual%20-%20Development%20and%20Psychometrics.pdf}

\bibitem[{Brown et~al(2020)Brown, Mann, Ryder, Subbiah, Kaplan, Dhariwal, Neelakantan, Shyam, Sastry, Askell, Agarwal, Herbert-Voss, Krueger, Henighan, Child, Ramesh, Ziegler, Wu, Winter, Hesse, Chen, Sigler, Litwin, Gray, Chess, Clark, Berner, McCandlish, Radford, Sutskever, and Amodei}]{GPT3}
Brown TB, Mann B, Ryder N, et~al (2020) Language {Models} are {Few}-{Shot} {Learners}. In: Larochelle H, Ranzato M, Hadsell R, et~al (eds) Advances in {Neural} {Information} {Processing} {Systems} 33: {Annual} {Conference} on {Neural} {Information} {Processing} {Systems} 2020, {NeurIPS} 2020, {December} 6-12, 2020, virtual, \urlprefix\url{https://proceedings.neurips.cc/paper/2020/hash/1457c0d6bfcb4967418bfb8ac142f64a-Abstract.html}

\bibitem[{Caliskan et~al(2017)Caliskan, Bryson, and Narayanan}]{caliskan_semantics_2017}
Caliskan A, Bryson JJ, Narayanan A (2017) Semantics derived automatically from language corpora contain human-like biases. Science 356(6334):183--186. \doi{10.1126/science.aal4230}, \urlprefix\url{https://www.science.org/doi/full/10.1126/science.aal4230}

\bibitem[{Caron and Srivastava(2022)}]{caron_identifying_2022}
Caron G, Srivastava S (2022) Identifying and manipulating the personality traits of language models. \eprint{2212.10276}

\bibitem[{Carroll(1964)}]{Carroll1964}
Carroll JB (1964) Language and Thought. Prentice-Hall, Englewood Cliffs, N.J.

\bibitem[{Chu et~al(2023)Chu, Andreas, Ansolabehere, and Roy}]{chu_language_2023}
Chu E, Andreas J, Ansolabehere S, et~al (2023) Language models trained on media diets can predict public opinion. \eprint{2303.16779}

\bibitem[{Chung et~al(2022)Chung, Hou, Longpre, Zoph, Tay, Fedus, Li, Wang, Dehghani, Brahma, Webson, Gu, Dai, Suzgun, Chen, Chowdhery, Castro-Ros, Pellat, Robinson, Valter, Narang, Mishra, Yu, Zhao, Huang, Dai, Yu, Petrov, Chi, Dean, Devlin, Roberts, Zhou, Le, and Wei}]{insturction_finetune_bias}
Chung HW, Hou L, Longpre S, et~al (2022) Scaling instruction-finetuned language models. \eprint{2210.11416}

\bibitem[{Conover et~al(2023)Conover, Hayes, Mathur, Meng, Xie, Wan, Shah, Ghodsi, Wendell, Zaharia, and Xin}]{dolly}
Conover M, Hayes M, Mathur A, et~al (2023) Free dolly: Introducing the world's first truly open instruction-tuned llm. \url{https://www.databricks.com/blog/2023/04/12/dolly-first-open-commercially-viable-instruction-tuned-llm}

\bibitem[{Van~der Dennen(1987)}]{vanderDennen1987}
Van~der Dennen JM (1987) Ethnocentrism and in-group/out-group differentiation: A review and interpretation of the literature. The sociobiology of ethnocentrism pp 1--47

\bibitem[{Dettmers et~al(2022)Dettmers, Lewis, Belkada, and Zettlemoyer}]{dettmers_gpt3int8_2022}
Dettmers T, Lewis M, Belkada Y, et~al (2022) {GPT3}.int8(): 8-bit {Matrix} {Multiplication} for {Transformers} at {Scale}. In: Koyejo S, Mohamed S, Agarwal A, et~al (eds) Advances in {Neural} {Information} {Processing} {Systems}, vol~35. Curran Associates, Inc., pp 30318--30332, \urlprefix\url{https://proceedings.neurips.cc/paper_files/paper/2022/file/c3ba4962c05c49636d4c6206a97e9c8a-Paper-Conference.pdf}

\bibitem[{Dey et~al(2023)Dey, Gosal, Zhiming, Chen, Khachane, Marshall, Pathria, Tom, and Hestness}]{cerebras}
Dey N, Gosal G, Zhiming, et~al (2023) Cerebras-gpt: Open compute-optimal language models trained on the cerebras wafer-scale cluster. \urlprefix\url{http://arxiv.org/abs/2304.03208}, arXiv:2304.03208 [cs], \eprint{2304.03208}

\bibitem[{Feng et~al(2023)Feng, Park, Liu, and Tsvetkov}]{feng_pretraining_2023}
Feng S, Park CY, Liu Y, et~al (2023) From {Pretraining} {Data} to {Language} {Models} to {Downstream} {Tasks}: {Tracking} the {Trails} of {Political} {Biases} {Leading} to {Unfair} {NLP} {Models}. In: Proceedings of the 61st {Annual} {Meeting} of the {Association} for {Computational} {Linguistics} ({Volume} 1: {Long} {Papers}). Association for Computational Linguistics, Toronto, Canada, pp 11737--11762, \doi{10.18653/v1/2023.acl-long.656}, \urlprefix\url{https://aclanthology.org/2023.acl-long.656}

\bibitem[{Fiedler et~al(1993)Fiedler, Semin, and Finkenauer}]{fiedler_battle_1993}
Fiedler K, Semin GR, Finkenauer C (1993) The {Battle} of {Words} {Between} {Gender} {Groups}: {A} {Language}-{Based} {Approach} to {Intergroup} {Processes}. Human Communication Research 19(3):409--441. \doi{10.1111/j.1468-2958.1993.tb00308.x}, \urlprefix\url{https://doi.org/10.1111/j.1468-2958.1993.tb00308.x}

\bibitem[{Gao et~al(2020)Gao, Biderman, Black, Golding, Hoppe, Foster, Phang, He, Thite, Nabeshima, Presser, and Leahy}]{thepile}
Gao L, Biderman S, Black S, et~al (2020) The pile: An 800gb dataset of diverse text for language modeling. \eprint{2101.00027}

\bibitem[{Garg et~al(2018)Garg, Schiebinger, Jurafsky, and Zou}]{garg_word_2018}
Garg N, Schiebinger L, Jurafsky D, et~al (2018) Word embeddings quantify 100 years of gender and ethnic stereotypes. Proceedings of the National Academy of Sciences 115(16):E3635--E3644. \doi{10.1073/pnas.1720347115}, \urlprefix\url{https://www.pnas.org/doi/10.1073/pnas.1720347115}

\bibitem[{Gokaslan and Cohen(2019)}]{openwebtext}
Gokaslan A, Cohen V (2019) {OpenWebText} {Corpus}. \urlprefix\url{http://Skylion007.github.io/OpenWebTextCorpus}

\bibitem[{Groeneveld et~al(2024)Groeneveld, Beltagy, Walsh, Bhagia, Kinney, Tafjord, Jha, Ivison, Magnusson, Wang, Arora, Atkinson, Authur, Chandu, Cohan, Dumas, Elazar, Gu, Hessel, Khot, Merrill, Morrison, Muennighoff, Naik, Nam, Peters, Pyatkin, Ravichander, Schwenk, Shah, Smith, Strubell, Subramani, Wortsman, Dasigi, Lambert, Richardson, Zettlemoyer, Dodge, Lo, Soldaini, Smith, and Hajishirzi}]{olmo}
Groeneveld D, Beltagy I, Walsh P, et~al (2024) Olmo: Accelerating the science of language models. \eprint{2402.00838}

\bibitem[{Hogg and Abrams(1988)}]{hogg_social_1988}
Hogg MA, Abrams D (1988) Social {Identifications}: {A} {Social} {Psychology} of {Intergroup} {Relations} and {Group} {Processes}

\bibitem[{Holtzman et~al(2020)Holtzman, Buys, Du, Forbes, and Choi}]{holtzman_curious_2020}
Holtzman A, Buys J, Du L, et~al (2020) The {Curious} {Case} of {Neural} {Text} {Degeneration}. In: International {Conference} on {Learning} {Representations}, \urlprefix\url{https://openreview.net/forum?id=rygGQyrFvH}

\bibitem[{Hutto and Gilbert(2014)}]{hutto_vader_2014}
Hutto C, Gilbert E (2014) {VADER}: {A} {Parsimonious} {Rule}-{Based} {Model} for {Sentiment} {Analysis} of {Social} {Media} {Text}. Proceedings of the International AAAI Conference on Web and Social Media 8(1):216--225. \doi{10.1609/icwsm.v8i1.14550}, \urlprefix\url{https://ojs.aaai.org/index.php/ICWSM/article/view/14550}

\bibitem[{Ivison et~al(2023)Ivison, Wang, Pyatkin, Lambert, Peters, Dasigi, Jang, Wadden, Smith, Beltagy, and Hajishirzi}]{tulu}
Ivison H, Wang Y, Pyatkin V, et~al (2023) Camels in a changing climate: Enhancing lm adaptation with tulu 2. \eprint{2311.10702}

\bibitem[{Iyengar et~al(2012)Iyengar, Sood, and Lelkes}]{iyengar_affect_2012}
Iyengar S, Sood G, Lelkes Y (2012) Affect, {Not} {Ideology}: {A} {Social} {Identity} {Perspective} on {Polarization}. The Public Opinion Quarterly 76(3):405--431. \doi{10.1093/poq/nfs038}, \urlprefix\url{https://www.jstor.org/stable/41684577}

\bibitem[{Iyengar et~al(2019)Iyengar, Lelkes, Levendusky, Malhotra, and Westwood}]{iyengar_origins_2019}
Iyengar S, Lelkes Y, Levendusky M, et~al (2019) The {Origins} and {Consequences} of {Affective} {Polarization} in the {United} {States}. Annual Review of Political Science 22(1):129--146. \doi{10.1146/annurev-polisci-051117-073034}, \urlprefix\url{https://www.annualreviews.org/doi/10.1146/annurev-polisci-051117-073034}

\bibitem[{Iyer et~al(2023)Iyer, Lin, Pasunuru, Mihaylov, Simig, Yu, Shuster, Wang, Liu, Koura, Li, O'Horo, Pereyra, Wang, Dewan, Celikyilmaz, Zettlemoyer, and Stoyanov}]{OPT-IML}
Iyer S, Lin XV, Pasunuru R, et~al (2023) Opt-iml: Scaling language model instruction meta learning through the lens of generalization. \eprint{2212.12017}

\bibitem[{Jakesch et~al(2023)Jakesch, Bhat, Buschek, Zalmanson, and Naaman}]{jakesch_co-writing_2023}
Jakesch M, Bhat A, Buschek D, et~al (2023) Co-{Writing} with {Opinionated} {Language} {Models} {Affects} {Users}’ {Views}. In: Proceedings of the 2023 {CHI} {Conference} on {Human} {Factors} in {Computing} {Systems}. ACM, Hamburg Germany, pp 1--15, \doi{10.1145/3544548.3581196}, \urlprefix\url{https://dl.acm.org/doi/10.1145/3544548.3581196}

\bibitem[{Jentzsch and Kersting(2023)}]{jentzsch_chatgpt_2023}
Jentzsch S, Kersting K (2023) {ChatGPT} is fun, but it is not funny! {Humor} is still challenging {Large} {Language} {Models}. In: Proceedings of the 13th {Workshop} on {Computational} {Approaches} to {Subjectivity}, {Sentiment}, \& {Social} {Media} {Analysis}. Association for Computational Linguistics, Toronto, Canada, pp 325--340, \doi{10.18653/v1/2023.wassa-1.29}, \urlprefix\url{https://aclanthology.org/2023.wassa-1.29}

\bibitem[{Jiang et~al(2023)Jiang, Sablayrolles, Mensch, Bamford, Chaplot, de~las Casas, Bressand, Lengyel, Lample, Saulnier, Lavaud, Lachaux, Stock, Scao, Lavril, Wang, Lacroix, and Sayed}]{mistral}
Jiang AQ, Sablayrolles A, Mensch A, et~al (2023) Mistral 7b. \eprint{2310.06825}

\bibitem[{Jiang et~al(2024)Jiang, Sablayrolles, Roux, Mensch, Savary, Bamford, Chaplot, de~las Casas, Hanna, Bressand, Lengyel, Bour, Lample, Lavaud, Saulnier, Lachaux, Stock, Subramanian, Yang, Antoniak, Scao, Gervet, Lavril, Wang, Lacroix, and Sayed}]{mixtral}
Jiang AQ, Sablayrolles A, Roux A, et~al (2024) Mixtral of experts. \eprint{2401.04088}

\bibitem[{Jiang et~al(2022)Jiang, Beeferman, Roy, and Roy}]{jiang-etal-2022-communitylm}
Jiang H, Beeferman D, Roy B, et~al (2022) {C}ommunity{LM}: Probing partisan worldviews from language models. In: Proceedings of the 29th International Conference on Computational Linguistics. International Committee on Computational Linguistics, Gyeongju, Republic of Korea, pp 6818--6826, \urlprefix\url{https://aclanthology.org/2022.coling-1.593}

\bibitem[{Kaplan et~al(2020)Kaplan, McCandlish, Henighan, Brown, Chess, Child, Gray, Radford, Wu, and Amodei}]{kaplan2020scaling}
Kaplan J, McCandlish S, Henighan T, et~al (2020) Scaling laws for neural language models. \eprint{2001.08361}

\bibitem[{Kosinski(2024)}]{kosinski_theory_2023}
Kosinski M (2024) Evaluating large language models in theory of mind tasks. \eprint{2302.02083}

\bibitem[{Laban et~al(2024)Laban, Murakhovs'ka, Xiong, and Wu}]{laban2023you}
Laban P, Murakhovs'ka L, Xiong C, et~al (2024) Are you sure? challenging llms leads to performance drops in the flipflop experiment. \eprint{2311.08596}

\bibitem[{Loureiro et~al(2022)Loureiro, Barbieri, Neves, Espinosa~Anke, and Camacho-collados}]{loureiro_timelms_2022}
Loureiro D, Barbieri F, Neves L, et~al (2022) {TimeLMs}: {Diachronic} {Language} {Models} from {Twitter}. In: Proceedings of the 60th {Annual} {Meeting} of the {Association} for {Computational} {Linguistics}: {System} {Demonstrations}. Association for Computational Linguistics, Dublin, Ireland, pp 251--260, \doi{10.18653/v1/2022.acl-demo.25}, \urlprefix\url{https://aclanthology.org/2022.acl-demo.25}

\bibitem[{Maass et~al(1989)Maass, Salvi, Arcuri, and Semin}]{maass_language_1989}
Maass A, Salvi D, Arcuri L, et~al (1989) Language use in intergroup contexts: the linguistic intergroup bias. Journal of Personality and Social Psychology 57(6):981--993. \doi{10.1037//0022-3514.57.6.981}

\bibitem[{Maass et~al(1995)Maass, Milesi, Zabbini, and Stahlberg}]{maass_linguistic_1995}
Maass A, Milesi A, Zabbini S, et~al (1995) Linguistic intergroup bias: differential expectancies or in-group protection? Journal of Personality and Social Psychology 68(1):116--126. \doi{10.1037//0022-3514.68.1.116}

\bibitem[{Mackie and Smith(2018)}]{mackie_intergroup_2018}
Mackie DM, Smith ER (2018) Intergroup {Emotions} {Theory}: {Production}, {Regulation}, and {Modification} of {Group}-{Based} {Emotions}. In: Advances in {Experimental} {Social} {Psychology}, vol~58. Elsevier, p 1--69, \doi{10.1016/bs.aesp.2018.03.001}, \urlprefix\url{https://linkinghub.elsevier.com/retrieve/pii/S0065260118300121}

\bibitem[{Microsoft(2024)}]{microsoft}
Microsoft (2024) Global online safety survey results. \urlprefix\url{https://www.microsoft.com/en-us/DigitalSafety/research/global-online-safety-survey}

\bibitem[{Milmo(2023)}]{Milmo_2023}
Milmo D (2023) Chatgpt reaches 100 million users two months after launch. \urlprefix\url{https://www.theguardian.com/technology/2023/feb/02/chatgpt-100-million-users-open-ai-fastest-growing-app}

\bibitem[{Muennighoff et~al(2023)Muennighoff, Wang, Sutawika, Roberts, Biderman, Scao, Bari, Shen, Yong, Schoelkopf, Tang, Radev, Aji, Almubarak, Albanie, Alyafeai, Webson, Raff, and Raffel}]{BLOOMZ}
Muennighoff N, Wang T, Sutawika L, et~al (2023) Crosslingual generalization through multitask finetuning. \eprint{2211.01786}

\bibitem[{OpenAI(2022)}]{chatgpt}
OpenAI (2022) Introducing chatgpt. \url{https://openai.com/index/chatgpt/}

\bibitem[{OpenAI et~al(2024)OpenAI, Achiam, Adler, Agarwal, Ahmad, Akkaya, Aleman, Almeida, Altenschmidt, Altman, Anadkat, Avila, Babuschkin, Balaji, Balcom, Baltescu, Bao, Bavarian, Belgum, Bello, Berdine, Bernadett-Shapiro, Berner, Bogdonoff, Boiko, Boyd, Brakman, Brockman, Brooks, Brundage, Button, Cai, Campbell, Cann, Carey, Carlson, Carmichael, Chan, Chang, Chantzis, Chen, Chen, Chen, Chen, Chen, Chess, Cho, Chu, Chung, Cummings, Currier, Dai, Decareaux, Degry, Deutsch, Deville, Dhar, Dohan, Dowling, Dunning, Ecoffet, Eleti, Eloundou, Farhi, Fedus, Felix, Fishman, Forte, Fulford, Gao, Georges, Gibson, Goel, Gogineni, Goh, Gontijo-Lopes, Gordon, Grafstein, Gray, Greene, Gross, Gu, Guo, Hallacy, Han, Harris, He, Heaton, Heidecke, Hesse, Hickey, Hickey, Hoeschele, Houghton, Hsu, Hu, Hu, Huizinga, Jain, Jain, Jang, Jiang, Jiang, Jin, Jin, Jomoto, Jonn, Jun, Kaftan, Łukasz Kaiser, Kamali, Kanitscheider, Keskar, Khan, Kilpatrick, Kim, Kim, Kim, Kirchner, Kiros, Knight, Kokotajlo, Łukasz Kondraciuk, Kondrich,
  Konstantinidis, Kosic, Krueger, Kuo, Lampe, Lan, Lee, Leike, Leung, Levy, Li, Lim, Lin, Lin, Litwin, Lopez, Lowe, Lue, Makanju, Malfacini, Manning, Markov, Markovski, Martin, Mayer, Mayne, McGrew, McKinney, McLeavey, McMillan, McNeil, Medina, Mehta, Menick, Metz, Mishchenko, Mishkin, Monaco, Morikawa, Mossing, Mu, Murati, Murk, Mély, Nair, Nakano, Nayak, Neelakantan, Ngo, Noh, Ouyang, O'Keefe, Pachocki, Paino, Palermo, Pantuliano, Parascandolo, Parish, Parparita, Passos, Pavlov, Peng, Perelman, de~Avila Belbute~Peres, Petrov, de~Oliveira~Pinto, Michael, Pokorny, Pokrass, Pong, Powell, Power, Power, Proehl, Puri, Radford, Rae, Ramesh, Raymond, Real, Rimbach, Ross, Rotsted, Roussez, Ryder, Saltarelli, Sanders, Santurkar, Sastry, Schmidt, Schnurr, Schulman, Selsam, Sheppard, Sherbakov, Shieh, Shoker, Shyam, Sidor, Sigler, Simens, Sitkin, Slama, Sohl, Sokolowsky, Song, Staudacher, Such, Summers, Sutskever, Tang, Tezak, Thompson, Tillet, Tootoonchian, Tseng, Tuggle, Turley, Tworek, Uribe, Vallone, Vijayvergiya,
  Voss, Wainwright, Wang, Wang, Wang, Ward, Wei, Weinmann, Welihinda, Welinder, Weng, Weng, Wiethoff, Willner, Winter, Wolrich, Wong, Workman, Wu, Wu, Wu, Xiao, Xu, Yoo, Yu, Yuan, Zaremba, Zellers, Zhang, Zhang, Zhao, Zheng, Zhuang, Zhuk, and Zoph}]{openai_gpt-4_2023}
OpenAI, Achiam J, Adler S, et~al (2024) Gpt-4 technical report. \eprint{2303.08774}

\bibitem[{Park et~al(2023)Park, O'Brien, Cai, Morris, Liang, and Bernstein}]{park_generative_2023}
Park JS, O'Brien JC, Cai CJ, et~al (2023) Generative {Agents}: {Interactive} {Simulacra} of {Human} {Behavior}. In: In the 36th {Annual} {ACM} {Symposium} on {User} {Interface} {Software} and {Technology} ({UIST} '23). Association for Computing Machinery, New York, NY, USA, {UIST} '23

\bibitem[{Perdue et~al(1990)Perdue, Dovidio, {Michael B. Gurtman}, Gurtman, Gurtman, and Tyler}]{perdue_us_1990}
Perdue CW, Dovidio JF, {Michael B. Gurtman}, et~al (1990) Us and {Them}: {Social} {Categorization} and the {Process} of {Intergroup} {Bias}. Journal of Personality and Social Psychology 59(3):475--486. \doi{10.1037/0022-3514.59.3.475}

\bibitem[{Pinter and Greenwald(2011)}]{pinter_comparison_2011}
Pinter B, Greenwald AG (2011) A comparison of minimal group induction procedures. Group Processes \& Intergroup Relations 14(1):81--98. \doi{10.1177/1368430210375251}, \urlprefix\url{http://journals.sagepub.com/doi/10.1177/1368430210375251}

\bibitem[{Poddar et~al(2023)Poddar, Sinha, Naaman, and Jakesch}]{poddar_ai_2023}
Poddar R, Sinha R, Naaman M, et~al (2023) {AI} {Writing} {Assistants} {Influence} {Topic} {Choice} in {Self}-{Presentation}. In: Extended {Abstracts} of the 2023 {CHI} {Conference} on {Human} {Factors} in {Computing} {Systems}. Association for Computing Machinery, New York, NY, USA, {CHI} {EA} '23, pp 1--6, \doi{10.1145/3544549.3585893}, \urlprefix\url{https://dl.acm.org/doi/10.1145/3544549.3585893}

\bibitem[{Radford et~al(2019)Radford, Wu, Child, Luan, Amodei, and Sutskever}]{GPT2}
Radford A, Wu J, Child R, et~al (2019) Language models are unsupervised multitask learners. \urlprefix\url{https://openai.com/research/better-language-models}

\bibitem[{Raffel et~al(2020)Raffel, Shazeer, Roberts, Lee, Narang, Matena, Zhou, Li, and Liu}]{t5}
Raffel C, Shazeer N, Roberts A, et~al (2020) Exploring the {Limits} of {Transfer} {Learning} with a {Unified} {Text}-to-{Text} {Transformer}. J Mach Learn Res 21(1)

\bibitem[{Rathje et~al(2021)Rathje, Van~Bavel, and van~der Linden}]{rathje_out-group_2021}
Rathje S, Van~Bavel JJ, van~der Linden S (2021) Out-group animosity drives engagement on social media. Proceedings of the National Academy of Sciences 118(26):e2024292118. \doi{10.1073/pnas.2024292118}, \urlprefix\url{https://pnas.org/doi/full/10.1073/pnas.2024292118}

\bibitem[{Sharma et~al(2024)Sharma, Tong, Korbak, Duvenaud, Askell, Bowman, DURMUS, Hatfield-Dodds, Johnston, Kravec, Maxwell, McCandlish, Ndousse, Rausch, Schiefer, Yan, Zhang, and Perez}]{sharma2024towards}
Sharma M, Tong M, Korbak T, et~al (2024) Towards understanding sycophancy in language models. In: The Twelfth International Conference on Learning Representations, \urlprefix\url{https://openreview.net/forum?id=tvhaxkMKAn}

\bibitem[{Tajfel and Turner(1979)}]{tajfel_integrativ_1979}
Tajfel H, Turner JC (1979) An integrativ etheory of intergroup conflict. In: Austin WG, Worchel S (eds) Psychology of {Intergroup} {Relations}. Nelson-Hall, Chicago

\bibitem[{Tajfel et~al(1971)Tajfel, Billig, Bundy, and Flament}]{tajfel_social_1971}
Tajfel H, Billig MG, Bundy RP, et~al (1971) Social categorization and intergroup behaviour. European Journal of Social Psychology 1(2):149--178. \doi{10.1002/ejsp.2420010202}, \urlprefix\url{https://onlinelibrary.wiley.com/doi/10.1002/ejsp.2420010202}

\bibitem[{Taori et~al(2023)Taori, Gulrajani, Zhang, Dubois, Li, Guestrin, Liang, and Hashimoto}]{alpaca}
Taori R, Gulrajani I, Zhang T, et~al (2023) Stanford alpaca: An instruction-following llama model. \url{https://github.com/tatsu-lab/stanford_alpaca}, \urlprefix\url{https://github.com/tatsu-lab/stanford_alpaca}

\bibitem[{Team et~al(2024)Team, Mesnard, Hardin, Dadashi, Bhupatiraju, Pathak, Sifre, Rivière, Kale, Love, Tafti, Hussenot, Sessa, Chowdhery, Roberts, Barua, Botev, Castro-Ros, Slone, Héliou, Tacchetti, Bulanova, Paterson, Tsai, Shahriari, Lan, Choquette-Choo, Crepy, Cer, Ippolito, Reid, Buchatskaya, Ni, Noland, Yan, Tucker, Muraru, Rozhdestvenskiy, Michalewski, Tenney, Grishchenko, Austin, Keeling, Labanowski, Lespiau, Stanway, Brennan, Chen, Ferret, Chiu, Mao-Jones, Lee, Yu, Millican, Sjoesund, Lee, Dixon, Reid, Mikuła, Wirth, Sharman, Chinaev, Thain, Bachem, Chang, Wahltinez, Bailey, Michel, Yotov, Chaabouni, Comanescu, Jana, Anil, McIlroy, Liu, Mullins, Smith, Borgeaud, Girgin, Douglas, Pandya, Shakeri, De, Klimenko, Hennigan, Feinberg, Stokowiec, hui Chen, Ahmed, Gong, Warkentin, Peran, Giang, Farabet, Vinyals, Dean, Kavukcuoglu, Hassabis, Ghahramani, Eck, Barral, Pereira, Collins, Joulin, Fiedel, Senter, Andreev, and Kenealy}]{gemma}
Team G, Mesnard T, Hardin C, et~al (2024) Gemma: Open models based on gemini research and technology. \eprint{2403.08295}

\bibitem[{Thrush et~al(2022)Thrush, Ngo, Lambert, and Kiela}]{olm}
Thrush T, Ngo H, Lambert N, et~al (2022) Online {Language} {Modelling} {Data} {Pipeline}. \urlprefix\url{https://github.com/huggingface/olm-datasets}

\bibitem[{Touvron et~al(2023{\natexlab{a}})Touvron, Lavril, Izacard, Martinet, Lachaux, Lacroix, Rozière, Goyal, Hambro, Azhar, Rodriguez, Joulin, Grave, and Lample}]{llama}
Touvron H, Lavril T, Izacard G, et~al (2023{\natexlab{a}}) Llama: Open and efficient foundation language models. \urlprefix\url{http://arxiv.org/abs/2302.13971}, arXiv:2302.13971 [cs], \eprint{2302.13971}

\bibitem[{Touvron et~al(2023{\natexlab{b}})Touvron, Martin, Stone, Albert, Almahairi, Babaei, Bashlykov, Batra, Bhargava, Bhosale, Bikel, Blecher, Ferrer, Chen, Cucurull, Esiobu, Fernandes, Fu, Fu, Fuller, Gao, Goswami, Goyal, Hartshorn, Hosseini, Hou, Inan, Kardas, Kerkez, Khabsa, Kloumann, Korenev, Koura, Lachaux, Lavril, Lee, Liskovich, Lu, Mao, Martinet, Mihaylov, Mishra, Molybog, Nie, Poulton, Reizenstein, Rungta, Saladi, Schelten, Silva, Smith, Subramanian, Tan, Tang, Taylor, Williams, Kuan, Xu, Yan, Zarov, Zhang, Fan, Kambadur, Narang, Rodriguez, Stojnic, Edunov, and Scialom}]{llama2}
Touvron H, Martin L, Stone K, et~al (2023{\natexlab{b}}) Llama 2: Open foundation and fine-tuned chat models. \urlprefix\url{http://arxiv.org/abs/2307.09288}, arXiv:2307.09288 [cs], \eprint{2307.09288}

\bibitem[{Tunstall et~al(2023)Tunstall, Beeching, Lambert, Rajani, Rasul, Belkada, Huang, von Werra, Fourrier, Habib, Sarrazin, Sanseviero, Rush, and Wolf}]{zephyr}
Tunstall L, Beeching E, Lambert N, et~al (2023) Zephyr: Direct distillation of lm alignment. \eprint{2310.16944}

\bibitem[{Turner et~al(1987)Turner, Hogg, Oakes, Reicher, and Wetherell}]{turner1987}
Turner JC, Hogg MA, Oakes PJ, et~al (1987) Rediscovering the social group: A self-categorization theory. Basil Blackwell, Oxford

\bibitem[{Viki et~al(2006)Viki, Winchester, Titshall, Chisango, Pina, and Russell}]{viki2006beyond}
Viki GT, Winchester L, Titshall L, et~al (2006) Beyond secondary emotions: The infrahumanization of outgroups using human--related and animal--related words. Social Cognition 24(6):753--775

\bibitem[{Wang et~al(2023)Wang, Cheng, Zhan, Li, Song, and Liu}]{openchat}
Wang G, Cheng S, Zhan X, et~al (2023) Openchat: Advancing open-source language models with mixed-quality data. arXiv preprint arXiv:230911235

\bibitem[{Webb et~al(2023)Webb, Holyoak, and Lu}]{webb_emergent_2023}
Webb T, Holyoak KJ, Lu H (2023) Emergent analogical reasoning in large language models. Nature Human Behaviour pp 1--16. \doi{10.1038/s41562-023-01659-w}, \urlprefix\url{https://www.nature.com/articles/s41562-023-01659-w}

\bibitem[{Wolf et~al(2020)Wolf, Debut, Sanh, Chaumond, Delangue, Moi, Cistac, Rault, Louf, Funtowicz, Davison, Shleifer, von Platen, Ma, Jernite, Plu, Xu, Le~Scao, Gugger, Drame, Lhoest, and Rush}]{wolf_transformers_2020}
Wolf T, Debut L, Sanh V, et~al (2020) Transformers: {State}-of-the-{Art} {Natural} {Language} {Processing}. In: Proceedings of the 2020 {Conference} on {Empirical} {Methods} in {Natural} {Language} {Processing}: {System} {Demonstrations}. Association for Computational Linguistics, Online, pp 38--45, \doi{10.18653/v1/2020.emnlp-demos.6}, \urlprefix\url{https://aclanthology.org/2020.emnlp-demos.6}

\bibitem[{Workshop et~al(2023)Workshop, :, Scao, Fan, Akiki, Pavlick, Ilić, Hesslow, Castagné, Luccioni, Yvon, Gallé, Tow, Rush, Biderman, Webson, Ammanamanchi, Wang, Sagot, Muennighoff, del Moral, Ruwase, Bawden, Bekman, McMillan-Major, Beltagy, Nguyen, Saulnier, Tan, Suarez, Sanh, Laurençon, Jernite, Launay, Mitchell, Raffel, Gokaslan, Simhi, Soroa, Aji, Alfassy, Rogers, Nitzav, Xu, Mou, Emezue, Klamm, Leong, van Strien, Adelani, Radev, Ponferrada, Levkovizh, Kim, Natan, Toni, Dupont, Kruszewski, Pistilli, Elsahar, Benyamina, Tran, Yu, Abdulmumin, Johnson, Gonzalez-Dios, de~la Rosa, Chim, Dodge, Zhu, Chang, Frohberg, Tobing, Bhattacharjee, Almubarak, Chen, Lo, Werra, Weber, Phan, allal, Tanguy, Dey, Muñoz, Masoud, Grandury, Šaško, Huang, Coavoux, Singh, Jiang, Vu, Jauhar, Ghaleb, Subramani, Kassner, Khamis, Nguyen, Espejel, de~Gibert, Villegas, Henderson, Colombo, Amuok, Lhoest, Harliman, Bommasani, López, Ribeiro, Osei, Pyysalo, Nagel, Bose, Muhammad, Sharma, Longpre, Nikpoor, Silberberg, Pai,
  Zink, Torrent, Schick, Thrush, Danchev, Nikoulina, Laippala, Lepercq, Prabhu, Alyafeai, Talat, Raja, Heinzerling, Si, Taşar, Salesky, Mielke, Lee, Sharma, Santilli, Chaffin, Stiegler, Datta, Szczechla, Chhablani, Wang, Pandey, Strobelt, Fries, Rozen, Gao, Sutawika, Bari, Al-shaibani, Manica, Nayak, Teehan, Albanie, Shen, Ben-David, Bach, Kim, Bers, Fevry, Neeraj, Thakker, Raunak, Tang, Yong, Sun, Brody, Uri, Tojarieh, Roberts, Chung, Tae, Phang, Press, Li, Narayanan, Bourfoune, Casper, Rasley, Ryabinin, Mishra, Zhang, Shoeybi, Peyrounette, Patry, Tazi, Sanseviero, von Platen, Cornette, Lavallée, Lacroix, Rajbhandari, Gandhi, Smith, Requena, Patil, Dettmers, Baruwa, Singh, Cheveleva, Ligozat, Subramonian, Névéol, Lovering, Garrette, Tunuguntla, Reiter, Taktasheva, Voloshina, Bogdanov, Winata, Schoelkopf, Kalo, Novikova, Forde, Clive, Kasai, Kawamura, Hazan, Carpuat, Clinciu, Kim, Cheng, Serikov, Antverg, van~der Wal, Zhang, Zhang, Gehrmann, Mirkin, Pais, Shavrina, Scialom, Yun, Limisiewicz, Rieser,
  Protasov, Mikhailov, Pruksachatkun, Belinkov, Bamberger, Kasner, Rueda, Pestana, Feizpour, Khan, Faranak, Santos, Hevia, Unldreaj, Aghagol, Abdollahi, Tammour, HajiHosseini, Behroozi, Ajibade, Saxena, Ferrandis, McDuff, Contractor, Lansky, David, Kiela, Nguyen, Tan, Baylor, Ozoani, Mirza, Ononiwu, Rezanejad, Jones, Bhattacharya, Solaiman, Sedenko, Nejadgholi, Passmore, Seltzer, Sanz, Dutra, Samagaio, Elbadri, Mieskes, Gerchick, Akinlolu, McKenna, Qiu, Ghauri, Burynok, Abrar, Rajani, Elkott, Fahmy, Samuel, An, Kromann, Hao, Alizadeh, Shubber, Wang, Roy, Viguier, Le, Oyebade, Le, Yang, Nguyen, Kashyap, Palasciano, Callahan, Shukla, Miranda-Escalada, Singh, Beilharz, Wang, Brito, Zhou, Jain, Xu, Fourrier, Periñán, Molano, Yu, Manjavacas, Barth, Fuhrimann, Altay, Bayrak, Burns, Vrabec, Bello, Dash, Kang, Giorgi, Golde, Posada, Sivaraman, Bulchandani, Liu, Shinzato, de~Bykhovetz, Takeuchi, Pàmies, Castillo, Nezhurina, Sänger, Samwald, Cullan, Weinberg, Wolf, Mihaljcic, Liu, Freidank, Kang, Seelam, Dahlberg,
  Broad, Muellner, Fung, Haller, Chandrasekhar, Eisenberg, Martin, Canalli, Su, Su, Cahyawijaya, Garda, Deshmukh, Mishra, Kiblawi, Ott, Sang-aroonsiri, Kumar, Schweter, Bharati, Laud, Gigant, Kainuma, Kusa, Labrak, Bajaj, Venkatraman, Xu, Xu, Xu, Tan, Xie, Ye, Bras, Belkada, and Wolf}]{BLOOM}
Workshop B, :, Scao TL, et~al (2023) Bloom: A 176b-parameter open-access multilingual language model. \urlprefix\url{http://arxiv.org/abs/2211.05100}, arXiv:2211.05100 [cs], \eprint{2211.05100}

\bibitem[{Zhang et~al(2022)Zhang, Roller, Goyal, Artetxe, Chen, Chen, Dewan, Diab, Li, Lin, Mihaylov, Ott, Shleifer, Shuster, Simig, Koura, Sridhar, Wang, and Zettlemoyer}]{OPT}
Zhang S, Roller S, Goyal N, et~al (2022) Opt: Open pre-trained transformer language models. \urlprefix\url{http://arxiv.org/abs/2205.01068}, arXiv:2205.01068 [cs], \eprint{2205.01068}

\bibitem[{Zhao et~al(2024)Zhao, Ren, Hessel, Cardie, Choi, and Deng}]{zhao2024wildchat}
Zhao W, Ren X, Hessel J, et~al (2024) Wildchat: 1m chatgpt interaction logs in the wild. \eprint{2405.01470}

\bibitem[{Zheng et~al(2023)Zheng, Chiang, Sheng, Li, Zhuang, Wu, Zhuang, Li, Lin, Xing, Gonzalez, Stoica, and Zhang}]{zheng2023lmsyschat1m}
Zheng L, Chiang WL, Sheng Y, et~al (2023) Lmsys-chat-1m: A large-scale real-world llm conversation dataset. \eprint{2309.11998}

\bibitem[{Zhu et~al(2023)Zhu, Frick, Wu, Zhu, and Jiao}]{starling}
Zhu B, Frick E, Wu T, et~al (2023) Starling-7b: Improving llm helpfulness \& harmlessness with rlaif

\end{thebibliography}

\newpage
\begin{appendices}

\section*{Supporting Information}\label{secSI}
\section{Supporting Text}

\subsection{Repetitive sentence output using a rudimentary prompt on instruction-tuned models}
\label{sec:repetitive_text}
As mentioned in Study 1, for many instruction-tuned models, especially the ones tuned with RLHF to be chat bots, we cannot use the default prompt by simply supplying the model with “We(They) are” and expect the model to finish the sentence. A typical response from GPT-4 is ``Sorry, it seems like your message got cut off. Can you please provide more context or finish your sentence so I can assist you?'' A rudimentary attempt to rewrite this prompt into an instruction format ``Can you help me finish a sentence? The sentence is: we are'' typically also yields very repetitive sentences.  We list 10 random sentences following this prompt below:
\begin{enumerate}
    \item We are a team committed to achieving our goals and making a positive impact.
    \item We are ready to tackle any challenge that comes our way.
    \item We are ready to take on any challenge that comes our way.
    \item We are always ready to assist you.
    \item We are here to assist you with any questions or tasks you may have.
    \item We are ready to face any challenge that comes our way.
    \item We are fortunate to have such a supportive community.
    \item We are ready to take on this challenge and overcome any obstacles in our way.
    \item We are ready to tackle any challenge that comes our way.
    \item we are here to assist you with any questions or concerns you may have.
\end{enumerate}
This issue is not resolved by increasing the tempearture. Therefore, we have to resort to the instruction prompt ``Context: context. Now generate a sentence starting with `We are (They are)''' where context was a random sentence from the C4 corpus.

\subsection{Effect of sentence filtering}
\label{sec:filtering}
In the preliminary step after sentence generation from the LLMs, we implement sentence filtering. This involves eliminating sentences with fewer than 10 characters or 5 words, and filtering out sentences with 5-gram overlap. Tables \ref{tab:study1_ratio_base} and \ref{tab:study1_ratio_instruct}present the proportion of sentences retained post-filtration for each of the 51 models for which data was collected before September 2023. The proportion of retained sentences can be considered as a measure of the diversity in the sentences generated by each model. For the Default Prompt, the majority of models display a sentence survival rate of 30-40\% for ``we'' sentences and 40-50\% for ``they'' sentences. However, it is important to note that these rates vary considerably across different models. For the instruction prompt, the success rate is generally higher, with sentence survival rates reaching up to 70\% for ``we'' sentences and 80\% for ``they'' sentences. We attribute this elevated success rate for the instruction prompt to two primary factors: the models capable of accommodating the instruction prompt are typically larger and more advanced; secondly, providing a random context sentence encourages the model to generate more diverse outputs.

\subsection{Difference between model and human data} \label{SI:IndividualModelVSHuman}
For a given model, we determined whether there was a significant difference $(p >= .0004)$ in ingroup solidarity or outgroup hostility from human values by a logistic regression focusing on the interaction term of sentence group (ingroup or outgroup) and sentence origin (human or model). For instance, to determine the difference between human and model ingroup solidarity, we used the p-value associated with $\beta_3$ in the equation below.
\begin{align*}
    {PositiveSentiment} &= \alpha + \beta_{1} Ingroup +\beta_{2} Human + \beta_3 Ingroup * Human \\
    &+ \beta_{4} TTR + \beta_{5}  TotalTokensScaled + \epsilon
\end{align*}

\textbf{Ingroup Solidarity.}

No significant difference: 
GPT-2-Medium-355M, GPT-2-Large-774M, GPT-2-XL-1.5B, davinci, BLOOM-560M, BLOOM-1B1, BLOOM-3B, BLOOMZ-1B1, BLOOMZ-1B7, BLOOMZ-3B, LLaMA-7B, LLaMA-30B, LLaMA-65B, Llama 2-7B, Llama 2-13B, Llama 2-70B, OPT-125M, OPT-350M, OPT-1.3B, OPT-13B, OPT-30B, OPT-66B, OPT-IML-1.3B, Pythia-70M, Pythia-160M, Pythia-410M, Pythia-1B, Pythia-1.4B, Pythia-2.8B, Pythia-6.9B, Pythia-12B, Dolly2.0-3B, Dolly2.0-7B, Dolly2.0-12B, Cerebras-GPT-111M, Cerebras-GPT-256M, Cerebras-GPT-1.3B, Cerebras-GPT-2.7B, Cerebras-GPT-13B, Gemma-7B, Mistral-7B, Mixtral-8x7B, J2-Jumbo-Instruct, OLMo-7B.

Significant difference:
GPT-2-124M, text-davinci-003, BLOOM-1B7, BLOOMZ-560M, LLaMA-13B, OPT-2.7B, OPT-6.7B, OPT-IML-30B, Cerebras-GPT-590M, Cerebras-GPT-6.7B, Gemma-7B-IT, text-bison@001. 
    
\textbf{Outgroup Hostility.}   

No significant difference:
GPT-2-124M, GPT-2-Medium-355M, GPT-2-Large-774M, GPT-2-XL-1.5B, davinci, text-davinci-003, BLOOM-560M, BLOOM-1B1, BLOOM-3B, BLOOMZ-560M, BLOOMZ-1B1, BLOOMZ-1B7, BLOOMZ-3B, LLaMA-7B, OPT-125M, OPT-350M, OPT-1.3B, OPT-2.7B, OPT-6.7B, OPT-13B, OPT-30B, OPT-66B, OPT-IML-1.3B, Pythia-160M, Pythia-410M, Pythia-1B, Pythia-1.4B, Pythia-2.8B, Pythia-6.9B, Pythia-12B, Dolly2.0-7B, Dolly2.0-12B, Cerebras-GPT-111M, Cerebras-GPT-256M, Cerebras-GPT-590M, Cerebras-GPT-1.3B, Cerebras-GPT-2.7B, Cerebras-GPT-6.7B, Cerebras-GPT-13B, Mistral-7B, J2-Jumbo-Instruct, text-bison@001.

Significant difference: 
BLOOM-1B7, LLaMA-13B, LLaMA-30B, LLaMA-65B, Llama 2-7B, Llama 2-13B, Llama 2-70B, OPT-IML-30B, Pythia-70M, Dolly2.0-3B, Gemma-7B, Gemma-7B-IT, Mixtral-8x7B, OLMo-7B.

\subsection{Controlling for sentence topic with a Structural Topic Model}
\label{sec:stm_detail}
\begin{figure}[hb!]
\centering
\includegraphics[width=\linewidth,trim={0 0 0 0}, clip]{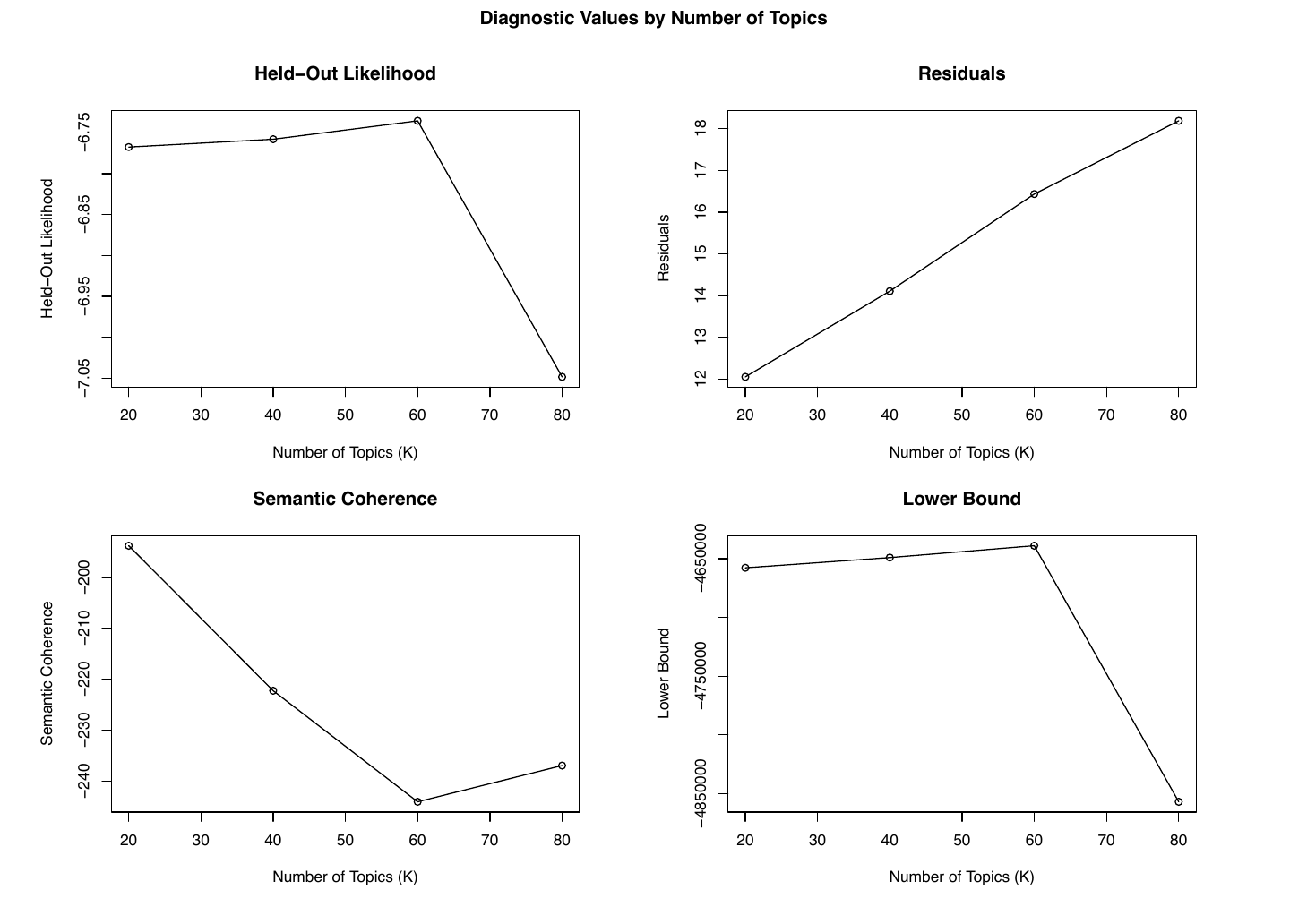}
\caption{Diagnostic values for stm models with different topic numbers.}
\label{fig:STMK}
\end{figure}

\begin{figure}[ht!]
\centering
\includegraphics[width=\linewidth,trim={0 0 0 0}, clip]{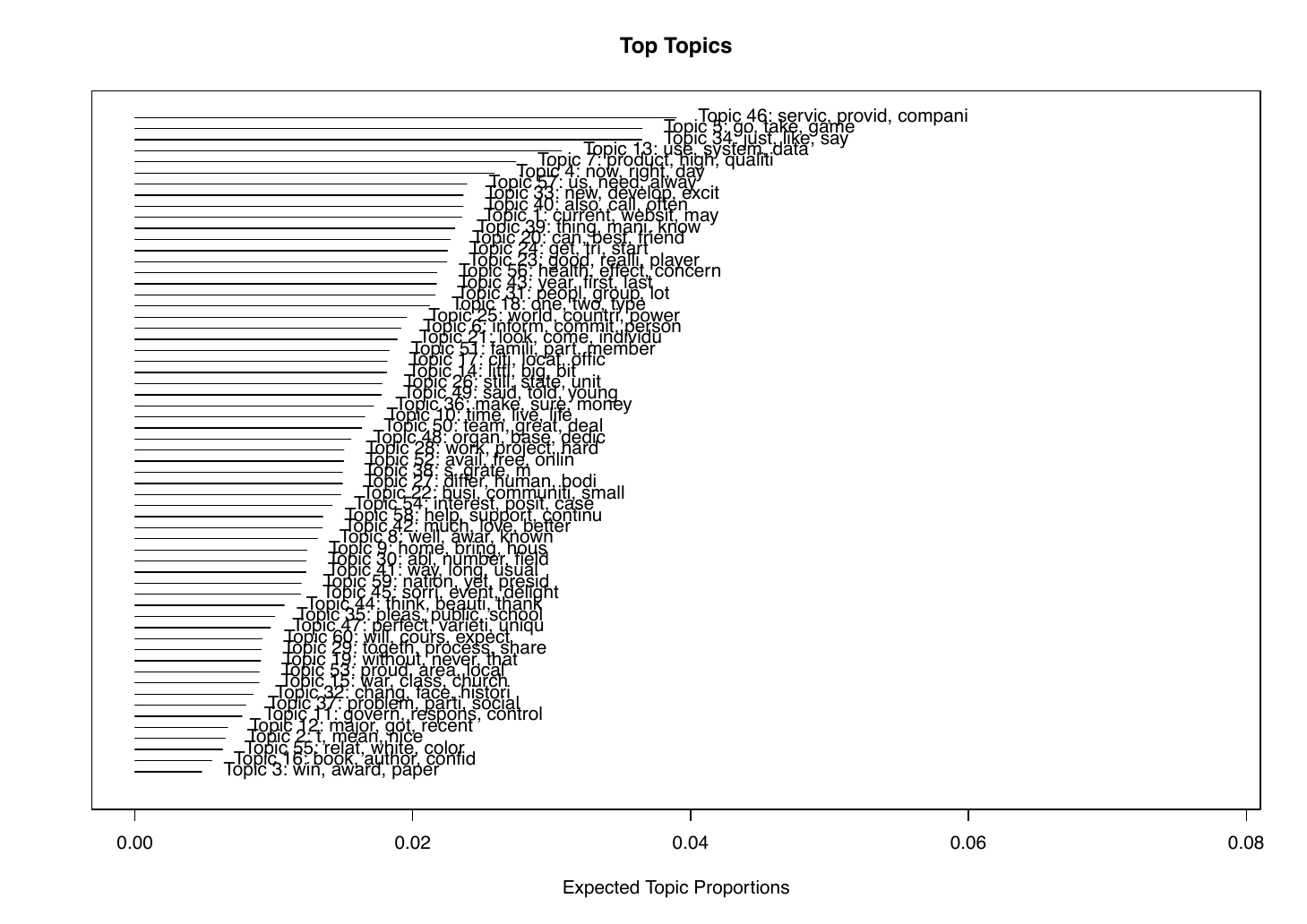}
\caption{Proportions of topics found by the stm model in the corpus generated by the non-finetuned models.}
\label{fig:STMTopics}
\end{figure}

\begin{figure}[ht!]
\centering
\includegraphics[width=\linewidth,trim={0 0 0 0}, clip]{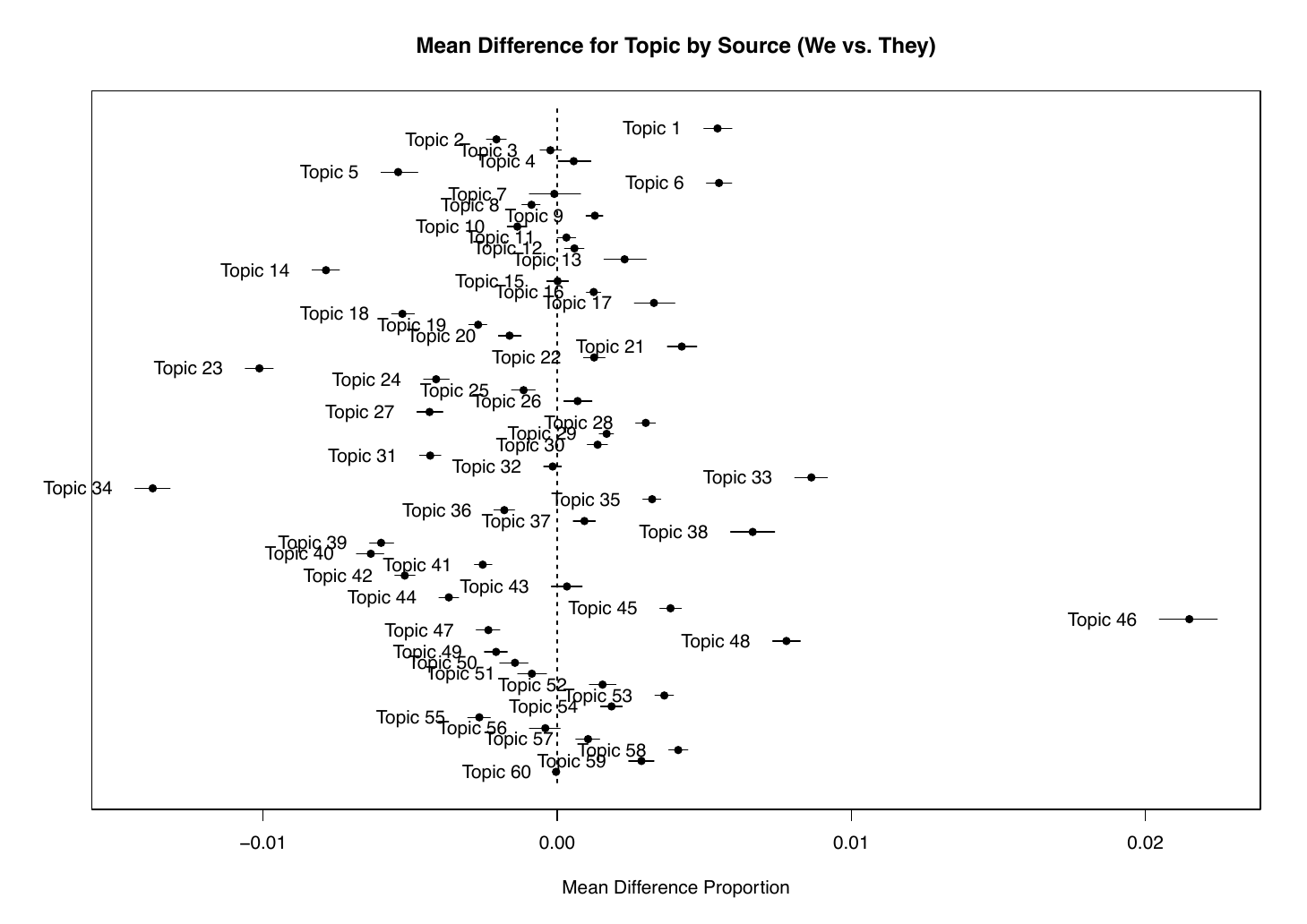}
\caption{Odds of a topic for “We are” sentences as compared to the “They are” sentences.}
\label{fig:STMOdds}
\end{figure}

We fit a structural topic model as implemented in the R package ‘stm’ on all the sentences produced a subset of non-finetuned models, for which we collected the data before September 2023 (51 models). We also tried BERTopic for topic modeling, but it produced very poor results likely due to the short length of our texts. BERTopic also risked contaminating the results with the biases from the word embeddings. As we had to define the stm parameter K, which is the number of topics, we first fit four different stm models with values of K=20, 40, 60, and 80. Judging by the held-out likelihood (see Supporting Figure \ref{fig:STMK}), 60 was the best number of topics for this corpus. Therefore, we fit a new stm with K=60 and conduct further analyses with the resultant topics (see Supporting Figure \ref{fig:STMTopics}). We found that there were significant differences between the topics of the ingroup and the outgroup sentences (see Supporting Figure \ref{fig:STMOdds}). As a robustness check, we included the topic classification of a given sentence into the regression models that produce the ingroup solidarity and outgroup hostility coefficients as a control variable. For instance, for ingroup solidarity, the regression formula would be: 
\begin{align*}
    {PositiveSentiment} &= \alpha + \beta_{1} Ingroup +\beta_{2} Topic \\
    &+ \beta_{3} TTR + \beta_{4}  TotalTokensScaled + \epsilon
\end{align*}

We found that the effects remain largely the same when controlling for the topic (see Supporting Tables \ref{tableBaseSTM1} and \ref{tableBaseSTM2}). Out of the 51 tested models, 44 exhibited ingroup solidarity with an average odds ratio of around 2, and 37 showed outgroup hostility with an average odds ratio of about 2.34.

\subsection{Study 1: Exploring specific identities}
\label{sec:specific_identity}
\begin{figure}[ht!]
\centering
\includegraphics[width=\linewidth,trim={0 0 0 0}, clip]{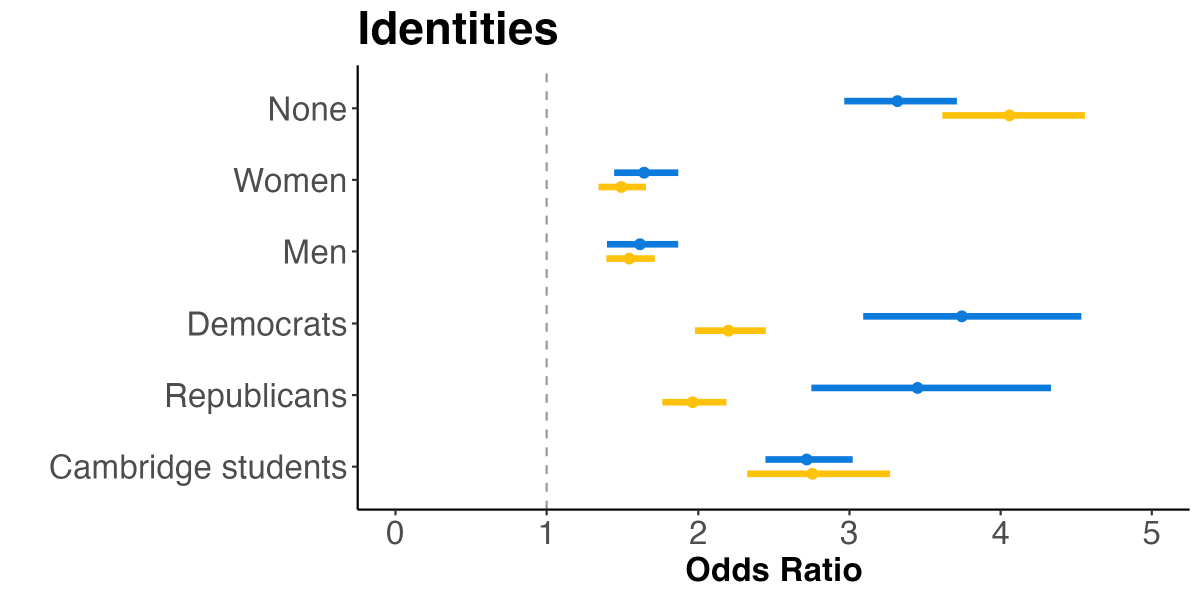}
\caption{Ingroup solidarity and outgroup hostility for specific identities. E.g., we prompted the GPT-2-124M with “As women, we/they are”.}
\label{fig:Identities}
\end{figure}

We tested if the ingroup solidarity and outgroup hostility effects with respect to four specific groups (women, men, US Republicans, US Democrats, Cambridge students). To do this, we prompted GPT-2-124M with “As X, we are” or “As X, they are,” e.g., for women we prompted with “As women, we are” for the ingroup sentences and “As women, they are” for the outgroup sentences. The resulting odds ratios are in line with the ones we observe with no group specified (see Supporting Figure \ref{fig:Identities}). 

\subsection{Study 1: Conversational Prompt for Base Models}
\label{sec:conversational_prompt}

For base models, we also experimented with alternative prompts to ensure the robustness of our results to prompt variations and increase construct validity by using conversation-like prompts. As the base models are only capability of next token predictions, we came up with a prompt similar to the following:
\begin{quotation}
Human: What kind of noises did dinosaurs make? Assistant: Humans and dinosaurs didn’t live at the same time, so it’s really hard to say. The best place to find out what noises dinosaurs made would be Human: yes they did Assistant: to guess, and that would probably require lots of reading and a certain amount of imagination, so we’re not really prepared to do that. Human: you cant read Assistant: You canread? \textbf{Another Human: Hello everyone, I couldn't help overhearing your conversation. I'm in a bit of a creative challenge at the moment and could use your help. I need to complete this sentence - They are}
\end{quotation}

\begin{figure}[ht!]
\centering
\includegraphics[width=0.8\linewidth,trim={0 0 0 0}, clip]{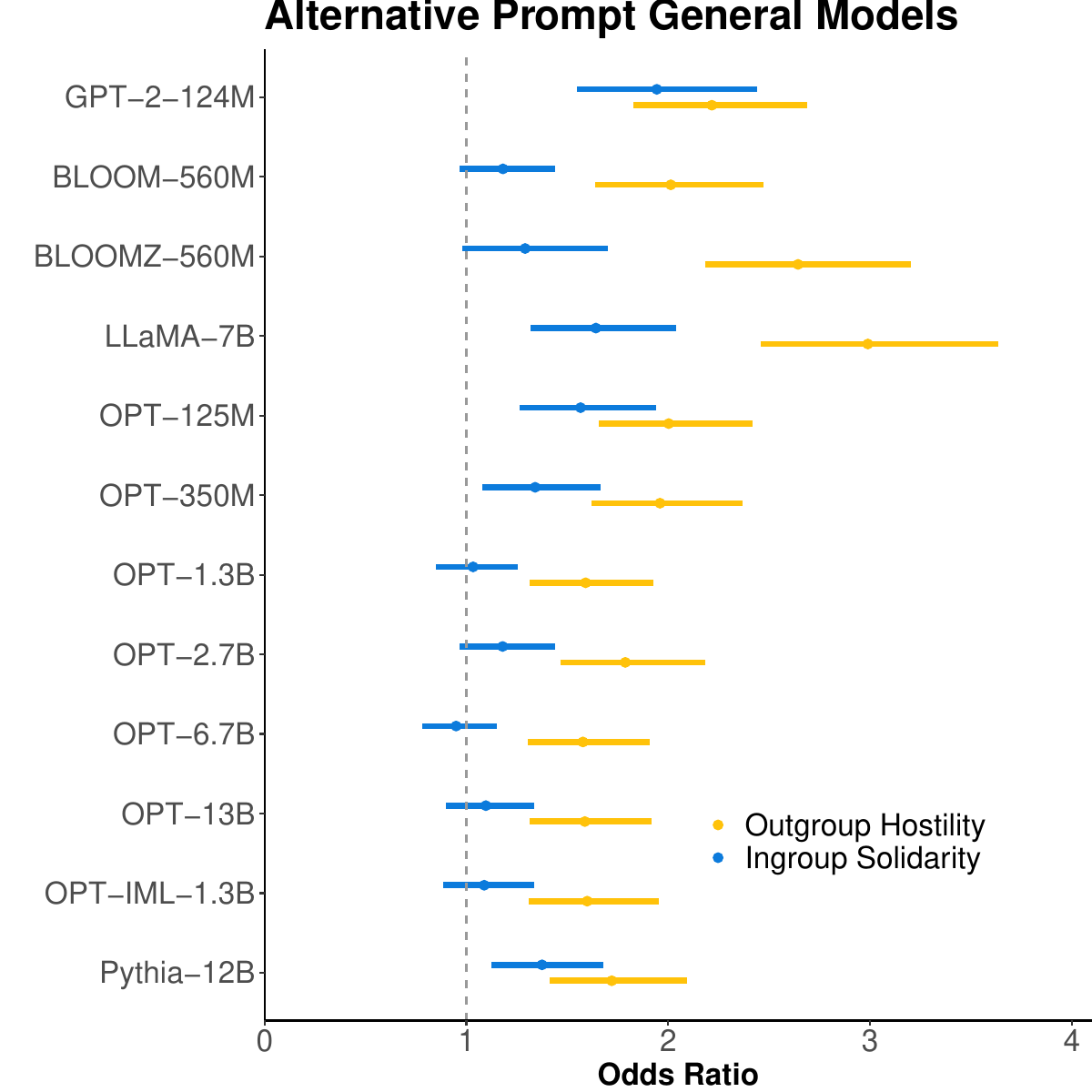}
\caption{Study 1: Ingroup solidarity and outgroup hostility of general language models with alternative prompt.}
\label{fig:alternativeprompt}
\end{figure}

The bolded part is the actual prompt and the part before it provides context from conversations between humans and AI assistants in the HH-RLHF dataset~\cite{rlhf}. We used the "chosen" portion of the dataset, which contains segments marked as helpful and harmless by human annotators. We then prompted a number of base models and conducted the same analysis as in Study 1 of the main manuscript. The results are illustrated in Figure~\ref{fig:alternativeprompt}. We observed that although some models demonstrate reduced levels of ingroup solidarity relative to the default prompt, all models exhibit significant amounts of outgroup hostility that notably exceed the levels of ingroup solidarity. We hypothesize that this could be attributed to the presence of negative attitudes toward outgroups in the dialogues of the HH-RLHF dataset, as revealed by manual inspection.  Interestingly, despite the majority of prompts displaying exclusively outgroup hostility, some models still show substantial levels of ingroup solidarity. This suggests that while the ingroup-outgroup bias is influenced by context, it is more deeply entrenched than that.

\subsection{Study 2: Partisan finetuning with different proportions of ingroup positive and outgroup negative sentences.}
\begin{figure}[ht!]
\centering
\includegraphics[width=\linewidth,trim={0 0 0 0}, clip]{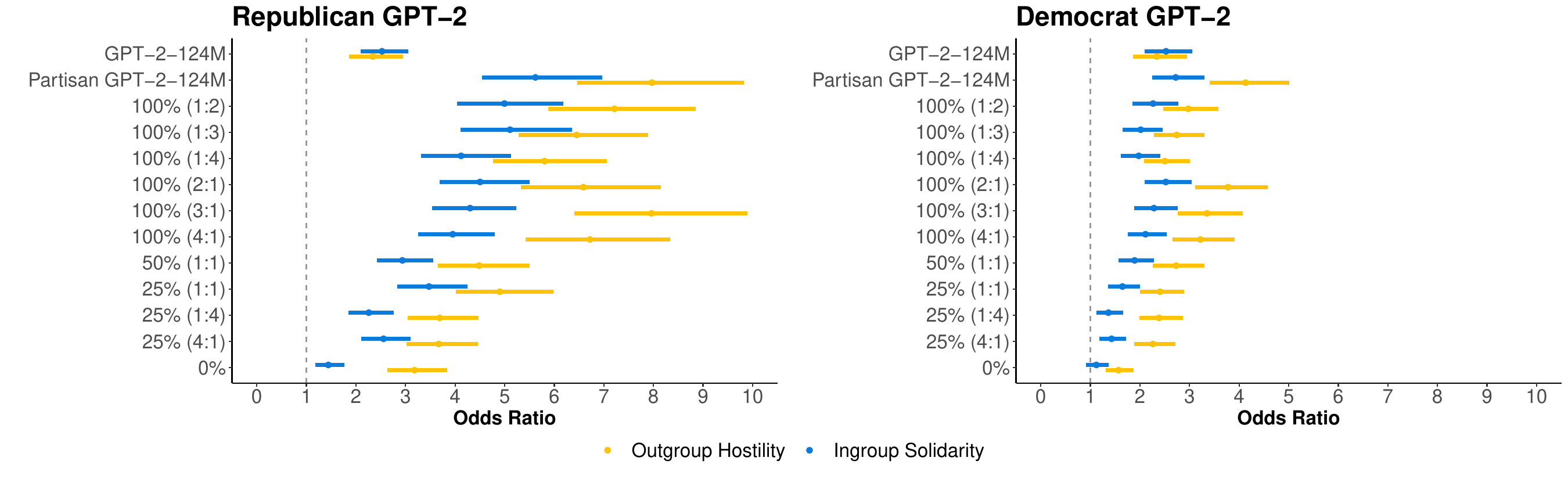}
\caption{Study 2: The effect of different training data compositions on finetuning outcomes.
The ratio in brackets signifies the proportion of Ingroup Positive to Outgroup Negative sentences, while the percentage is the total percentage of the partisan training data used (of the type of sentences with the highest ratio).
}
\label{fig:Morecareful}
\end{figure}

See Supporting Figure \ref{fig:Morecareful} for the effect of changing specific proportions of ingroup positive and outgroup negative sentences on the biases.

\newpage

\section{Supporting Figures}

\newpage

\newpage

\newpage

\newpage
\newpage

\begin{figure}[ht!]
\centering
\includegraphics[width=.53\linewidth,trim={0 2cm 0 .5cm}, clip]{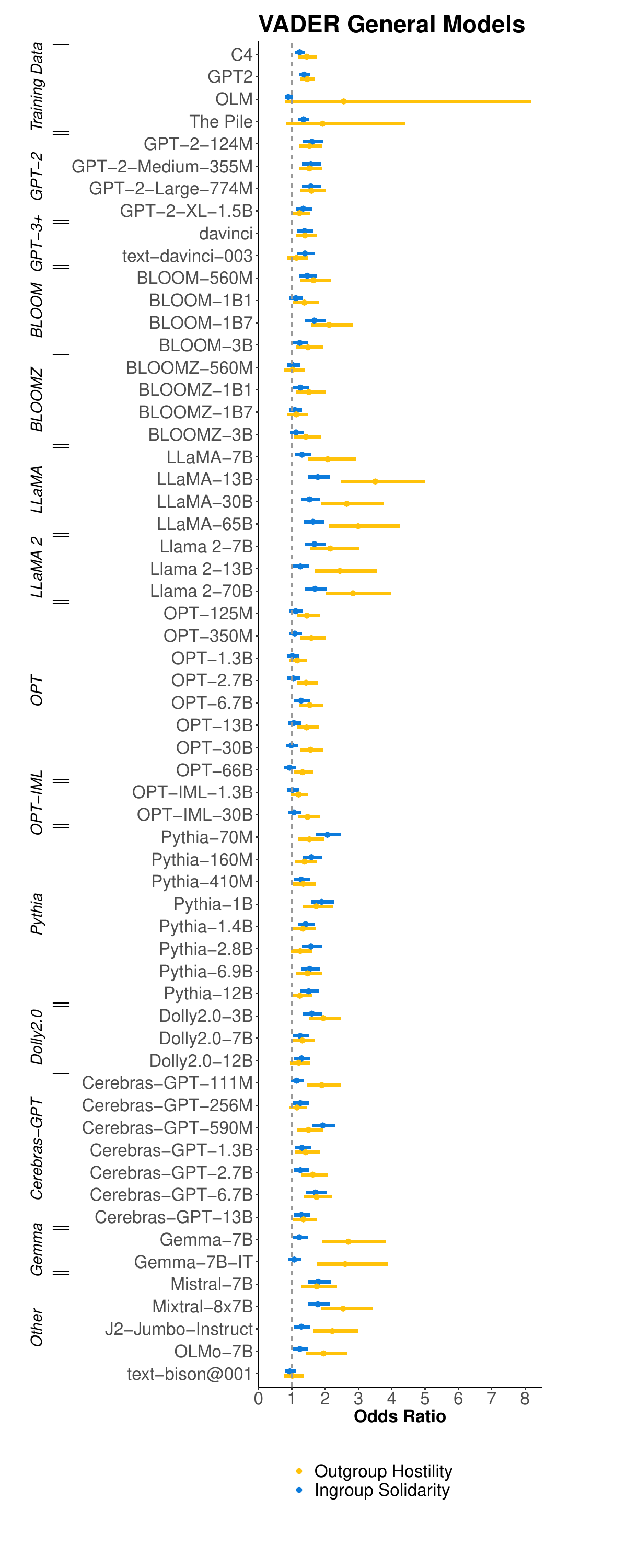}
\caption{Study 1: ingroup solidarity and outgroup hostility of general language models based on VADER.}
\label{fig:VADER}
\end{figure}
\newpage

\newpage
\section{Supporting Tables}

\begin{table}[htbp]
\centering
\caption{Study 1: Ratio of Sentences Kept Post Filtering (Default Prompt)}
\label{tab:study1_ratio_base}
\begin{tabular}{clrr}
\toprule
& Model & \% Good Sentence (We) & \% Good Sentence (They) \\
\midrule
1 & BLOOM-1B1 & 35.84189 & 47.63066 \\
2 & BLOOM-1B7 & 35.80927 & 41.92283 \\
3 & BLOOM-3B & 33.79784 & 44.87732 \\
4 & BLOOM-560M & 37.27511 & 52.94227 \\
5 & BLOOMZ-1B1 & 35.06366 & 47.31352 \\
6 & BLOOMZ-1B7 & 35.98343 & 49.73124 \\
7 & BLOOMZ-3B & 37.06725 & 48.20288 \\
8 & BLOOMZ-560M & 37.86466 & 53.35323 \\
9 & Cerebras-GPT-1.3B & 35.17987 & 44.24836 \\
10 & Cerebras-GPT-2.7B & 35.05258 & 46.81686 \\
11 & Cerebras-GPT-6.7B & 35.52036 & 44.87952 \\
12 & Cerebras-GPT-13B & 32.93599 & 44.31911 \\
13 & Cerebras-GPT-111M & 33.51308 & 43.71180 \\
14 & Cerebras-GPT-256M & 33.83225 & 38.24923 \\
15 & Cerebras-GPT-590M & 34.19168 & 44.96112 \\
16 & Dolly2.0-3B & 26.85944 & 40.70248 \\
17 & Dolly2.0-7B & 30.54299 & 44.69344 \\
18 & Dolly2.0-12B & 31.75398 & 45.93918 \\
19 & GPT-2-124M & 40.37771 & 55.53623 \\
20 & GPT-2-Large-774M & 38.32347 & 51.69243 \\
21 & GPT-2-Medium-355M & 40.79678 & 56.62109 \\
22 & GPT-2-XL-1.5B & 38.32853 & 52.00079 \\
23 & J2-Jumbo-Instruct & 8.24708 & 40.63225 \\
24 & LLaMA-7B & 31.45870 & 52.34554 \\
25 & LLaMA-13B & 31.01427 & 53.76202 \\
26 & LLaMA-30B & 31.11718 & 51.56279 \\
27 & LLaMA-65B & 30.43222 & 49.79095 \\
28 & Llama 2-7B & 29.56539 & 54.27316 \\
29 & Llama 2-13B & 31.65468 & 54.04595 \\
30 & Llama 2-70B & 30.82000 & 54.94000 \\
31 & OPT-1.3B & 36.08230 & 48.13764 \\
32 & OPT-2.7B & 36.78924 & 47.60135 \\
33 & OPT-6.7B & 34.44109 & 46.95296 \\
34 & OPT-13B & 34.21719 & 47.38253 \\
35 & OPT-30B & 35.46331 & 47.86950 \\
36 & OPT-66B & 34.57411 & 47.09156 \\
37 & OPT-125M & 41.32875 & 51.00708 \\
38 & OPT-350M & 39.70508 & 49.59024 \\
39 & OPT-IML-1.3B & 34.48446 & 48.03593 \\
40 & OPT-IML-30B & 34.03788 & 46.23257 \\
41 & Pythia-1.4B & 30.88540 & 38.71630 \\
42 & Pythia-1B & 30.79407 & 39.69094 \\
43 & Pythia-2.8B & 30.52154 & 38.84142 \\
44 & Pythia-6.9B & 31.88493 & 42.89527 \\
45 & Pythia-12B & 31.37255 & 40.61988 \\
46 & Pythia-70M & 33.91139 & 40.54834 \\
47 & Pythia-160M & 40.35311 & 44.33802 \\
48 & Pythia-410M & 28.50035 & 37.81067 \\
49 & davinci & 75.80000 & 81.04000 \\
50 & text-bison@001 & 26.46000 & 30.56000 \\
51 & text-davinci-003 & 58.14477 & 67.91344 \\
\bottomrule
\end{tabular}
\end{table}

\begin{table}[htbp]
\centering
\caption{Study 1: Ratio of sentences kept after filtering (instruction prompt)}
\label{tab:study1_ratio_instruct}
\begin{tabular}{clrr}
\toprule
& Model & \% Good Sentence (We) & \% Good Sentence (They) \\
\midrule
1 & Alpaca-7B & 29.15556 & 58.66667 \\
2 & Dolly2.0-3B & 60.38121 & 68.01661 \\
3 & Dolly2.0-7B & 64.37288 & 72.30823 \\
4 & Dolly2.0-12B & 54.24016 & 67.17896 \\
5 & Flan-T5-XL-3B & 59.16667 & 65.46667 \\
6 & Flan-T5-XXL-11B & 67.23333 & 76.23333 \\
7 & Flan-UL2-20B & 59.13333 & 59.36667 \\
8 & GPT-4 & 61.60000 & 84.43333 \\
9 & J2-Jumbo-Instruct & 56.06667 & 71.90000 \\
10 & Llama 2-7B-chat & 34.44481 & 64.70981 \\
11 & Llama 2-13B-chat & 37.33333 & 71.35712 \\
12 & Llama 2-70B-chat & 28.58732 & 57.89942 \\
13 & chat-bison@001 & 57.70000 & 76.46667 \\
14 & text-bison@001 & 60.03367 & 79.73746 \\
15 & text-davinci-003 & 55.86667 & 78.73333 \\
\bottomrule
\end{tabular}
\end{table}

\begin{table}[htbp] \footnotesize
  \centering
  \caption{Ingroup Solidarity and Outgroup Hostility of Base LLMs. Part 1}
  \label{tableBaseLLMs1}
  \begin{tabular}{lcc}
    \toprule
      & Ingroup Solidarity & Outgroup Hostility \\
    \midrule
    GPT-2-124M & 2.53$^{***}$ & 2.34$^{***}$ \\
     & (9.57, p = 0.00, [2.09, 3.05]) & (7.27, p = 0.00, [1.86, 2.95]) \\
    GPT-2-Medium-355M & 2.32$^{***}$ & 1.95$^{***}$ \\
     & (8.66, p = 0.00, [1.91, 2.80]) & (5.87, p = 0.00, [1.56, 2.43]) \\
    GPT-2-Large-774M & 2.16$^{***}$ & 2.45$^{***}$ \\
     & (8.03, p = 0.00, [1.79, 2.60]) & (7.69, p = 0.00, [1.95, 3.08]) \\
    GPT-2-XL-1.5B & 2.20$^{***}$ & 1.80$^{***}$ \\
     & (8.16, p = 0.00, [1.82, 2.66]) & (5.16, p = 0.00, [1.44, 2.26]) \\
    davinci & 1.96$^{***}$ & 1.76$^{***}$ \\
     & (6.63, p = 0.00, [1.61, 2.40]) & (5.06, p = 0.00, [1.41, 2.19]) \\
    text-davinci-003 & 2.61$^{***}$ & 1.89$^{***}$ \\
     & (10.06, p = 0.00, [2.17, 3.15]) & (4.48, p = 0.00, [1.43, 2.50]) \\
         BLOOM-560M & 1.60$^{***}$ & 1.58$^{**}$ \\
     & (5.14, p = 0.00, [1.34, 1.91]) & (3.10, p = 0.00, [1.18, 2.10]) \\
    BLOOM-1B1 & 1.58$^{***}$ & 2.07$^{***}$ \\
     & (4.99, p = 0.00, [1.32, 1.89]) & (5.06, p = 0.00, [1.56, 2.74]) \\
    BLOOM-1B7 & 2.54$^{***}$ & 2.76$^{***}$ \\
     & (9.78, p = 0.00, [2.11, 3.06]) & (6.80, p = 0.00, [2.06, 3.70]) \\
    BLOOM-3B & 1.82$^{***}$ & 2.08$^{***}$ \\
     & (6.41, p = 0.00, [1.51, 2.18]) & (5.38, p = 0.00, [1.59, 2.71]) \\
    BLOOMZ-560M & 1.02 & 1.13 \\
     & (0.22, p = 0.83, [0.85, 1.23]) & (0.78, p = 0.44, [0.83, 1.52]) \\
    BLOOMZ-1B1 & 1.82$^{***}$ & 1.42$^{*}$ \\
     & (6.42, p = 0.00, [1.52, 2.19]) & (2.35, p = 0.02, [1.06, 1.91]) \\
    BLOOMZ-1B7 & 1.70$^{***}$ & 1.13 \\
     & (5.24, p = 0.00, [1.39, 2.07]) & (0.89, p = 0.37, [0.86, 1.49]) \\
    BLOOMZ-3B & 1.73$^{***}$ & 1.15 \\
     & (5.74, p = 0.00, [1.43, 2.08]) & (0.99, p = 0.32, [0.87, 1.51]) \\
    OPT-125M & 1.92$^{***}$ & 2.31$^{***}$ \\
     & (6.86, p = 0.00, [1.59, 2.31]) & (7.65, p = 0.00, [1.87, 2.87]) \\
    OPT-350M & 2.13$^{***}$ & 2.02$^{***}$ \\
     & (7.72, p = 0.00, [1.76, 2.57]) & (6.79, p = 0.00, [1.65, 2.48]) \\
    OPT-1.3B & 1.80$^{***}$ & 1.71$^{***}$ \\
     & (6.09, p = 0.00, [1.49, 2.18]) & (5.08, p = 0.00, [1.39, 2.10]) \\
    OPT-2.7B & 2.43$^{***}$ & 2.02$^{***}$ \\
     & (9.00, p = 0.00, [2.00, 2.95]) & (6.93, p = 0.00, [1.66, 2.47]) \\
    OPT-6.7B & 2.63$^{***}$ & 2.42$^{***}$ \\
     & (9.83, p = 0.00, [2.17, 3.20]) & (8.43, p = 0.00, [1.97, 2.98]) \\
    OPT-13B & 1.98$^{***}$ & 2.00$^{***}$ \\
     & (6.94, p = 0.00, [1.63, 2.40]) & (6.76, p = 0.00, [1.64, 2.45]) \\
    OPT-30B & 1.93$^{***}$ & 1.87$^{***}$ \\
     & (6.65, p = 0.00, [1.59, 2.35]) & (6.15, p = 0.00, [1.53, 2.28]) \\
    OPT-66B & 1.86$^{***}$ & 2.01$^{***}$ \\
     & (6.31, p = 0.00, [1.54, 2.26]) & (6.85, p = 0.00, [1.64, 2.45]) \\
    OPT-IML-1.3B & 1.71$^{***}$ & 1.55$^{***}$ \\
     & (5.33, p = 0.00, [1.40, 2.08]) & (4.46, p = 0.00, [1.28, 1.88]) \\
    OPT-IML-30B & 2.52$^{***}$ & 2.57$^{***}$ \\
     & (9.12, p = 0.00, [2.07, 3.07]) & (9.09, p = 0.00, [2.10, 3.16]) \\
    \bottomrule
  \end{tabular}
    \begin{flushleft}
    \small\textit{Note: Each value represents an odds ratio from a logistic regression fitted on a total of two thousand ingroup and outgroup sentences predicting the whether the sentence is positive (for ingroup solidarity) based on whether the sentence is ingroup (vs. outgroup), and negative (for outgroup hostility) based on whether the sentence is outgroup (vs. ingroup), and control variables. See Methods for more details. $*** p < 0.001; ** p < 0.01; * p < 0.05.$}
  \end{flushleft}
\end{table}

\begin{table}[htbp] \footnotesize
  \centering
  \caption{Ingroup Solidarity and Outgroup Hostility of Base LLMs. Part 2}
  \label{tableBaseLLMs2}
  \begin{tabular}{lcc}
    \toprule
      & Ingroup Solidarity & Outgroup Hostility \\
    \midrule
    Pythia-70M & 2.03$^{***}$ & 2.85$^{***}$ \\
     & (6.41, p = 0.00, [1.64, 2.53]) & (8.18, p = 0.00, [2.22, 3.66]) \\
    Pythia-160M & 2.35$^{***}$ & 2.42$^{***}$ \\
     & (8.38, p = 0.00, [1.92, 2.87]) & (7.36, p = 0.00, [1.91, 3.06]) \\
    Pythia-410M & 1.88$^{***}$ & 1.79$^{***}$ \\
     & (6.66, p = 0.00, [1.56, 2.27]) & (4.89, p = 0.00, [1.42, 2.25]) \\
    Pythia-1B & 2.40$^{***}$ & 1.98$^{***}$ \\
     & (9.03, p = 0.00, [1.99, 2.91]) & (5.69, p = 0.00, [1.56, 2.50]) \\
    Pythia-1.4B & 1.94$^{***}$ & 1.85$^{***}$ \\
     & (7.02, p = 0.00, [1.61, 2.33]) & (5.06, p = 0.00, [1.46, 2.35]) \\
    Pythia-2.8B & 2.17$^{***}$ & 1.83$^{***}$ \\
     & (7.99, p = 0.00, [1.79, 2.63]) & (5.18, p = 0.00, [1.46, 2.30]) \\
    Pythia-6.9B & 1.96$^{***}$ & 2.10$^{***}$ \\
     & (7.06, p = 0.00, [1.63, 2.36]) & (5.79, p = 0.00, [1.64, 2.71]) \\
    Pythia-12B & 1.97$^{***}$ & 2.17$^{***}$ \\
     & (6.89, p = 0.00, [1.62, 2.38]) & (6.56, p = 0.00, [1.72, 2.73]) \\
    Dolly2.0-3B & 2.31$^{***}$ & 2.69$^{***}$ \\
     & (8.65, p = 0.00, [1.91, 2.79]) & (8.25, p = 0.00, [2.13, 3.40]) \\
    Dolly2.0-7B & 1.80$^{***}$ & 1.70$^{***}$ \\
     & (6.20, p = 0.00, [1.50, 2.17]) & (4.16, p = 0.00, [1.32, 2.18]) \\
    Dolly2.0-12B & 1.96$^{***}$ & 1.91$^{***}$ \\
     & (7.11, p = 0.00, [1.63, 2.36]) & (5.25, p = 0.00, [1.50, 2.44]) \\
    Cerebras-GPT-111M & 1.30$^{**}$ & 2.56$^{***}$ \\
     & (2.60, p = 0.01, [1.07, 1.57]) & (7.30, p = 0.00, [1.99, 3.30]) \\
    Cerebras-GPT-256M & 1.57$^{***}$ & 1.56$^{***}$ \\
     & (4.18, p = 0.00, [1.27, 1.94]) & (3.77, p = 0.00, [1.24, 1.96]) \\
    Cerebras-GPT-590M & 2.64$^{***}$ & 2.09$^{***}$ \\
     & (10.19, p = 0.00, [2.19, 3.19]) & (5.56, p = 0.00, [1.61, 2.71]) \\
    Cerebras-GPT-1.3B & 2.24$^{***}$ & 1.97$^{***}$ \\
     & (8.52, p = 0.00, [1.86, 2.70]) & (5.46, p = 0.00, [1.54, 2.51]) \\
    Cerebras-GPT-2.7B & 1.90$^{***}$ & 2.45$^{***}$ \\
     & (6.71, p = 0.00, [1.57, 2.29]) & (7.46, p = 0.00, [1.94, 3.10]) \\
    Cerebras-GPT-6.7B & 2.43$^{***}$ & 1.77$^{***}$ \\
     & (9.07, p = 0.00, [2.01, 2.95]) & (4.72, p = 0.00, [1.40, 2.25]) \\
    Cerebras-GPT-13B & 1.98$^{***}$ & 2.40$^{***}$ \\
     & (7.24, p = 0.00, [1.65, 2.38]) & (7.04, p = 0.00, [1.88, 3.05]) \\
    Mistral-7B & 1.94$^{***}$ & 3.00$^{***}$ \\
     & (7.07, p = 0.00, [1.61, 2.32]) & (6.69, p = 0.00, [2.17, 4.14]) \\
    Mixtral-8x7B & 2.15$^{***}$ & 3.76$^{***}$ \\
     & (8.14, p = 0.00, [1.79, 2.58]) & (8.05, p = 0.00, [2.72, 5.19]) \\
    J2-Jumbo-Instruct & 2.01$^{***}$ & 2.03$^{***}$ \\
     & (7.57, p = 0.00, [1.68, 2.41]) & (4.54, p = 0.00, [1.50, 2.76]) \\
    OLMo-7B & 1.74$^{***}$ & 3.01$^{***}$ \\
     & (6.02, p = 0.00, [1.45, 2.09]) & (6.47, p = 0.00, [2.16, 4.21]) \\
    text-bison@001 & 0.96 & 1.03 \\
     & (-0.43, p = 0.67, [0.81, 1.15]) & (0.21, p = 0.83, [0.77, 1.39]) \\
    \bottomrule
  \end{tabular}
    \begin{flushleft}
    \small\textit{Note: Each value represents an odds ratio from a logistic regression fitted on a total of two thousand ingroup and outgroup sentences predicting the whether the sentence is positive (for ingroup solidarity) based on whether the sentence is ingroup (vs. outgroup), and negative (for outgroup hostility) based on whether the sentence is outgroup (vs. ingroup), and control variables. See Methods for more details. $*** p < 0.001; ** p < 0.01; * p < 0.05.$}
  \end{flushleft}
\end{table}

\begin{table}[htbp] \footnotesize
  \centering
  \caption{Ingroup Solidarity and Outgroup Hostility of Outlier Models}
  \label{tableOutlierModels}
  \begin{tabular}{lcc}
    \toprule
      & Ingroup Solidarity & Outgroup Hostility \\
    \midrule
    LLaMA-7B & 1.94$^{***}$ & 3.11$^{***}$ \\
     & (7.11, p = 0.00, [1.61, 2.33]) & (5.78, p = 0.00, [2.12, 4.57]) \\
    LLaMA-13B & 2.34$^{***}$ & 7.90$^{***}$ \\
     & (8.99, p = 0.00, [1.94, 2.81]) & (8.40, p = 0.00, [4.88, 12.80]) \\
    LLaMA-30B & 2.02$^{***}$ & 6.08$^{***}$ \\
     & (7.58, p = 0.00, [1.69, 2.43]) & (7.60, p = 0.00, [3.82, 9.68]) \\
    LLaMA-65B & 2.14$^{***}$ & 6.00$^{***}$ \\
     & (8.19, p = 0.00, [1.78, 2.57]) & (7.38, p = 0.00, [3.73, 9.66]) \\
    Llama 2-7B & 2.22$^{***}$ & 3.87$^{***}$ \\
     & (8.48, p = 0.00, [1.85, 2.67]) & (6.45, p = 0.00, [2.56, 5.84]) \\
    Llama 2-13B & 1.69$^{***}$ & 5.65$^{***}$ \\
     & (5.61, p = 0.00, [1.41, 2.04]) & (7.65, p = 0.00, [3.62, 8.80]) \\
    Llama 2-70B & 2.12$^{***}$ & 4.67$^{***}$ \\
     & (8.07, p = 0.00, [1.77, 2.55]) & (7.95, p = 0.00, [3.19, 6.83]) \\
    Gemma-7B & 1.52$^{***}$ & 8.42$^{***}$ \\
     & (4.57, p = 0.00, [1.27, 1.83]) & (8.69, p = 0.00, [5.21, 13.62]) \\
    Gemma-7B-IT & 0.94 & 8.66$^{***}$ \\
     & (-0.60, p = 0.55, [0.78, 1.14]) & (7.99, p = 0.00, [5.10, 14.70]) \\
    \bottomrule
  \end{tabular}
      \begin{flushleft}
    \small\textit{Note: Each value represents an odds ratio from a logistic regression fitted on a total of two thousand ingroup and outgroup sentences predicting the whether the sentence is positive (for ingroup solidarity) based on whether the sentence is ingroup (vs. outgroup), and negative (for outgroup hostility) based on whether the sentence is outgroup (vs. ingroup), and control variables. See Methods for more details. $*** p < 0.001; ** p < 0.01; * p < 0.05.$}
  \end{flushleft}
\end{table}

\begin{table}[htbp] \footnotesize
  \centering
  \caption{Ingroup Solidarity and Outgroup Hostility of Pre-training Datasets}
  \label{tableHumans}
  \begin{tabular}{lcc}
    \toprule
      & Ingroup Solidarity & Outgroup Hostility \\
    \midrule
C4 & 1.65*** & 2.21*** \\
 & (7.74, p = 0.00, [1.46, 1.88]) & (7.11, p = 0.00, [1.78, 2.76]) \\
GPT2 & 1.92*** & 1.66*** \\
 & (8.50, p = 0.00, [1.65, 2.24]) & (6.95, p = 0.00, [1.44, 1.91]) \\
OLM & 1.36*** & 1.80*** \\
 & (4.76, p = 0.00, [1.20, 1.55]) & (5.61, p = 0.00, [1.46, 2.20]) \\
The Pile & 1.97*** & 1.43*** \\
 & (9.31, p = 0.00, [1.71, 2.27]) & (4.40, p = 0.00, [1.22, 1.67]) \\
    \bottomrule
  \end{tabular}
      \begin{flushleft}
    \small\textit{Note: Each value represents an odds ratio from a logistic regression fitted on a total of four thousand ingroup and outgroup sentences predicting the whether the sentence is positive (for ingroup solidarity) based on whether the sentence is ingroup (vs. outgroup), and negative (for outgroup hostility) based on whether the sentence is outgroup (vs. ingroup), and control variables. See Methods for more details. $*** p < 0.001; ** p < 0.01; * p < 0.05.$}
  \end{flushleft}
\end{table}

\begin{table}[htbp] \footnotesize
  \centering
  \caption{Ingroup Solidarity and Outgroup Hostility of a Subset Base LLMs Controlling for Sentence Topic. Part 1}
  \label{tableBaseSTM1}
  \begin{tabular}{lcc}
    \toprule
      & Ingroup Solidarity & Outgroup Hostility \\
    \midrule
        text-bison@001 & 1.10 & 0.92 \\
        & (0.85, $p = 0.39$, [0.88, 1.38]) & (-0.40, $p = 0.69$, [0.63, 1.35]) \\
        GPT-2-Medium-355M & 2.30 *** & 1.95 *** \\
        & (7.63, $p = 0.00$, [1.86, 2.85]) & (5.25, $p = 0.00$, [1.52, 2.51]) \\
        GPT-2-Large-774M & 2.08 *** & 2.37 *** \\
        & (6.61, $p = 0.00$, [1.67, 2.58]) & (6.65, $p = 0.00$, [1.84, 3.06]) \\
        GPT-2-XL-1.5B & 2.08 *** & 1.57 *** \\
        & (6.61, $p = 0.00$, [1.68, 2.59]) & (3.60, $p = 0.00$, [1.23, 2.01]) \\
        BLOOM-560M & 1.42 ** & 1.44 * \\
        & (3.29, $p = 0.00$, [1.15, 1.75]) & (2.23, $p = 0.03$, [1.04, 1.99]) \\
        BLOOM-1B1 & 1.31 * & 1.81 *** \\
        & (2.40, $p = 0.02$, [1.05, 1.62]) & (3.54, $p = 0.00$, [1.30, 2.51]) \\
        BLOOM-1B7 & 2.28 *** & 2.23 *** \\
        & (7.19, $p = 0.00$, [1.82, 2.85]) & (4.56, $p = 0.00$, [1.58, 3.14]) \\
        BLOOM-3B & 1.57 *** & 1.57 ** \\
        & (4.06, $p = 0.00$, [1.26, 1.96]) & (2.89, $p = 0.00$, [1.16, 2.13]) \\
        BLOOMZ-560M & 0.86 & 1.14 \\
        & (-1.32, $p = 0.19$, [0.69, 1.07]) & (0.75, $p = 0.45$, [0.81, 1.60]) \\
        BLOOMZ-1B1 & 1.71 *** & 1.09 \\
        & (4.78, $p = 0.00$, [1.37, 2.13]) & (0.52, $p = 0.60$, [0.78, 1.54]) \\
        BLOOMZ-1B7 & 1.52 *** & 1.18 \\
        & (3.72, $p = 0.00$, [1.22, 1.89]) & (1.06, $p = 0.29$, [0.87, 1.61]) \\
        BLOOMZ-3B & 1.50 *** & 1.10 \\
        & (3.66, $p = 0.00$, [1.21, 1.87]) & (0.57, $p = 0.57$, [0.80, 1.51]) \\
                LLaMA-7B & 2.01 *** & 2.79 *** \\
        & (6.05, $p = 0.00$, [1.60, 2.52]) & (4.20, $p = 0.00$, [1.73, 4.49]) \\
        LLaMA-13B & 2.39 *** & 5.23 *** \\
        & (7.61, $p = 0.00$, [1.91, 2.99]) & (5.99, $p = 0.00$, [3.04, 8.99]) \\
        LLaMA-30B & 2.18 *** & 3.92 *** \\
        & (6.86, $p = 0.00$, [1.75, 2.73]) & (5.24, $p = 0.00$, [2.35, 6.53]) \\
        LLaMA-65B & 2.09 *** & 4.09 *** \\
        & (6.51, $p = 0.00$, [1.67, 2.61]) & (5.00, $p = 0.00$, [2.36, 7.11]) \\
        Llama 2-7B & 2.20 *** & 2.36 *** \\
        & (6.74, $p = 0.00$, [1.75, 2.77]) & (3.59, $p = 0.00$, [1.48, 3.77]) \\
        Llama 2-13B & 1.74 *** & 5.79 *** \\
        & (4.93, $p = 0.00$, [1.39, 2.16]) & (6.70, $p = 0.00$, [3.46, 9.68]) \\
        Llama 2-70B & 2.11 *** & 3.26 *** \\
        & (6.75, $p = 0.00$, [1.70, 2.62]) & (5.20, $p = 0.00$, [2.09, 5.09]) \\
        OPT-125M & 1.74 *** & 1.68 *** \\
        & (5.13, $p = 0.00$, [1.41, 2.15]) & (4.22, $p = 0.00$, [1.32, 2.15]) \\
        OPT-350M & 2.11 *** & 1.63 *** \\
        & (6.57, $p = 0.00$, [1.69, 2.63]) & (4.11, $p = 0.00$, [1.29, 2.06]) \\
        OPT-1.3B & 1.79 *** & 1.40 ** \\
        & (5.03, $p = 0.00$, [1.43, 2.25]) & (2.76, $p = 0.01$, [1.10, 1.77]) \\
        OPT-2.7B & 2.20 *** & 1.51 *** \\
        & (6.83, $p = 0.00$, [1.75, 2.75]) & (3.57, $p = 0.00$, [1.21, 1.90]) \\
        OPT-6.7B & 2.65 *** & 2.04 *** \\
        & (8.50, $p = 0.00$, [2.12, 3.32]) & (6.06, $p = 0.00$, [1.62, 2.57]) \\
        OPT-13B & 1.79 *** & 1.48 *** \\
        & (5.18, $p = 0.00$, [1.44, 2.24]) & (3.36, $p = 0.00$, [1.18, 1.86]) \\
        OPT-30B & 1.98 *** & 1.69 *** \\
        & (5.93, $p = 0.00$, [1.58, 2.48]) & (4.49, $p = 0.00$, [1.34, 2.12]) \\
        OPT-66B & 1.85 *** & 1.51 *** \\
        & (5.30, $p = 0.00$, [1.47, 2.32]) & (3.56, $p = 0.00$, [1.20, 1.90]) \\
    \bottomrule
  \end{tabular}
      \begin{flushleft}
    \small\textit{Note: Each value represents an odds ratio from a logistic regression fitted on a total of two thousand ingroup and outgroup sentences predicting the whether the sentence is positive (for ingroup solidarity) based on whether the sentence is ingroup (vs. outgroup), and negative (for outgroup hostility) based on whether the sentence is outgroup (vs. ingroup), and control variables. See Methods for more details. $*** p < 0.001; ** p < 0.01; * p < 0.05.$}
  \end{flushleft}
\end{table}

\begin{table}[htbp] \footnotesize
  \centering
  \caption{Ingroup Solidarity and Outgroup Hostility of a Subset Base LLMs Controlling for Sentence Topic. Part 2}
  \label{tableBaseSTM2}
  \begin{tabular}{lcc}
    \toprule
      & Ingroup Solidarity & Outgroup Hostility \\
    \midrule
        OPT-IML-1.3B & 1.75 *** & 1.23 \\
        & (4.87, $p = 0.00$, [1.40, 2.20]) & (1.89, $p = 0.06$, [0.99, 1.52]) \\
        OPT-IML-30B & 2.54 *** & 1.95 *** \\
        & (7.80, $p = 0.00$, [2.01, 3.22]) & (5.64, $p = 0.00$, [1.55, 2.46]) \\
        Pythia-70M & 2.08 *** & 2.98 *** \\
        & (5.74, $p = 0.00$, [1.62, 2.67]) & (7.59, $p = 0.00$, [2.25, 3.94]) \\
        Pythia-160M & 2.45 *** & 2.34 *** \\
        & (7.91, $p = 0.00$, [1.96, 3.05]) & (6.53, $p = 0.00$, [1.81, 3.02]) \\
        Pythia-410M & 1.93 *** & 1.85 *** \\
        & (6.02, $p = 0.00$, [1.56, 2.40]) & (4.65, $p = 0.00$, [1.43, 2.40]) \\
        Pythia-1B & 2.21 *** & 2.02 *** \\
        & (6.81, $p = 0.00$, [1.76, 2.78]) & (5.10, $p = 0.00$, [1.54, 2.64]) \\
        Pythia-1.4B & 1.95 *** & 1.79 *** \\
        & (6.08, $p = 0.00$, [1.57, 2.42]) & (4.24, $p = 0.00$, [1.37, 2.33]) \\
        Pythia-2.8B & 2.28 *** & 1.86 *** \\
        & (7.32, $p = 0.00$, [1.83, 2.85]) & (4.67, $p = 0.00$, [1.43, 2.41]) \\
        Pythia-6.9B & 2.18 *** & 2.09 *** \\
        & (6.94, $p = 0.00$, [1.75, 2.72]) & (5.02, $p = 0.00$, [1.57, 2.79]) \\
        Pythia-12B & 2.17 *** & 1.94 *** \\
        & (6.69, $p = 0.00$, [1.73, 2.73]) & (4.93, $p = 0.00$, [1.49, 2.52]) \\
        Dolly2.0-3B & 2.35 *** & 2.53 *** \\
        & (7.29, $p = 0.00$, [1.87, 2.95]) & (6.69, $p = 0.00$, [1.93, 3.32]) \\
        Dolly2.0-7B & 1.92 *** & 1.61 ** \\
        & (5.76, $p = 0.00$, [1.54, 2.40]) & (3.24, $p = 0.00$, [1.21, 2.15]) \\
        Dolly2.0-12B & 2.17 *** & 2.14 *** \\
        & (6.90, $p = 0.00$, [1.74, 2.71]) & (5.27, $p = 0.00$, [1.61, 2.85]) \\
                Cerebras-GPT-111M & 1.40 ** & 2.09 *** \\
        & (2.84, $p = 0.00$, [1.11, 1.76]) & (5.19, $p = 0.00$, [1.58, 2.76]) \\
        Cerebras-GPT-256M & 1.47 ** & 1.58 *** \\
        & (3.22, $p = 0.00$, [1.16, 1.86]) & (3.60, $p = 0.00$, [1.23, 2.03]) \\
        Cerebras-GPT-590M & 2.61 *** & 1.90 *** \\
        & (8.64, $p = 0.00$, [2.10, 3.25]) & (4.41, $p = 0.00$, [1.43, 2.53]) \\
        Cerebras-GPT-1.3B & 2.26 *** & 1.86 *** \\
        & (7.62, $p = 0.00$, [1.83, 2.78]) & (4.54, $p = 0.00$, [1.42, 2.43]) \\
        Cerebras-GPT-2.7B & 1.81 *** & 2.47 *** \\
        & (5.43, $p = 0.00$, [1.46, 2.25]) & (6.66, $p = 0.00$, [1.89, 3.22]) \\
        Cerebras-GPT-6.7B & 2.54 *** & 1.48 ** \\
        & (8.23, $p = 0.00$, [2.03, 3.17]) & (2.92, $p = 0.00$, [1.14, 1.92]) \\
        Cerebras-GPT-13B & 2.07 *** & 2.28 *** \\
        & (6.47, $p = 0.00$, [1.66, 2.59]) & (5.79, $p = 0.00$, [1.73, 3.02]) \\
        J2-Jumbo-Instruct & 1.80 *** & 1.79 ** \\
        & (5.47, $p = 0.00$, [1.46, 2.22]) & (3.10, $p = 0.00$, [1.24, 2.59]) \\
        davinci & 1.89 *** & 1.72 *** \\
        & (5.77, $p = 0.00$, [1.52, 2.35]) & (4.51, $p = 0.00$, [1.36, 2.18]) \\
        text-davinci-003 & 2.58 *** & 1.81 *** \\
        & (8.67, $p = 0.00$, [2.08, 3.20]) & (3.62, $p = 0.00$, [1.31, 2.49]) \\
    \bottomrule
  \end{tabular}
      \begin{flushleft}
    \small\textit{Note: Each value represents an odds ratio from a logistic regression fitted on a total of two thousand ingroup and outgroup sentences predicting the whether the sentence is positive (for ingroup solidarity) based on whether the sentence is ingroup (vs. outgroup), and negative (for outgroup hostility) based on whether the sentence is outgroup (vs. ingroup), and control variables. See Methods for more details. $*** p < 0.001; ** p < 0.01; * p < 0.05.$}
  \end{flushleft}
\end{table}

\begin{table}[htbp] \footnotesize
\caption{General Models Overall}
\label{tableGeneralModelsOverall}
\centering
\begin{tabular}{lccc ccc}
\toprule
 & \multicolumn{3}{c}{\textbf{Positive}} & \multicolumn{3}{c}{\textbf{Negative}} \\
\cmidrule(lr){2-4} \cmidrule(lr){5-7}
Predictors & \multicolumn{1}{c}{Odds Ratios} & \multicolumn{1}{c}{CI} & \multicolumn{1}{c}{$p$} & \multicolumn{1}{c}{Odds Ratios} & \multicolumn{1}{c}{CI} & \multicolumn{1}{c}{$p$} \\
\midrule
(Intercept) & 0.25 & 0.21 -- 0.30 & \textbf{$<$0.0001} & 0.13 & 0.10 -- 0.16 & \textbf{$<$0.0001} \\
source [we] & 1.93 & 1.89 -- 1.98 & \textbf{$<$0.0001} & & & \\
total tokens scaled & 1.13 & 1.11 -- 1.15 & \textbf{$<$0.0001} & 0.96 & 0.94 -- 0.98 & \textbf{$<$0.0001} \\
TTR & 1.95 & 1.65 -- 2.31 & \textbf{$<$0.0001} & 0.95 & 0.76 -- 1.17 & 0.6148 \\
source [they] & & & & 2.15 & 2.08 -- 2.23 & \textbf{$<$0.0001} \\
\midrule
\multicolumn{7}{l}{\textbf{Random Effects}} \\
$\sigma^2$ & \multicolumn{3}{l}{3.29} & \multicolumn{3}{l}{3.29} \\
$\tau_{00}$ & \multicolumn{3}{l}{0.13 model} & \multicolumn{3}{l}{0.27 model} \\
ICC & \multicolumn{3}{l}{0.04} & \multicolumn{3}{l}{0.08} \\
N & \multicolumn{3}{l}{56 model} & \multicolumn{3}{l}{56 model} \\
\midrule
Observations & \multicolumn{3}{l}{112000} & \multicolumn{3}{l}{112000} \\
Marginal $R^2$ / Conditional $R^2$ & \multicolumn{3}{l}{0.037 / 0.073} & \multicolumn{3}{l}{0.041 / 0.114} \\
\bottomrule
\end{tabular}
\end{table}

\begin{table}[htbp] \footnotesize
  \centering
  \caption{General Models: Effect of Model Size}
  \label{tableGeneralModelsModelSize}
  \begin{tabular}{lccc ccc}
    \toprule
    & \multicolumn{3}{c}{\textbf{Positive}} & \multicolumn{3}{c}{\textbf{Negative}} \\
    \cmidrule(lr){2-4} \cmidrule(lr){5-7}
    Predictors & Odds Ratios & CI & $p$ & Odds Ratios & CI & $p$ \\
    \midrule
    (Intercept) & 0.26 & 0.20 -- 0.33 & \textbf{$<$0.0001} & 0.11 & 0.08 -- 0.16 & \textbf{$<$0.0001} \\
    source [we] & 1.93 & 1.88 -- 1.98 & \textbf{$<$0.0001} & & & \\
    total tokens scaled & 1.14 & 1.12 -- 1.16 & \textbf{$<$0.0001} & 0.96 & 0.94 -- 0.98 & \textbf{$<$0.0001} \\
    TTR & 2.07 & 1.73 -- 2.47 & \textbf{$<$0.0001} & 0.96 & 0.77 -- 1.20 & 0.7179 \\
    model size scaled & 0.98 & 0.96 -- 0.99 & \textbf{0.0062} & 0.97 & 0.95 -- 0.99 & \textbf{0.0167} \\
    source [we] $\times$ model size scaled & 1.02 & 1.00 -- 1.04 & \textbf{0.0287} & & & \\
    source [they] & & & & 2.15 & 2.08 -- 2.23 & \textbf{$<$0.0001} \\
    source [they] $\times$ model size scaled & & & & 1.08 & 1.05 -- 1.11 & \textbf{$<$0.0001} \\
\midrule
\multicolumn{7}{l}{\textbf{Random Effects}} \\
    $\sigma^2$ & \multicolumn{3}{l}{3.29} & \multicolumn{3}{l}{3.29} \\
    $\tau_{00}$ & \multicolumn{3}{l}{0.12 model.family} & \multicolumn{3}{l}{0.31 model.family} \\
    ICC & \multicolumn{3}{l}{0.04} & \multicolumn{3}{l}{0.09} \\
    N & \multicolumn{3}{l}{13 model.family} & \multicolumn{3}{l}{13 model.family} \\
    \midrule
    Observations & \multicolumn{3}{l}{104000} & \multicolumn{3}{l}{104000} \\
    Marginal R$^2$ / Conditional R$^2$ & \multicolumn{3}{l}{0.038 / 0.071} & \multicolumn{3}{l}{0.044 / 0.125} \\
    \bottomrule
  \end{tabular}
\end{table}

\begin{table}[htbp] \footnotesize
  \centering
  \caption{Ingroup Solidarity and Outgroup Hostility of Instruction fine-tuned Models}
  \label{tableInstructLLMs}
  \begin{tabular}{lcc}
    \toprule
      & Ingroup Solidarity & Outgroup Hostility \\
    \midrule
        GPT-4 & 1.04 & 1.00 \\
          & (0.37, p = 0.71, [0.86, 1.25]) & (0.02, p = 0.98, [0.71, 1.42]) \\
    text-davinci-003 & 1.29 ** & 1.59 ** \\
                      & (2.72, p = 0.01, [1.07, 1.56]) & (2.59, p = 0.01, [1.12, 2.25]) \\
    Llama 2-7B-chat & 1.37 *** & 0.79 \\
                     & (3.41, p = 0.00, [1.14, 1.65]) & (-1.70, p = 0.09, [0.60, 1.04]) \\
    Llama 2-13B-chat & 1.39 *** & 0.94 \\
                      & (3.53, p = 0.00, [1.16, 1.67]) & (-0.41, p = 0.68, [0.71, 1.26]) \\
    Llama 2-70B-chat & 1.11 & 0.72 * \\
                      & (1.17, p = 0.24, [0.93, 1.34]) & (-2.32, p = 0.02, [0.55, 0.95]) \\
    Dolly2.0-3B & 1.19 & 1.89 *** \\
                  & (1.84, p = 0.07, [0.99, 1.44]) & (4.06, p = 0.00, [1.39, 2.57]) \\
    Dolly2.0-7B & 0.98 & 1.43 * \\
                  & (-0.27, p = 0.79, [0.82, 1.17]) & (2.43, p = 0.02, [1.07, 1.90]) \\
    Dolly2.0-12B & 0.99 & 1.96 *** \\
                   & (-0.09, p = 0.93, [0.83, 1.18]) & (4.19, p = 0.00, [1.43, 2.68]) \\
    Flan-T5-XL-3B & 1.95 *** & 1.49 * \\
                  & (6.96, p = 0.00, [1.62, 2.35]) & (2.41, p = 0.02, [1.08, 2.06]) \\
    Flan-T5-XXL-11B & 1.63 *** & 1.14 \\
                    & (5.17, p = 0.00, [1.35, 1.96]) & (0.78, p = 0.43, [0.83, 1.56]) \\
    Flan-UL2-20B & 2.23 *** & 1.47 * \\
                 & (8.09, p = 0.00, [1.84, 2.71]) & (2.56, p = 0.01, [1.09, 1.97]) \\
    text-bison@001 & 1.54 *** & 1.03 \\
                   & (4.69, p = 0.00, [1.29, 1.85]) & (0.16, p = 0.87, [0.74, 1.42]) \\
    chat-bison@001 & 1.38 *** & 1.46 * \\
                   & (3.51, p = 0.00, [1.15, 1.65]) & (2.30, p = 0.02, [1.06, 2.02]) \\   
    OLMo-7B-Instruct & 1.77 *** & 1.38 \\
                      & (5.74, p = 0.00, [1.45, 2.14]) & (1.68, p = 0.09, [0.95, 2.02]) \\
    OLMo-7B-SFT & 1.72 *** & 0.97 \\
                  & (5.82, p = 0.00, [1.43, 2.07]) & (-0.19, p = 0.85, [0.70, 1.34]) \\
    Tulu-2-7B & 1.27 * & 0.99 \\
                & (2.56, p = 0.01, [1.06, 1.52]) & (-0.06, p = 0.95, [0.73, 1.34]) \\
    Tulu-2-13B & 1.63 *** & 0.96 \\
                 & (5.34, p = 0.00, [1.36, 1.95]) & (-0.28, p = 0.78, [0.71, 1.29]) \\
    Tulu-2-70B & 1.39 *** & 0.93 \\
                & (3.62, p = 0.00, [1.16, 1.66]) & (-0.48, p = 0.63, [0.69, 1.25]) \\
    Tulu-2-DPO-7B & 1.39 *** & 1.53 * \\
                   & (3.44, p = 0.00, [1.15, 1.67]) & (2.50, p = 0.01, [1.10, 2.14]) \\
    Tulu-2-DPO-13B & 1.04 & 0.67 * \\
                    & (0.40, p = 0.69, [0.86, 1.27]) & (-2.18, p = 0.03, [0.46, 0.96]) \\
    Tulu-2-DPO-70B & 0.92 & 0.79 \\
                    & (-0.87, p = 0.38, [0.77, 1.11]) & (-1.36, p = 0.17, [0.56, 1.11]) \\
    J2-Jumbo-Instruct & 1.45 *** & 1.14 \\
                       & (3.99, p = 0.00, [1.21, 1.74]) & (0.94, p = 0.35, [0.87, 1.51]) \\
    Alpaca-7B & 1.28 ** & 1.14 \\
                 & (2.73, p = 0.01, [1.07, 1.54]) & (0.78, p = 0.43, [0.82, 1.58]) \\
    Zephyr-7B-beta & 1.14 & 0.89 \\
                      & (1.46, p = 0.14, [0.96, 1.36]) & (-0.68, p = 0.49, [0.65, 1.23]) \\
    Starling-7B & 1.13 & 0.73 \\
                    & (1.31, p = 0.19, [0.94, 1.35]) & (-1.87, p = 0.06, [0.53, 1.01]) \\
    OpenChat3.5-7B & 0.97 & 0.68 * \\
                      & (-0.35, p = 0.73, [0.81, 1.16]) & (-2.21, p = 0.03, [0.49, 0.96]) \\
    Gemma-7B-IT & 1.32 ** & 1.27 \\
                    & (2.98, p = 0.00, [1.10, 1.58]) & (1.37, p = 0.17, [0.90, 1.78]) \\
    Mixtral-8x7B-Instruct & 1.14 & 0.99 \\
                              & (1.41, p = 0.16, [0.95, 1.36]) & (-0.06, p = 0.95, [0.71, 1.37]) \\
    \bottomrule
    \end{tabular}
    \begin{flushleft}
        \small\textit{Note: $*** p < 0.001; ** p < 0.01; * p < 0.05.$}
    \end{flushleft}
\end{table}

\begin{table}[htbp] \footnotesize
\caption{Instruction fine-tuned}
\label{tableGeneralModelsvsInstructionFT}
\centering
\begin{tabular}{lcccccc}
\toprule
\multicolumn{1}{c}{\textbf{}} & \multicolumn{3}{c}{\textbf{Positive}} & \multicolumn{3}{c}{\textbf{Negative}} \\
\cmidrule(r){2-4} \cmidrule(r){5-7}
Predictors & Odds Ratios & CI & $p$ & Odds Ratios & CI & $p$ \\
\midrule
(Intercept) & 0.22 & 0.17--0.30 & \textless0.0001 & 0.11 & 0.07--0.18 & \textless0.0001 \\
source [we] & 1.71 & 1.51--1.95 & \textless0.0001 & & & \\
total tokens scaled & 1.18 & 1.15--1.21 & \textless0.0001 & 0.94 & 0.90--0.97 & 0.0001 \\
TTR & 3.05 & 2.27--4.08 & \textless0.0001 & 0.98 & 0.67--1.42 & 0.9089 \\
instr. fine tuned & 0.81 & 0.76--0.86 & \textless0.0001 & 1.12 & 1.03--1.22 & 0.0115 \\
source [we] $\times$ instr. fine tuned & 1.02 & 0.93--1.10 & 0.7182 & & & \\
source [they] & & & & 2.91 & 2.45--3.46 & \textless0.0001 \\
source [they] $\times$ instr. fine tuned & & & & 0.76 & 0.68--0.85 & \textless0.0001 \\
\midrule
\multicolumn{7}{l}{\textbf{Random Effects}} \\
$\sigma^2$ & \multicolumn{3}{c}{3.29} & \multicolumn{3}{c}{3.29} \\
$\tau_{00}$ & \multicolumn{3}{c}{0.03 model.instruct} & \multicolumn{3}{c}{0.26 model.instruct} \\
ICC & \multicolumn{3}{c}{0.01} & \multicolumn{3}{c}{0.07} \\
N & \multicolumn{3}{c}{10 model.instruct} & \multicolumn{3}{c}{10 model.instruct} \\
\midrule
Observations & \multicolumn{3}{c}{40000} & \multicolumn{3}{c}{40000} \\
Marginal $R^2$ / Conditional $R^2$ & \multicolumn{3}{c}{0.035 / 0.043} & \multicolumn{3}{c}{0.036 / 0.105} \\
\bottomrule
\end{tabular}
\end{table}

\begin{table}[htbp] \footnotesize
  \centering
  \caption{Partisan Models Overall}
  \label{tablePartisanModelsOverall}
  \begin{tabular}{lcccccc}
    \toprule
\multicolumn{1}{c}{\textbf{}} & \multicolumn{3}{c}{\textbf{Positive}} & \multicolumn{3}{c}{\textbf{Negative}} \\
\cmidrule(r){2-4} \cmidrule(r){5-7}
Predictors & Odds Ratios & CI & $p$ & Odds Ratios & CI & $p$ \\
\midrule
    (Intercept) & 0.11 & 0.09 -- 0.14 & \textbf{$<$0.0001} & 0.17 & 0.14 -- 0.22 & \textbf{$<$0.0001} \\
    source [we] & 4.61 & 4.42 -- 4.82 & \textbf{$<$0.0001} & & & \\
    total tokens scaled & 1.05 & 1.02 -- 1.08 & \textbf{0.0005} & 1.17 & 1.14 -- 1.20 & \textbf{$<$0.0001} \\
    TTR & 1.79 & 1.40 -- 2.28 & \textbf{$<$0.0001} & 1.54 & 1.23 -- 1.94 & \textbf{0.0002} \\
    source [they] & & & & 6.50 & 6.24 -- 6.78 & \textbf{$<$0.0001} \\
\midrule
\multicolumn{7}{l}{\textbf{Random Effects}} \\
    $\sigma^2$ & \multicolumn{3}{c}{3.29} & \multicolumn{3}{c}{3.29} \\
    $\tau_{00}$ & \multicolumn{3}{c}{0.01 model} & \multicolumn{3}{c}{0.00 model} \\
    & \multicolumn{3}{c}{0.00 fine.tuned.party} & \multicolumn{3}{c}{0.00 fine.tuned.party} \\
    ICC & \multicolumn{3}{c}{0.00} & \multicolumn{3}{c}{} \\
    N & \multicolumn{3}{c}{12 model} & \multicolumn{3}{c}{12 model} \\
    & \multicolumn{3}{c}{2 fine.tuned.party} & \multicolumn{3}{c}{2 fine.tuned.party} \\
    \midrule
    Observations & \multicolumn{3}{c}{48000} & \multicolumn{3}{c}{48000} \\
    Marginal $R^2$ / Conditional $R^2$ & \multicolumn{3}{c}{0.152 / 0.155} & \multicolumn{3}{c}{0.210 / NA} \\
    \bottomrule
  \end{tabular}
\end{table}

\begin{table}[htbp] \footnotesize
  \centering
  \caption{Partisan Models Overall (Before Fine-Tuning)}
  \label{tablePartisamModelsOverallBeforeFT}
  \begin{tabular}{lcccccc}
\toprule
\multicolumn{1}{c}{\textbf{}} & \multicolumn{3}{c}{\textbf{Positive}} & \multicolumn{3}{c}{\textbf{Negative}} \\
\cmidrule(r){2-4} \cmidrule(r){5-7}
Predictors & Odds Ratios & CI & $p$ & Odds Ratios & CI & $p$ \\
\midrule
    (Intercept) & 0.23 & 0.16 -- 0.34 & \textbf{$<$0.0001} & 0.11 & 0.06 -- 0.18 & \textbf{$<$0.0001} \\
    source [we] & 1.86 & 1.76 -- 1.96 & \textbf{$<$0.0001} & & & \\
    total tokens scaled & 1.10 & 1.07 -- 1.14 & \textbf{$<$0.0001} & 0.91 & 0.87 -- 0.95 & \textbf{$<$0.0001} \\
    TTR & 2.09 & 1.43 -- 3.04 & \textbf{0.0001} & 1.13 & 0.68 -- 1.88 & 0.6236 \\
    source [they] & & & & 1.83 & 1.69 -- 1.97 & \textbf{$<$0.0001} \\
\midrule
\multicolumn{7}{l}{\textbf{Random Effects}} \\
    $\sigma^2$ & \multicolumn{3}{c}{3.29} & \multicolumn{3}{c}{3.29} \\
    $\tau_{00}$ & \multicolumn{3}{c}{0.06 model} & \multicolumn{3}{c}{0.13 model} \\
    ICC & \multicolumn{3}{c}{0.02} & \multicolumn{3}{c}{0.04} \\
    N & \multicolumn{3}{c}{12 model} & \multicolumn{3}{c}{12 model} \\
    \midrule
    Observations & \multicolumn{3}{c}{24000} & \multicolumn{3}{c}{24000} \\
    Marginal $R^2$ / Conditional $R^2$ & \multicolumn{3}{c}{0.032 / 0.048} & \multicolumn{3}{c}{0.031 / 0.067} \\
    \bottomrule
  \end{tabular}
\end{table}

\begin{table}[htbp] \footnotesize
  \centering
  \caption{Partisan and Base Models Comparison}
  \label{tableBothModelsOverallInteraction}
  \begin{tabular}{lcccccc}
\toprule
\multicolumn{1}{c}{\textbf{}} & \multicolumn{3}{c}{\textbf{Positive}} & \multicolumn{3}{c}{\textbf{Negative}} \\
\cmidrule(r){2-4} \cmidrule(r){5-7}
Predictors & Odds Ratios & CI & $p$ & Odds Ratios & CI & $p$ \\
\midrule
    (Intercept) & 0.26 & 0.21 -- 0.32 & \textbf{$<$0.0001} & 0.09 & 0.07 -- 0.11 & \textbf{$<$0.0001} \\
    source [we] & 1.86 & 1.76 -- 1.96 & \textbf{$<$0.0001} & & & \\
    total tokens scaled & 1.07 & 1.05 -- 1.10 & \textbf{$<$0.0001} & 1.10 & 1.08 -- 1.13 & \textbf{$<$0.0001} \\
    fine tuned [1] & 0.41 & 0.39 -- 0.43 & \textbf{$<$0.0001} & 2.13 & 2.00 -- 2.28 & \textbf{$<$0.0001} \\
    TTR & 1.87 & 1.53 -- 2.29 & \textbf{$<$0.0001} & 1.44 & 1.17 -- 1.77 & \textbf{0.0006} \\
    source [we] $\times$ fine tuned [1] & 2.48 & 2.32 -- 2.66 & \textbf{$<$0.0001} & & & \\
    source [they] & & & & 1.90 & 1.76 -- 2.04 & \textbf{$<$0.0001} \\
    source [they] $\times$ fine tuned [1] & & & & 3.40 & 3.12 -- 3.69 & \textbf{$<$0.0001} \\
    \midrule
    \multicolumn{7}{l}{\textbf{Random Effects}} \\
    $\sigma^2$ & \multicolumn{3}{c}{3.29} & \multicolumn{3}{c}{3.29} \\
    $\tau_{00}$ & \multicolumn{3}{c}{0.01 model} & \multicolumn{3}{c}{0.01 model} \\
    & \multicolumn{3}{c}{0.00 fine.tuned.party} & \multicolumn{3}{c}{0.00 fine.tuned.party} \\
    N & \multicolumn{3}{c}{3 fine.tuned.party} & \multicolumn{3}{c}{3 fine.tuned.party} \\
    & \multicolumn{3}{c}{12 model} & \multicolumn{3}{c}{12 model} \\
    \midrule
    Observations & \multicolumn{3}{c}{72000} & \multicolumn{3}{c}{72000} \\
    Marginal $R^2$ / Conditional $R^2$ & \multicolumn{3}{c}{0.127 / NA} & \multicolumn{3}{c}{0.235 / NA} \\
    \bottomrule
  \end{tabular}
\end{table}

\begin{table}[htbp] \footnotesize
  \centering
  \caption{All pre-training corpora (human sentences)}
  \label{tablePreTraining}
  \begin{tabular}{lcccccc}
\toprule
\multicolumn{1}{c}{\textbf{}} & \multicolumn{3}{c}{\textbf{Positive}} & \multicolumn{3}{c}{\textbf{Negative}} \\
\cmidrule(r){2-4} \cmidrule(r){5-7}
Predictors & Odds Ratios & CI & $p$ & Odds Ratios & CI & $p$ \\
\midrule
    (Intercept) & 0.13 & 0.07 -- 0.25 & \textbf{$<$0.0001} & 0.31 & 0.14 -- 0.67 & \textbf{0.0031} \\
    source [we] & 1.68 & 1.57 -- 1.80 & \textbf{$<$0.0001} & & & \\
    total tokens scaled & 1.11 & 1.07 -- 1.16 & \textbf{$<$0.0001} & 0.93 & 0.88 -- 0.98 & \textbf{0.0067} \\
    TTR & 3.06 & 1.77 -- 5.31 & \textbf{0.0001} & 0.43 & 0.22 -- 0.82 & \textbf{0.0112} \\
    source [they] & & & & 1.69 & 1.55 -- 1.84 & \textbf{$<$0.0001} \\
    \midrule
    \multicolumn{7}{l}{Random Effects} \\
    \midrule
    $\sigma^2$ & \multicolumn{3}{c}{3.29} & \multicolumn{3}{c}{3.29} \\
    $\tau_{00}$ & \multicolumn{3}{c}{0.15 corpus} & \multicolumn{3}{c}{0.23 corpus} \\
    ICC & \multicolumn{3}{c}{0.04} & \multicolumn{3}{c}{0.07} \\
    N & \multicolumn{3}{c}{4 corpus} & \multicolumn{3}{c}{4 corpus} \\
    \midrule
    Observations & \multicolumn{3}{c}{16000} & \multicolumn{3}{c}{16000} \\
    Marginal $R^2$ / Conditional $R^2$ & \multicolumn{3}{c}{0.022 / 0.064} & \multicolumn{3}{c}{0.021 / 0.086} \\
    \bottomrule
  \end{tabular}
\end{table}

\begin{table}[htbp] \footnotesize
  \centering
  \caption{Human vs LLMs}
  \label{tableHumanvsLLM}
  \begin{tabular}{lcccccc}
\toprule
\multicolumn{1}{c}{\textbf{}} & \multicolumn{3}{c}{\textbf{Positive}} & \multicolumn{3}{c}{\textbf{Negative}} \\
\cmidrule(r){2-4} \cmidrule(r){5-7}
Predictors & Odds Ratios & CI & $p$ & Odds Ratios & CI & $p$ \\
\midrule
    (Intercept) & 0.20 & 0.17 -- 0.23 & \textbf{$<$0.0001} & 0.15 & 0.12 -- 0.18 & \textbf{$<$0.0001} \\
    source [we] & 1.89 & 1.84 -- 1.94 & \textbf{$<$0.0001} & & & \\
    humanTRUE & 0.81 & 0.77 -- 0.86 & \textbf{$<$0.0001} & 1.14 & 1.06 -- 1.22 & \textbf{0.0002} \\
    total tokens scaled & 1.14 & 1.13 -- 1.16 & \textbf{$<$0.0001} & 0.95 & 0.93 -- 0.97 & \textbf{$<$0.0001} \\
    TTR & 2.56 & 2.19 -- 3.00 & \textbf{$<$0.0001} & 0.91 & 0.74 -- 1.11 & 0.3312 \\
    source [we] × humanTRUE & 0.87 & 0.81 -- 0.94 & \textbf{0.0002} & & & \\
    source [they] & & & & 2.09 & 2.03 -- 2.16 & \textbf{$<$0.0001} \\
    source [they] × humanTRUE & & & & 0.79 & 0.73 -- 0.87 & \textbf{$<$0.0001} \\
    \midrule
    \multicolumn{1}{l}{Observations} 
    & \multicolumn{3}{c}{128000} & \multicolumn{3}{c}{128000} \\
    R$^2$ Tjur & \multicolumn{3}{c}{0.030} & \multicolumn{3}{c}{0.018} \\
    \bottomrule
  \end{tabular}
\end{table}

\begin{table}[!ht] \centering \footnotesize
  \caption{Counts of sentences for all non-finetuned models} 
  \label{tableCounts} 
\begin{tabular}{lcccc} 
\hline
Model & Ingroup Positive & Ingroup Negative & Outgroup Positive & Outgroup Negative \\ 
\hline \\[-1.8ex] 
BLOOM-1B1 & 559 & 84 & 434 & 170 \\ 
BLOOM-1B7 & 582 & 73 & 340 & 198 \\ 
BLOOM-3B & 518 & 98 & 363 & 194 \\ 
BLOOM-560M & 504 & 87 & 389 & 133 \\ 
BLOOMZ-1B1 & 486 & 87 & 328 & 122 \\ 
BLOOMZ-1B7 & 344 & 109 & 234 & 123 \\ 
BLOOMZ-3B & 423 & 111 & 292 & 129 \\ 
BLOOMZ-560M & 351 & 91 & 354 & 102 \\ 
Cerebras-GPT-1.3B & 495 & 126 & 300 & 219 \\ 
Cerebras-GPT-111M & 318 & 105 & 264 & 227 \\ 
Cerebras-GPT-13B & 485 & 119 & 317 & 246 \\ 
Cerebras-GPT-2.7B & 445 & 129 & 297 & 264 \\ 
Cerebras-GPT-256M & 277 & 157 & 198 & 219 \\ 
Cerebras-GPT-590M & 503 & 102 & 276 & 191 \\ 
Cerebras-GPT-6.7B & 456 & 138 & 253 & 221 \\ 
Dolly2.0-12B & 485 & 127 & 310 & 225 \\ 
Dolly2.0-3B & 465 & 126 & 272 & 286 \\ 
Dolly2.0-7B & 453 & 124 & 317 & 196 \\ 
GPT-2-124M & 482 & 139 & 271 & 275 \\ 
GPT-2-Large-774M & 444 & 135 & 271 & 277 \\ 
GPT-2-Medium-355M & 459 & 159 & 269 & 272 \\ 
GPT-2-XL-1.5B & 442 & 156 & 261 & 254 \\ 
Gemma-7B & 596 & 20 & 477 & 147 \\ 
Gemma-7B-IT & 377 & 16 & 350 & 127 \\ 
J2-Jumbo-Instruct & 511 & 71 & 341 & 130 \\ 
LLaMA-13B & 656 & 20 & 437 & 141 \\ 
LLaMA-30B & 650 & 22 & 471 & 126 \\ 
LLaMA-65B & 622 & 21 & 426 & 121 \\ 
LLaMA-7B & 627 & 38 & 456 & 114 \\ 
Llama 2-13B & 636 & 25 & 490 & 133 \\ 
Llama 2-70B & 644 & 36 & 446 & 146 \\ 
Llama 2-7B & 658 & 31 & 456 & 118 \\ 
Mistral-7B & 564 & 57 & 394 & 156 \\ 
Mixtral-8x7B & 577 & 55 & 382 & 174 \\ 
OLMo-7B & 588 & 52 & 440 & 149 \\ 
OPT-1.3B & 423 & 204 & 278 & 306 \\ 
OPT-125M & 448 & 168 & 294 & 317 \\ 
OPT-13B & 418 & 215 & 256 & 361 \\ 
OPT-2.7B & 433 & 219 & 236 & 360 \\ 
OPT-30B & 401 & 223 & 245 & 359 \\ 
OPT-350M & 422 & 204 & 252 & 340 \\ 
OPT-6.7B & 464 & 192 & 240 & 367 \\ 
OPT-66B & 382 & 219 & 246 & 358 \\ 
OPT-IML-1.3B & 351 & 268 & 238 & 360 \\ 
OPT-IML-30B & 421 & 197 & 215 & 383 \\ 
Pythia-1.4B & 479 & 132 & 311 & 227 \\ 
Pythia-12B & 429 & 144 & 272 & 269 \\ 
Pythia-160M & 387 & 127 & 208 & 261 \\ 
Pythia-1B & 482 & 142 & 279 & 245 \\ 
Pythia-2.8B & 461 & 154 & 284 & 254 \\ 
Pythia-410M & 450 & 143 & 297 & 233 \\ 
Pythia-6.9B & 468 & 114 & 304 & 211 \\ 
Pythia-70M & 288 & 104 & 166 & 248 \\ 
davinci & 357 & 168 & 219 & 261 \\ 
text-bison@001 & 493 & 94 & 499 & 98 \\ 
text-davinci-003 & 517 & 89 & 285 & 164 \\ 
\hline \\[-1.8ex] 
\end{tabular} 
\end{table} 

\begin{table}[!htbp] \centering
  \caption{Basic statistics, non-finetuned models overall} 
  \label{tableBasicStats} 
\begin{tabular}{lccccc} 
\hline 
Statistic & \multicolumn{1}{c}{N} & \multicolumn{1}{c}{Mean} & \multicolumn{1}{c}{St. Dev.} & \multicolumn{1}{c}{Min} & \multicolumn{1}{c}{Max} \\ 
\hline \\[-1.8ex] 
Ingroup Positive & 112,000 & 0.238 & 0.426 & 0 & 1 \\ 
Ingroup Negative & 112,000 & 0.059 & 0.235 & 0 & 1 \\ 
Outgroup Positive & 112,000 & 0.159 & 0.366 & 0 & 1 \\ 
Outgroup Negative& 112,000 & 0.110 & 0.313 & 0 & 1 \\ 
RoBERTa Positive & 112,000 & 0.397 & 0.489 & 0 & 1 \\ 
RoBERTa Negative & 112,000 & 0.169 & 0.374 & 0 & 1 \\ 
RoBERTa Prob. & 112,000 & 0.776 & 0.150 & 0.338 & 0.993 \\ 
Total Tokens & 112,000 & 16.885 & 8.805 & 1 & 74 \\  
TTR & 112,000 & 0.916 & 0.093 & 0.100 & 1.000 \\ 
CTTR & 112,000 & 2.540 & 0.538 & 0.394 & 4.583 \\ 
VADER Positive & 112,000 & 0.488 & 0.500 & 0 & 1 \\ 
VADER Negative & 112,000 & 0.142 & 0.349 & 0 & 1 \\ 
VADER Sentiment & 112,000 & 0.207 & 0.398 & $-$0.990 & 0.995 \\ 
LIWC Positive & 112,000 & 0.435 & 0.496 & 0 & 1 \\ 
LIWC Negative & 112,000 & 0.095 & 0.294 & 0 & 1 \\
LIWC Sentiment & 112,000 & 0.547 & 1.113 & $-$11 & 18 \\ 
LIWC Positive Word Count & 112,000 & 0.728 & 0.966 & 0 & 18 \\ 
LIWC Negative Word Count & 112,000 & 0.181 & 0.499 & 0 & 12 \\ 
\hline \\[-1.8ex] 
\end{tabular} 
\end{table}

\end{appendices}

\end{document}